\documentclass[acmsmall,natbib=false]{acmart}

\usepackage{booktabs} % For formal tables
\usepackage{subcaption}
\usepackage{multirow}
\usepackage{enumitem}
\usepackage{array}
\usepackage{bm}
\usepackage{amsthm}
\usepackage{amsmath}
\usepackage{amsfonts}
\usepackage{threeparttable}

\usepackage{graphicx}
\usepackage{algorithm}
\usepackage{algpseudocode}
\usepackage{verbatim}
 \usepackage{mathtools}
\usepackage{xcolor}
\usepackage{amsthm}
\usepackage{makecell}

\usepackage{lineno}

\newcommand{\CHENG}{}
\newcommand{\LIANG}{}

\newcommand{\LIANGTwo}{}
\newcommand{\CHENGTwo}{}

\newcommand{\LIANGMajor}{}
\newcommand{\chengr}{}
\newcommand{\liangr}{}

\newcommand{\bodyr}{}

\AtBeginDocument{%
  }

\setcopyright{acmcopyright}
\copyrightyear{2022}
\acmYear{2022}
\acmDOI{XXXXXXX.XXXXXXX}

\acmJournal{JACM}
\acmVolume{37}
\acmNumber{4}
\acmArticle{111}
\acmMonth{8}

\RequirePackage[
  datamodel=acmdatamodel,
  style=acmauthoryear,
  ]{biblatex}

\begin{document}
% \linenumbers 
%%
%% The "title" command has an optional parameter,
%% allowing the author to define a "short title" to be used in page headers.
% \title{Road Network Representation Learning: Hypergraph is Also What You Need}

% \input{inputs/response.tex}
% \newpage
% \linenumbers

\title{{\bodyr{Road Network Representation Learning: A Dual Graph based Approach}}}

%%
%% The "author" command and its associated commands are used to define
%% the authors and their affiliations.
%% Of note is the shared affiliation of the first two authors, and the
%% "authornote" and "authornotemark" commands
%% used to denote shared contribution to the research.
\author{Liang~Zhang}
\email{liang012@e.ntu.edu.sg}
% \orcid{1234-5678-9012}

\author{Cheng~Long}
\email{c.long@ntu.edu.sg}
\affiliation{%
  \institution{School of Computer Science and Engineering, Nanyang Technological University}
  \streetaddress{50 Nanyang Ave}
  \country{Singapore}
  \postcode{639798}
}
\authornote{Corresponding author.}

 % \thanks{This study is supported under the RIE2020 Industry Alignment Fund Industry Collaboration Projects (IAF-ICP) Funding Initiative, as well as cash and in kind contribution from Singapore Telecommunications Limited Singtel, through Singtel Cognitive and Artificial Intelligence Lab for Enterprises (SCALE@NTU). This research is supported by the Ministry of Education, Singapore, under its Academic Research Fund (Tier 2 Award MOE-T2EP20221-0013). Any opinions, findings and conclusions or recommendations expressed in this material are those of the author(s) and do not reflect the views of the Ministry of Education, Singapore}

%%
%% By default, the full list of authors will be used in the page
%% headers. Often, this list is too long, and will overlap
%% other information printed in the page headers. This command allows
%% the author to define a more concise list
%% of authors' names for this purpose.
\renewcommand{\shortauthors}{Zhang et al.}

%%
%% The abstract is a short summary of the work to be presented in the
%% article.
\begin{abstract}
  Road network is a critical infrastructure powering many applications including transportation, mobility and logistics in real life. To leverage the input of a road network across these different applications, it is necessary to learn the representations of the roads in the form of vectors, which is named \emph{road network representation learning} (RNRL). While several models have been proposed for RNRL, they capture the pairwise relationships/connections among roads only (i.e., as a simple graph), and fail to capture among roads the high-order relationships (e.g., those roads that jointly form a local region usually have similar features such as speed limit) and long-range relationships (e.g., some roads that are far apart may have similar semantics such as being roads in residential areas). Motivated by this, we propose to construct a \emph{hypergraph}, where each hyperedge corresponds to a set of multiple roads forming a region. The constructed hypergraph would naturally capture the high-order relationships among roads with hyperedges. We then allow information propagation via both the edges in the simple graph and the hyperedges in the hypergraph in a graph neural network context. 
  % In this way, two nodes far apart can exchange their information via hyperedges as bridges (i.e., long-range relationships would be captured). 
  In addition, we introduce different pretext tasks based on both the simple graph (i.e., graph reconstruction) and the hypergraph (including hypergraph reconstruction and hyperedge classification) for optimizing the representations of roads. 
%   {\LIANGMajor{Equipped with the specially designed hyperedge classification task, two roads far apart on the map can establish connections via hyperedges as bridges (i.e., long-range relationships could be captured).}} 
{\bodyr {\chengr The graph reconstruction and hypergraph reconstruction tasks are conventional ones and can capture structural information. The hyperedge classification task can capture long-range relationships between pairs of roads that belong to hyperedges with the same label.}}
  We call the resulting model \emph{HyperRoad}. We further extend HyperRoad to problem settings when additional inputs of road attributes and/or trajectories that are generated on the roads are available. We conduct extensive experiments on two real datasets, for five downstream tasks, and under four problem settings, which demonstrate that our model achieves impressive improvements compared with existing baselines across datasets, tasks, problem settings and performance metrics.
\end{abstract}

%%
%% The code below is generated by the tool at http://dl.acm.org/ccs.cfm.
%% Please copy and paste the code instead of the example below.


%%
%% Keywords. The author(s) should pick words that accurately describe
%% the work being presented. Separate the keywords with commas.

% \received{20 February 2007}
% \received[revised]{12 March 2009}
% \received[accepted]{5 June 2009}

%%
%% This command processes the author and affiliation and title
%% information and builds the first part of the formatted document.
\maketitle

\section{Introduction} \label{sec:intro}

% As the fundamental component of the transportation systems, road network is closely related to numerous downstream transportation applications, including traffic based tasks such as traffic inference and forecasting \cite{guo2019attention}, and road segment based tasks such as road tag prediction \cite{yin2021multimodal}. Due to its important role, studying the road network representations can help us explore the functions and properties of transportation systems, and directly boost the effectiveness of all these tasks. 
Road network is a fundamental infrastructure supporting many real life applications, including but not limited to transportation, logistics, mobility, city management, planning, etc. Correspondingly, for various tasks such as traffic inference and forecasting~\cite{guo2019attention}, road tag prediction~\cite{yin2021multimodal} and arrival time estimation~\cite{yuan2020effective}, the road network data {\LIANGTwo is used as a key input}. Nevertheless, road network in its {\LIANGTwo raw} form cannot be directly inputted to machine learning models designed for these tasks. A common practice is to learn representations of a road network (specifically its roads) in the forms of vectors, called \emph{road network representation learning} (RNRL), which captures essential information of the road network, and then use the learned representations for various downstream tasks~\cite{wang2019learning,jepsen2020relational,wu2020learning,chen2021robust}. 

% {\color{red}{Which kind of references should be added here (last sentence)? About RNRL or downstream tasks?}} {\color{blue}Should be for RNRL}

% There is a wealth of information on the road network, including spatial information such as  road network structures and geo-locations, semantic information such as manually annotated road attributes, and trajectory information. Among them, the spatial information is inherent to the road network and can be used in any area. While the semantic and trajectory information may suffer from coverage and availability problems. Meanwhile, although many existing studies have been conducted on the road network representation learning task using different information, the spatial information modelling is still the most essential component of these models. For example, IRN2Vec \cite{wang2019learning} models road network structure as a base and consider road attributes as additional loss function. Therefore, how to model road network structures and positions effectively and incorporate other information flexibly is important for road network representation learning.
What is intrinsic to a road network is its structural information, which refers to the roads of the road network and the connections among the roads. Quite a few models have been proposed to capture the structural information of a road network for RNRL~\cite{wang2019learning,jepsen2020relational,wu2020learning,chen2021robust}. Among them, some studies~\cite{wang2019learning,chen2021robust} follow the node2vec idea and some others~\cite{jepsen2020relational,wu2020learning} adopt graph neural networks (GNN). Specifically, in~\cite{wang2019learning,chen2021robust}, it first collects some paths, each corresponding to a sequence of roads, via shortest path sampling~\cite{wang2019learning} or random walks~\cite{chen2021robust}, and then learns the representations of roads via the skip-gram model~\cite{mikolov2013distributed} based on collected paths. In~\cite{jepsen2020relational,wu2020learning}, it feeds the graph structure that models the structural information of a road network to GNN and learns the representations of roads by optimizing a graph reconstruction pretext task. 
It can be observed that all these models model the structural information of a road network as a \emph{simple graph} and captures only pairwise connections/relationships among roads for RNRL. 

\begin{figure}[t!]
	\centering
	\includegraphics[width=0.50\textwidth]{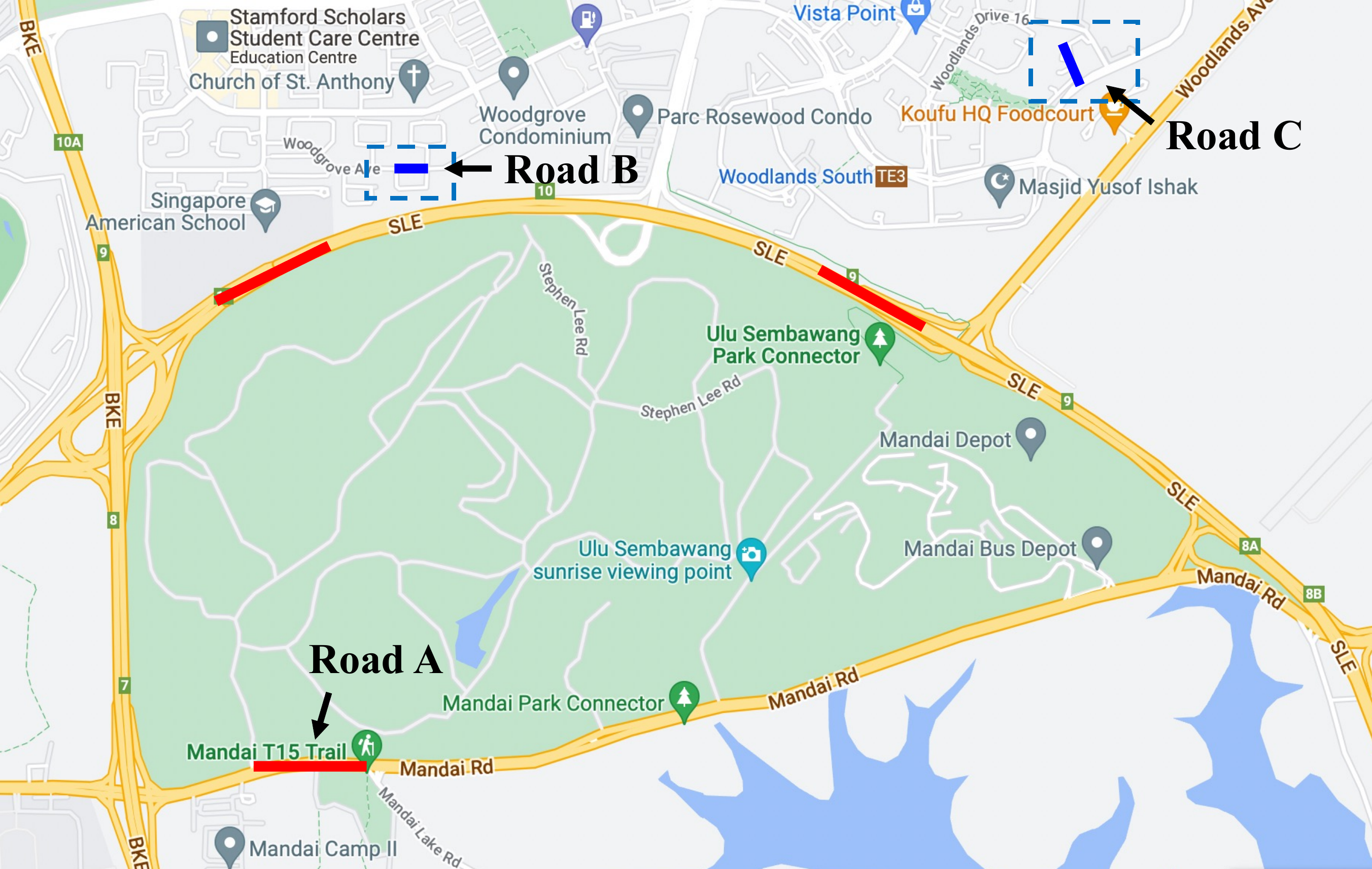}
	\caption{Road network example.}
	\label{fig:intro}
	\vspace{-0.1cm}
\end{figure}

We observe that apart from pairwise relationships, \emph{high-order relationships} and \emph{long-range relationships} commonly exist in road networks, yet the latter two are hardly captured by the simple graph structure used by existing models~\cite{wang2019learning,jepsen2020relational,wu2020learning,chen2021robust}. 
{\CHENGTwo Here, a high-order relationship is one among more than two roads and a long-range relationship is one among two roads that are not necessarily 
% to be directly linked or 
to have close spatial proximity.}
For an example of the high-order relationships, consider the three roads marked in red in Fig.~\ref{fig:intro}. These roads together with some others surrounding a park region (the area in green in the middle of the figure) constitute a highway loop. This relationship involves many roads and thus it is a high-order relationship, which helps identify the urban function of ``Road A''. 
% The data analysis results in Section \ref{sec:p_ie} also demonstrate strong correlations between a road and its high-order relationship (e.g. “Road A” and the highway loop).
% {\CHENGTwo We analyze some real dataset and find that roads that form regions collectively show much stronger correlations in their attributes compared with those that do not, which validates the existence of high-order relationships (Details can be found in Section~\ref{sec:p_ie}).}
However, existing models~\cite{wang2019learning,jepsen2020relational,wu2020learning,chen2021robust} fail to capture this high-order relationship and may learn low-quality representations. For example, for ``Road A'' in the figure, existing GNN-based models
%~\cite{jepsen2020relational,wu2020learning} 
may misinterpret its function and generate {\CHENGTwo its representation mainly based on its} pairwise connections with walk roads nearby. For an example of the long-range relationship, consider the two roads marked in blue in Fig.~\ref{fig:intro}, namely ``Road B'' and ``Road C''. These two roads share similar information (e.g., 
%they are both adjacent to the highway loop
they are both residential roads with low speed limit), despite being far apart on the map. 
% Our statistic results in Section \ref{sec:ssl} also validates that roads with long-range relationship are much more correlated than random roads.
% {\CHENGTwo Again, we validate the existence of long-range relationships based on the analysis on real datasets (Details can be found in Section~\ref{sec:ssl}).}
% }
% {\color{red}{we don't capture the correlations like near a highway loop explicitly, the example may cause misleading.}
% % 
% Cheng: I think we do since the propagation can be done with the common hyperedge (i.e., the highway). Am I right?
% % 
% Yes, my concern is that these two roads B \& C do not involve in the same highway loop (hyperedge). And I also refer this case in the hyperedge classification task. The emphasize of current description may be more consistent with the motivations. You may help to check whether they are consistent.
% % 
% Cheng: 
% }
Again, existing models~\cite{wang2019learning,jepsen2020relational,wu2020learning,chen2021robust} are largely incapable of capturing this long-range relationship.

\begin{figure*}[t!]
	\centering
	\includegraphics[width=0.99 \textwidth]{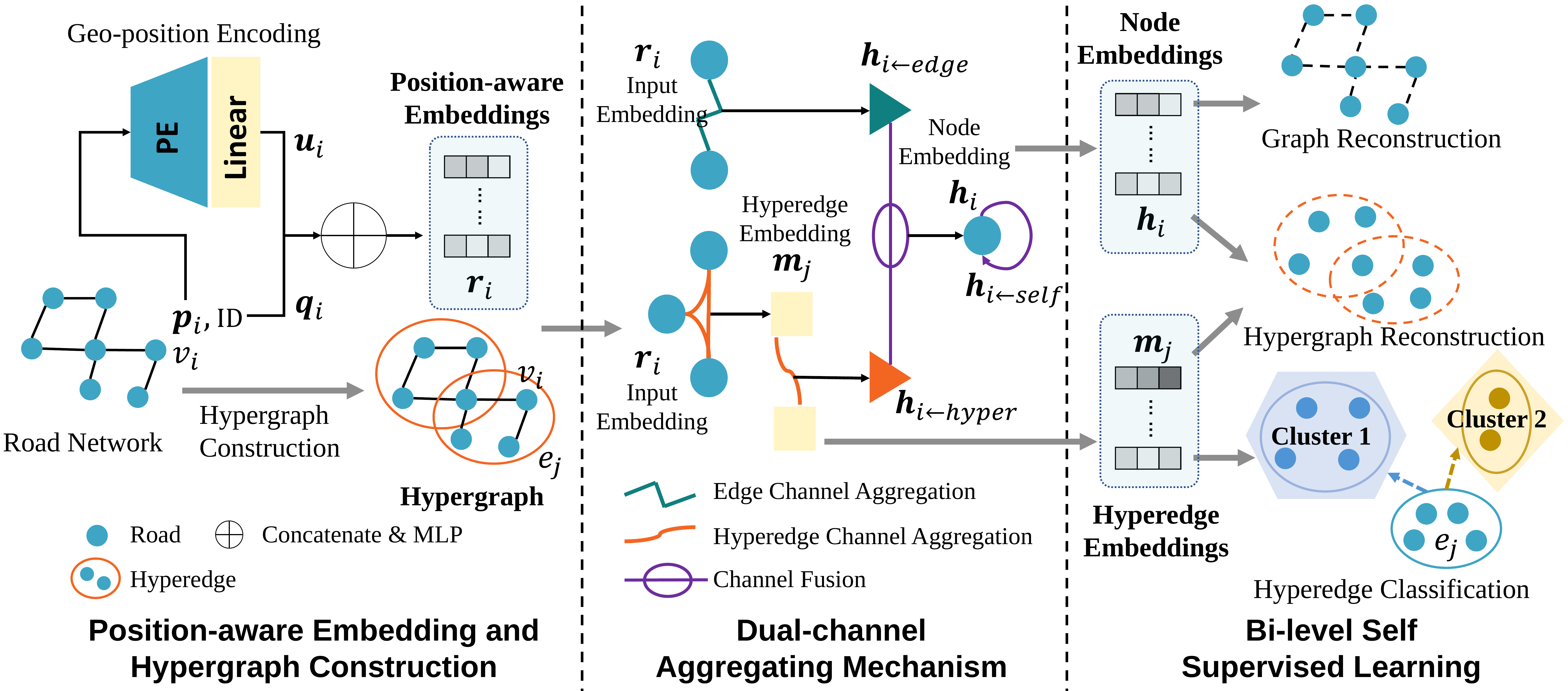}
	\caption{The architecture of HyperRoad.}
	\label{fig:framework}
\vspace*{-0.1cm}
\end{figure*}

In this paper, we propose a solution called HyperRoad (\underline{Hyper}graph-Oriented \underline{Road} network representation), which is capable of capturing the pairwise, high-order and long-range relationships among the roads all at the same time. We explain its intuitions and major ideas as follows. \underline{First}, based on the observation that some roads that form a region may have some similarity in terms of semantics and functions (e.g., a highway loop), we propose to first extract all regions that are naturally formed by roads (details will be introduced in Section~\ref{sec:p_ie}). Then for each extracted region, we construct a \emph{hyperedge} connecting all the roads that form the region. The constructed hyperedges would naturally capture the high-order relationships among the roads and constitute a \emph{hypergraph}. For example, for the road network shown in Fig.~\ref{fig:intro}, we would extract a region that corresponds to the park and construct a hyperedge connecting all roads that constitute the highway loop surrounding the park. 

\underline{Second}, we design a \emph{dual-channel aggregating mechanism} for propagating the information through two channels in a context of GNN, one for the simple graph and the other for the constructed hypergraph. 
% One is based on the \emph{simple} graph, which is based on pairwise relationships/connections among roads. The other is based on the constructed \emph{hypergraph}, which is based on the high-order relationships among roads. 
We then fuse the aggregated information from two channels via a gating mechanism as the \emph{road embeddings}. This dual-channel aggregating mechanism has its power reflected by {\bodyr {\LIANGMajor{(1) it captures the pairwise relationships via the channel based on simple graph as existing models do, and (2) it captures high-order relationships via the channel based on the hypergraph (which is new).}}
% and (3) it facilitates to capture the long-range relations since two roads that are far apart can share information from each other via the hyperedges (e.g., the hyperedges serve as ``bridges''). 
{\LIANGMajor{It is worth noting that although hypergraph and hypergraph neural network have been applied to enhance spatial-temporal mining in some existing works \cite{xia2021spatial, luo2022directed}, our work is the first attempt to {\chengr leverage} the power of hypergraph {\chengr for the RNRL problem and more generally road network modeling}.}} }

\underline{Third}, we design a \emph{bi-level self-supervised learning module} for optimizing the road embeddings. At the simple graph level, we adopt the graph reconstruction pretext task as some existing models do~\cite{jepsen2020relational,wu2020learning}. At the hypergraph level, we adopt two pretext tasks, namely 
{\bodyr {\LIANGMajor{hypergraph reconstruction and hyperedge classification.}}}
% hypergraph reconstruction and hypergraph classification. 
% The hypergraph reconstruction task is natural and adopted from the literature of hypergraph representation learning.
The hypergraph reconstruction task is natural for hypergraph representation learning.
The hyperedge classification task, which is to perform a classification task on hyperedges (with the labels defined based on some intrinsic properties of hyperedges such as their sizes), is newly proposed for the constructed hyperedges based on road networks. 
{\bodyr {\chengr The hyperedge classification task can capture long-range relationships between pairs of roads that belong to hyperedges with the same label.}}

Apart from the structural information, some other types of information are associated with or closely related to a road network and have been used for RNRL. These include (1) spatial information, which refers to the locations of roads (e.g., the coordinates of the middle points of roads), (2) road attributes (e.g., 
%road length, 
road type, lane number, etc), and (3) trajectories, which are sequences of roads that are generated by vehicles on the roads. 
% We argue that (1) spatial information is naturally associated with road networks and always available for RNRL, which is the same as the structural information and (2) attributes information and trajectories information are also naturally associated with road networks, but may not be always publicly available for RNRL.
Different from the structural and spatial information, the attributes and trajectories are not always available for RNRL.
% or complete for RNRL.
For example, for the road networks we downloaded from OpenStreetMap in City Xi'an, the lane attribute is missing for 69\% of the roads. The trajectories are available for some specific cities only and even for those cities some trajectories are available, they are not complete and cover only a small fraction of roads.
Therefore, we tackle the RNRL problem under four settings depending on the availability of the attributes and trajectories. The default setting is that we only have the structural and spatial information since it is deemed to be intrinsic to a road network.
% since this should be the mostly common case in practice. 
Other settings include one with additional attributes, one with additional trajectories, and the last one with both additional attributes and trajectories. For the default setting, we incorporate a \emph{position-aware embedding generation} module, which embeds the location of each road into its embedding to be inputted to the dual-channel aggregating mechanism. We still call the resulting model HyperRoad. For the setting with the additional attributes, we incorporate the attribute embeddings to the initial embeddings of roads and introduce an additional pretext task of attribute reconstruction in the bi-level self-supervised learning module. We call the resulting model \emph{HyperRoad-A}. For the setting with the additional trajectories, we follow an existing study~\cite{chen2021robust} and introduce a BERT like structure on top of HyperRoad for fine-tuning the road embeddings learnt by HyperRoad with the trajectories. We call the resulting model \emph{HyperRoad-T}. For the setting with both additional attributes and additional trajectories, we merge HyperRoad-A and HyperRoad-T and obtain a model called \emph{HyperRoad-AT}.
% Furthermore, we propose three advanced models that take different kinds of additional information into consideration, e.g., road attributes, trajectory, and both of them, which is able to handle with different situations when learning road network representations. The proposed framework is generic in the sense that it is able to support any road network modelling especially a newly constructed road network without any extra information, and it also outperforms the state-of-the-art models when different extra information can be available.  

% We review the related work in Section~\ref{sec:relate} and present the problem definition in Section\ref{sec:pf}
% The major contributions are summarized as follows.
To summarize, our contributions are three-fold:
\begin{itemize}[leftmargin=*]
\item  We observe the insufficiency of existing RNRL models for capturing the high-order relations and long-range relations among roads. We then design a novel model called HyperRoad, which (1) constructs a \emph{hypergraph} on the road network for the first time based on how roads form regions, (2) {\bodyr {\LIANGMajor{involves a dual-channel aggregating mechanism based on both the simple graph of the road network and the constructed hypergraph for capturing the pairwise and high-order relationships 
% and long-range relationships all 
at the same time}}}, and (3) uses a bi-level self-supervised learning module, which involves quite a few pretext tasks at both the simple graph level and the hypergraph level, {\bodyr {\LIANGMajor{for optimizing the road embeddings as well as {\chengr capturing} long-range relationships. In this way, our model is capable of capturing the {\chengr pairwise, high-order and long-range relationships} among the roads all at the same time.
}}} (Section~\ref{sec:basic_model}) 

\item We summarize the types of information of a road network, which can be used for RNRL. They include structural and spatial information (which is intrinsic and always available) and road attributes and trajectories (which are not always available or complete). Correspondingly, we identify four settings for RNRL depending on the availability of the road attributes and trajectories and propose corresponding solutions all with HyperRoad as the core. This shows that our HyperRoad model is fundamental and can be enhanced flexibly when additional information is available. This is the first systematical study on various settings of RNRL. (Section~\ref{sec:extension})

\item We conduct extensive experiments on two real datasets for five downstream tasks including both road based and trajectory based ones under different problem settings. The results show that our model outperforms existing models across \emph{all} tasks, under \emph{all} settings, and on \emph{both} datasets. (Section~\ref{sec:expsetup} and~\ref{sec:expresult})
\end{itemize}

% The remainder of the paper is organized as follows. Section~\ref{} reviews the related work, Section~\ref{} presents the problem definition, Section~\ref{}

% To summarize, our contributions are three-fold:
% \begin{itemize}[leftmargin=*]
% \item  For the first time, we propose to model road network as hypergraph, and design hypergraph construction strategies on road network. Based on that, a novel road network representation learning framework called HyperRoad is proposed, which is featured with a hypergraph based dual-channel aggregation module to model high-order road relations, %and enhance model expressive power, 
% and a bi-level self-supervised learning module to provide complementary supervision signals. %and capture both local and global road correlations. 

% \item For the first time, we consider the road network representation learning problem under different settings systematically, and propose three advanced models to extend the applications of HyperRoad flexibly. 

% \item We conduct extensive experiments on five downstream applications including both road segment based and trajectory based, and the results show that our model outperforms existing models across all these tasks under various settings. 
% \end{itemize}
% \input{inputs/related_work_v2}

\section{Problem Statement} \label{sec:pf}
% In this section, we present the notations and several preliminaries first and then define our problem formally.

% In this paper, we aim to learn a low dimensional vector representation for each road, which can be denoted as $\{\mathbf{h}_i | \mathbf{h}_i \in \mathbb{R}^{d} \}_{v_i \in \mathcal{V}}$ with $d$ as the embedding size. 
%
We consider four types of information, which are associated with or closely related to a road network and have been used for road network representation learning (RNRL). These include (1) structural information, which refers to a set of roads {\LIANGTwo (or nodes)} and the pairwise connections among them, (2) spatial information, which refers to the locations of roads (e.g., the coordinates of the middle points of roads), (3) road attributes (e.g., 
% road length, 
road type, lane number, etc), and (4) trajectories, which are sequences of roads that are generated by vehicles on the roads. We introduce some notations of the four types of information as follows.

\smallskip\noindent \textbf{Structural information.} It is usually represented by a simple (directed) graph $\mathcal{G_R}=(\mathcal{V}, \mathcal{E_R}, \mathcal{P})$, where $\mathcal{V} = \{v_1, v_2, \dots, v_N\}$ is a set of nodes each representing a road
% (i.e., roads) \footnote{We use node and road interchangeably in this paper.} 
and $\mathcal{E_R}$ is a set of edges. The corresponding adjacency matrix of $\mathcal{G_R}$ is represented as $\mathbf{A} \in \mathbb{R}^{N \times N}$, where the entry $\mathbf{A}_{i,j}$ is a binary value indicating whether there exists a link from road $v_i$ to road $v_j$. 
The structural information is intrinsic to the road network.

\smallskip\noindent \textbf{Spatial information.} Each node (or road) $v_i$ has a coordinate vector $\mathbf{p}_i$ ($\mathbf{p}_i \in \mathcal{P}$) including the latitude and longitude of its middle point, mapping the road into the geographic space. Spatial information is another type of intrinsic information of road network.

\smallskip\noindent \textbf{Road attributes.} For a road $v_i$, we denote its road attributes as $\mathbf{a}_i = \{a_{i,1}, a_{i,2}, ..., a_{i,m}\}$, where $m$ denotes the number of attributes of $v_i$. The road attributes convey some semantics information of roads.

\smallskip\noindent \textbf{Trajectories.} A trajectory $T=[v_i]_{i=1}^{n}$ corresponds to a sequence of adjacent roads, where $v_i$ is a road in $\mathcal{G_R}$. The trajectories on a road network can reflect the mobility patterns on the road network.

\smallskip\noindent \textbf{Problem statement.} Formally, the road network representation learning (RNRL) problem is to learn a low dimensional vector representation for each road, denoted as $\{\mathbf{h}_i | \mathbf{h}_i \in \mathbb{R}^{d} \}_{v_i \in \mathcal{V}}$ with $d$ as the embedding size. 
Among the four types of information, the structural and spatial information is intrinsic information and always available, while the attributes and trajectories are not always available or complete.
%
% For example, for the road networks we downloaded from OpenStreetMap in City Xi'an, the lanes attribute is missing for 69\% of the roads. The trajectories are available for some specific cities only and even for those cities some trajectories are available, they are not complete and cover only a small fraction of roads.
Therefore, we study the RNRL problem under four settings depending on the availability of the attributes and trajectories. The default setting is that we only have the intrinsic structural and spatial information.
% since this should be the mostly common case in practice. 
Other settings include one with additional attributes, one with additional trajectories, and the last one with both additional attributes and trajectories. 

\section{HyperRoad: The Core} \label{sec:basic_model}

{\CHENG{In this section, we propose a novel model called HyperRoad for road network representation learning (RNRL). The overview of the model architecture is shown in Fig. \ref{fig:framework}. HyperRoad is based on three key ideas: (1) it incorporates a positional encoding scheme for embedding the roads (called position-aware embedding) and constructs a \emph{hypergraph}, where each hyperedge corresponds to a set of multiple roads forming a region, (2) it propagates information via both edges (of a simple graph capturing the pairwise relationships) and hyperedges (of the constructed hypergraph) in a context of graph neural network, and (3) it optimizes the road representations with a few pretext tasks based on both the simple graph (i.e., graph reconstruction) and the constructed hypergraph (including hypergraph reconstruction and hyperedge classification). Next, we introduce these three ideas in turn.}}
% In this section, we detail the basic model HyperRoad for road network representation learning problem by considering pure road network structures and geo-locations. The overview of the model architecture is shown in Fig. \ref{fig:framework}. Firstly, to cope with the expressive power issue of existing models, we propose to construct hypergraph which can connect arbitrary number of roads naturally, and design a new dual-channel aggregation module to capture high-order contextual information instead of the pairwise modelling. Secondly, to deal with the long-range correlation problem, we find that the hyperedge can transfer knowledge across roads far apart on the map, and design two novel hyperedge based self supervised training tasks for road network setting beyond simple graph reconstruction. 

\subsection{Position-aware Embedding and Hypergraph Construction}
\label{sec:p_ie}

\smallskip\noindent\textbf{Position-aware Embedding.}
To represent the innate spatial information in road networks, we follow the strategy in BERT \cite{devlin2018bert} to jointly embed roads' IDs and locations into dense representations as the model inputs.
{\LIANG{Specifically, for each road $v_i$, we introduce an embedding layer to encode its ID as an embedding vector $\mathbf{q}_i$.}} 
{\CHENG{Here, an embedding layer corresponds to a special MLP layer without bias and non-linear function.}}
All road IDs' embeddings are represented as an embedding matrix $\mathbf{Q} \in \mathbb{R}^{N\times d}$, which will be learned in an end-to-end manner.

% {\color{red} Comments: The embedding vector $\mathbf{q}_i$ is not clear?}

% {\LIANG{Have changed.}}

% {\color{red}$\mathbf{Q} \in \mathbb{R}^{d}$ has been changed to $\mathbf{Q} \in \mathbb{R}^{N\times d}$}

% Then, motivated by recent advances in
In addition, inspired by the positional encoding of transformers~\cite{devlin2018bert} and spatial representation learning~\cite{mai2020multi}, we propose a positional encoder module to encode the spatial information of a road network. In specific, given the coordinate vector $\mathbf{p}_i$ of a road $v_i$, we use sine and cosine functions of different frequencies to generate its positional encoding $\mathbf{u}_i$ as follows.

\begin{equation}
\label{eq:pe}
%PE(\mathbf{p}_i) = 
\left\{
    \begin{aligned}
        & \textbf{u}_i(4k) = sin((\textbf{p}_i^{[0]}/\phi) / \lambda^{4k/d}), \\
        & \textbf{u}_i(4k+1) = cos((\textbf{p}_i^{[0]}/\phi) / \lambda^{4k/d}), \\
        & \textbf{u}_i(4k+2) = sin((\textbf{p}_i^{[1]}/\phi) / \lambda^{4k/d}), \\ 
        & \textbf{u}_i(4k+3) = cos((\textbf{p}_i^{[1]}/\phi) / \lambda^{4k/d}),
    \end{aligned} \right. 
\end{equation}
{\color{black}where $d$ denotes the embedding size (which is a multiple of 4), 
% {\LIANG{$4k$ denotes the $4k$-th dimension and $ 0 \leq k \leq \lfloor (d-1)/4 \rfloor$}}, 
$k$ is an integer in [0, d/4-1],}
$\textbf{p}_i^{[0]}$ and $\textbf{p}_i^{[1]}$ denote the projected latitude and longitude values, respectively, $\phi$ is the scale parameter to constrain the magnitude of projected values, and $\lambda$ is the frequency parameter similar to the positional encoding in transformers. The difference from those conventional positional encoding in transformers,  which consider 1D {\CHENGTwo positions}, is that we consider 2D locations in geospace.
% We extend 1D discrete positions in NLP to 2D continuous positions in geospace.

% {\color{red}The meaning of $k$-th dimension is not clear. }
% {\LIANG{Have changed.}}
% {\color{red}With the above definitions of $k$, some dimensions may not be defined. You can consider the case where $d =  9$, then, we have $0 \le k \le 1$, then we only define 8 dimensions. Please double check.}

% {\LIANG{Sorry, d needs to be a multiple of 4. For example, in Bert, d should be a a multiple of 2. It is the same.}}

Finally, for each road $v_i$, we fuse its ID embedding and positional embedding into a single one as follows.
\begin{equation}
\label{eq:input}
\mathbf{r}_i = \text{MLP}(\mathbf{q}_i \mathbin\Vert \mathbf{u}_i),
\end{equation}
where $\mathbin\Vert$ is the concatenate operation and $\text{MLP}$ is a fully connected neural network.

{\color{black}
\smallskip\noindent\textbf{Hypergraph Construction.}
As explained in Section~\ref{sec:intro}, there would usually exist some high-order relationships among roads that form a region. For example, those roads that constitute a highway loop surrounding a park in Fig.~\ref{fig:intro} would tend to have similar semantics (e.g., highway roads). 
Therefore, we propose to construct hyperedges based on how roads form regions. Specifically, we apply a map segmentation algorithm proposed in~\cite{yuan2012discovering} to find the polygons that are formed by roads. 
{\CHENG{Note this process is based on the structural and spatial information of a road network solely.}}
Then, for each found polygon, we construct a hyperedge, which involves all roads that constitute the polygon.
%
%{\color{teal}{For example, for an example area in Singapore, the found polygons are shown in Fig. \ref{fig:polygon}. We can find that the red highway roads (in Fig. \ref{fig:intro}}) locate in the same hyperedge.}
%
{\bodyr {\LIANGMajor{There are two major benefits with the constructed hypergraph. \underline{First}, the constructed hyperedges would naturally capture the high-order relationships among roads of a road network. 
% \underline{Second}, as will be shown in Section~\ref{sec:dual-channel}, the constructed hyperedges would facilitate the information propagation between nodes far apart and thus it helps with capturing long-range relationships among roads. 
\underline{Second}, as will be shown in the next section (Section \ref{sec:ssl}), the intrinsic properties associated with the constructed hyperedges (e.g., its size) can be used as inductive bias to design new self supervised learning pretext tasks for learning the representations of roads with long-range relationships.
}} }
}

Formally, a hypergraph based on the road network is defined as $\mathcal{G_H}=(\mathcal{V}, \mathcal{E_H})$, where $\mathcal{V}$ is a set of roads, and  $\mathcal{E_H} = \{e_1, e_2, ... , e_M\}$ represents the set of hyperedges. The topological structure of $\mathcal{G_H}$ can also be represented by an incidence matrix $\mathbf{H} \in \mathbb{R}^{N \times M}$, with entries defined as follows.
\begin{equation}
\label{eq:hg}
\mathbf{H}_{i,j} = \left\{
    \begin{aligned}
        1 & , & \text{if} \,\, v_i \in e_j, \\
        0 & , & \text{if} \,\, v_i \notin e_j.
    \end{aligned} \right.
\end{equation}

% {\color{red}Comments: (1) I commented the previous discussions here since they do not look clear to me. Are they talking about the same intuitions explained in the introduction section? If so, I think dropping them would be fine. Otherwise, you can add more intuitions here. (2) I commented the description of the clustering and I think it is better to move it to the part when we introduce the hyperedge classification pretext task. Please also specify the clustering algorithm with enough information for reproduction of the results.}

% {\LIANG{It is fine to drop them.}}

\begin{figure}[t!]
	\centering
	\includegraphics[width=0.70\textwidth]{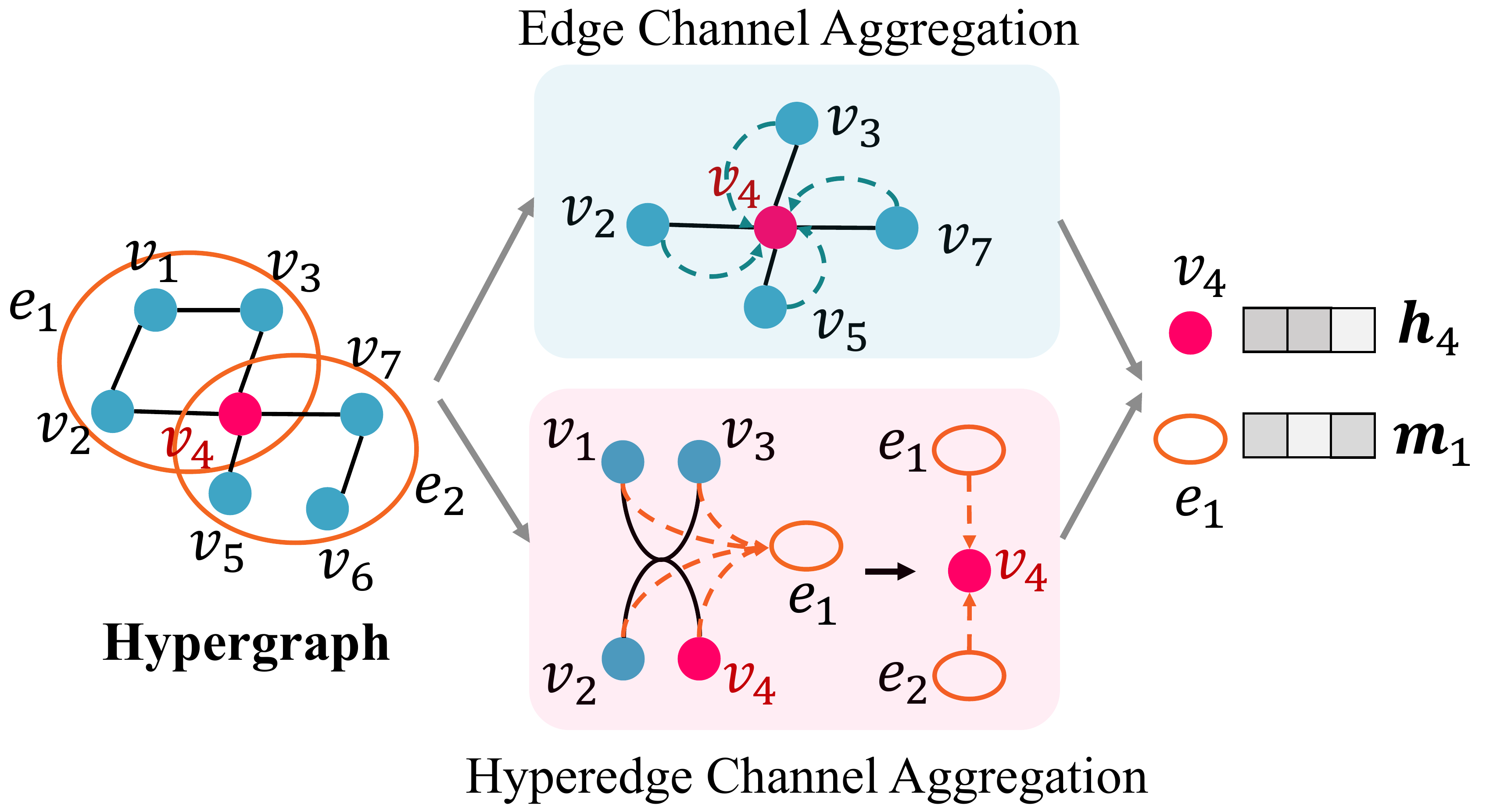}
	\caption{The illustration of the Dual-channel Aggregating Mechanism.}
	\label{fig:DAM}
\vspace{-0.1cm}
\end{figure}

\subsection{Dual-channel Aggregating Mechanism}
\label{sec:dual-channel}
% {\LIANGTwo The overview of our proposed Dual-channel Aggregating Mechanism is shown in Fig. \ref{fig:DAM}.} 
{\CHENGTwo An overview of the Dual-channel Aggregating Mechanism is shown in Fig.~\ref{fig:framework} (the middle part).}
In this module, 
% we describe our embedding aggregation scheme, which 
it iteratively aggregates the information from related roads to encode the structures. 
% Given the simple graph $\mathcal{G_R}$ and hypergraph $\mathcal{G_H}$, it is obvious that 
The intuition is that the edge level neighborhoods in the simple graph $\mathcal{G_R}$ capture pairwise relationships among roads and local receptive fields, and the hyperedge level neighborhoods in the hypergraph $\mathcal{G_H}$ capture the complex high-order relationships. Both are indispensable and complementary for road modeling. For example, the hyperedge context (e.g., highway loop) and local environment (e.g., residential area) can jointly determine the function of a ramp road. 
Therefore, we conduct message aggregation via two channels, namely edge channel and hyperedge channel, and devise a novel dual-channel aggregating mechanism. {\bodyr {\LIANGMajor{Note that some existing GNN methods \cite{morris2019weisfeiler, wang2021tree} also model high-order node relationships to help expand the receptive fields and/or improve the expressiveness of GNN model. For example, $k$-GNNs \cite{morris2019weisfeiler} aggregates the information from high-order neighbors defined by the $k$-order cliques, and TDGNN \cite{wang2021tree} aggregates the information from multi-hop neighbors defined by the tree decomposed subgraphs. These methods define different structures on the same simple graph. {\liangr In contrast}, our aggregation mechanism {\liangr takes} a dual graph perspective and aggregates information from both simple graph and hypergraph. In fact, we are aware of no existing methods which use multiple channels to combine a simple graph and a hypergraph together as ours does.}}}
In this section, we first illustrate the design of one-layer aggregation and then generalize it to multiple layers.

% {\color{red}Comments: There could be inconsistencies among ``e.g.,'', ``e.g.'' and ``\textit{e.g.}'', which need to be fixed. Better to stick to one form. I usually use ``e.g.,''. Same for ``i.e.,''.}

% {\LIANG{Have fixed.}}

\smallskip\noindent\textbf{Edge Channel Aggregation.} The goal of this sub-module is to propagate node information from the simple graph contexts. This process can be formulated as follows.
\begin{equation}
\label{eq:4}
	\mathbf{h}_{i \gets edge}^{(l+1)} = AGG(\{\mathbf{h}_{j}^{(l)}, \, v_j \in \mathcal{N}_i\}),
\end{equation}
where $\mathbf{h}_{i \gets edge}^{(l+1)}$ is the message received by node $v_i$ from the edge channel at layer $l+1$, 
{\CHENG{$\mathbf{h}_{i}^{(l)}$ denotes hidden state of node $v_j$ at layer $l$ ($\mathbf{h}_{j}^{(0)}$ is set to be the position-aware input embedding $\mathbf{r}_j$ for node $v_j$)}},
% $l$ represents the layer number, 
$\mathcal{N}_i$ denotes the neighborhoods of node $v_i$ defined by edges in $\mathcal{G_R}$, and $AGG(\cdot)$ is an aggregation function, which is implemented as a mean aggregator in this paper to keep the model simple. 

\begin{equation}
	\mathbf{h}_{i \gets edge}^{(l+1)} = \text{ReLU} ( \frac{1}{|\mathcal{N}_i|} \sum_{v_j \in \mathcal{N}_i} \, \mathbf{W}_{1}^{(l)} \mathbf{h}_{j}^{(l)} ),
\end{equation}
where $\mathbf{W}_{1}^{(l)}$ is a trainable weight matrix to distill
useful information for propagation and the normalization term $1 / |\mathcal{N}_i|$ is to constrain the embedding scales.

\smallskip\noindent\textbf{Hyperedge Channel Aggregation.} 
% In addition to edge aggregations, HyperRoad learns node representations through hypergraph convolutions, allowing to capture high-order context information of road to enhance the model representation ability. Specifically, we define our 
The hyperedge channel aggregation is based on the following two-step operations.
\begin{equation}
\begin{aligned}
	\mathbf{m}_{j}^{(l+1)} = AGG(\{\mathbf{h}_{i}^{(l)}, \, v_i \in \mathcal{R}_j\}), \\ 
	\mathbf{h}_{i \gets hyper}^{(l+1)} = AGG(\{\mathbf{m}_{j}^{(l+1)}, \, e_j \in \mathcal{H}_i\}),
\end{aligned}
\end{equation}
where $\mathbf{m}_{j}^{(l+1)}$ is the aggregated hyperedge representation of $e_j$ at layer $l+1$, $\mathcal{R}_j$ denotes the set of nodes connected to hyperedge $e_j$ in $\mathcal{G_H}$, $\mathbf{h}_{i \gets hyper}^{(l+1)}$ is the message received by node $v_i$ from hyperedge channel at layer $l+1$, and $\mathcal{H}_i$ represents the neighbor hyperedge set of node $v_i$. The two-step information propagation scheme performs ‘node-hyperedge-node’ feature transformation upon the hypergraph structure, and the hyperedge acts as an information medium during this process. 
{\CHENG{We note the above two-step operations are also adopted in some existing hypergraph learning studies~\cite{feng2019hypergraph}, and in this paper we use them in the context of a novel dual-channel aggregating mechanism.}}
In this paper, we simply implement those two $AGG(\cdot)$ functions by mean aggregators to avoid extra model parameters. 
% We leave the exploration of other operations to future work.
%
\begin{equation}
\label{eq:7}
\begin{aligned}
	\mathbf{m}_{j}^{(l+1)} = \text{ReLU} ( \frac{1}{|\mathcal{R}_j|} \sum_{v_i \in \mathcal{R}_j} \, \mathbf{W}_{2}^{(l)} \mathbf{h}_{i}^{(l)} ), \\ 
	\mathbf{h}_{i \gets hyper}^{(l+1)} = \text{ReLU} ( \frac{1}{|\mathcal{H}_i|} \sum_{e_j \in \mathcal{H}_i} \, \mathbf{W}_{3}^{(l)} \mathbf{m}_{j}^{(l+1)} ),
\end{aligned}
\end{equation}
where $\mathbf{W}_{2}^{(l)}$ and $\mathbf{W}_{3}^{(l)}$ are feature transformation matrices at different granularities. 

{\LIANGTwo An illustration of {\CHENGTwo the proposed dual-channel aggregating mechanism} on a toy example is shown in Fig. \ref{fig:DAM}.}

% {\color{red}Comments: Why we provide a reference for the two-step operations and then say ``the proposed two-step''. This is confusing since the former means it is an existing idea and the latter means it is a new idea.}

% {\LIANG{Have added.}}

{\LIANGTwo
\smallskip\noindent\textbf{Channel Fusion.} \label{sec:fusion}
In this stage, we aggregate all the messages to update the node representations. {\bodyr {\LIANGMajor{For a node, apart from aggregating information from other nodes, we add a residual connection \cite{he2016deep, vaswani2017attention} 
to preserve the node's own features. Meanwhile, the residual connection can also alleviate the over-smoothing problem when building deep graph neural networks and help model training, as demonstrated by existing work \cite{chen2020simple}. }}}
% and help model training. This method has also been applied in existing work \cite{chen2020simple} to help solve the over-smoothing problem in GNN training.}} 
We name this information channel as \emph{self channel}, which copies node embeddings directly to the next layer. Taking node $v_i$ as an example, we have message vector $\mathbf{h}_{i \gets self}^{(l+1)} = \mathbf{h}_{i}^{(l)}$. Formally, the new representation of $v_i$ can be formulated as follows.
\begin{equation}
	\mathbf{h}_{i}^{(l+1)} = f( \ \underbrace{\mathbf{h}_{i \gets self}^{(l+1)}}_{\text{self channel}}, \underbrace{\mathbf{h}_{i \gets edge}^{(l+1)}}_{\text{edge channel}}, \underbrace{\mathbf{h}_{i \gets hyper}^{(l+1)}}_{\text{hyperedge channel}}). 
\end{equation}

In this work, we implement $f(\cdot)$ using the \textit{Mean} function by default in HyperRoad, 
%to keep the model simple
and empirically compare other designs (e.g., attention mechanism) in Section \ref{sec:q2}. }
 
% too detailed 
\if 0
% one of the following three types of functions.
\begin{itemize}[leftmargin=*]
  \item \textit{Mean Fusion} averages the messages from the three channels as the output without trainable parameters.
    \begin{equation}
    	f_{mean} = (\mathbf{h}_{i \gets self}^{(l+1)} + \mathbf{h}_{i \gets edge}^{(l+1)} + \mathbf{h}_{i \gets hyper}^{(l+1)} ) / 3.
    \end{equation}
    
    \item \textit{Attention Fusion} learns message importance through an attention mechanism and regards  the output as the weighted sum of messages. 
    
    \begin{equation}
    f_{att} = \theta_s \cdot \mathbf{h}_{i \gets self}^{(l+1)} + \theta_e \cdot \mathbf{h}_{i \gets edge}^{(l+1)} + \theta_h \cdot \mathbf{h}_{i \gets hyper}^{(l+1)}.
    \end{equation}
    Without loss of generality, we take the weight $\theta_s$ as an example. It is defined as follows.
    
    \begin{equation}
    \begin{aligned}
    \tilde{\theta}_{s} = \text{MLP}(\mathbf{h}_{i \gets self}^{(l+1)}), \ \theta_{s} = \frac{\exp(\tilde{\theta}_{s})}{\sum_{c \in \{s,e,h\}} \exp(\tilde{\theta}_{c})}. 
    \end{aligned}
    \end{equation}
  
    \item \textit{MLP Fusion} concatenates all messages, followed by a fully connected neural network.
    
    \begin{equation}
    	f_{mlp} = \text{MLP}(\mathbf{h}_{i \gets self}^{(l+1)} \mathbin\Vert \mathbf{h}_{i \gets edge}^{(l+1)} \mathbin\Vert \mathbf{h}_{i \gets hyper}^{(l+1)}).
    \end{equation}
  
\end{itemize}
In this paper, we use $f_{mean}$ by default in HyperRoad and empirically compare the three fusions in Section \ref{sec:q2}.
\fi 

With the representations augmented by direct neighborhoods modeling, we further stack the aforementioned aggregation module to $L$ layers such that the model receptive fields are enlarged. Specifically, we name the node and hyperedge representations for $v_i$ and $e_j$ in the last layer as $\mathbf{h}_i$ and $\textbf{m}_j$, respectively. 

\subsection{Bi-level Self Supervised Learning} \label{sec:ssl}
Normally, a well-designed loss function plays a vital role in achieving desired properties for pre-training models in various domains \cite{devlin2018bert, lu2019vilbert}. 
We design two types of pretext tasks for RNRL, namely graph-level and hypergraph-level tasks, to effectively encode the road network structures into representations. {\bodyr {\LIANGMajor{The graph-level tasks refer to graph reconstruction, while the hypergraph-level tasks refer to hypergraph reconstruction and hyperedge classification. Among them, the graph reconstruction pretext task {\liangr has already been} applied in some existing RNRL models \cite{jepsen2020relational, wu2020learning} and the hypergraph reconstruction task is natural for hypergraph representation learning such as hypergraph auto-encoder \cite{wang2016structural, hu2021adaptive}. {\liangr We emphasize that} our self supervised learning component differs from existing methods in (1) it introduces the hypergraph reconstruction task in road network representation learning problem for the first time; and (2) it combines both graph-level tasks and hypergraph-level tasks in a unified framework for the first time; and (3) it proposes a new pretext task named hyperedge classification for the constructed hyperedges based on prior knowledge of road networks.}} }

\smallskip\noindent\textbf{Graph-level Pretext Task.} To make the learned road embeddings reserve the adjacent correlations between roads in $\mathcal{G_R}$, we design the first pretext task named \textit{graph reconstruction} to reconstruct adjacency matrix $\mathbf{A}$ based on corresponding embeddings. Specifically, we maximize the probability of connected road pairs {\CHENG{with a loss defined}} as follows.

\begin{equation}
\begin{aligned}
    \mathcal{L}_{GR} = - \sum_{v_i \in \mathcal{V}} \sum_{v_j \in \mathcal{N}_i} \left(\mathbf{h}_{i}^{T} \mathbf{h}_{j} - \sum_{n=1}^{N_G} 
    %\mathbb{E}_{v_n \sim P_{G}(v)} 
    \mathbf{h}_{i}^{T} \mathbf{h}_{n} \right), 
\end{aligned}
\end{equation}
where $\mathbf{h}_i$ is the learned embedding for road $v_i$, $N_G$ is the predefined negative size for the \textit{graph reconstruction} task, {\LIANGTwo and $v_n$ is the negative road sampled from outside $v_i$'s neighborhood {\CHENGTwo by following the distribution} $v_n \sim P_{G}(v_n|v_i)$ {\CHENGTwo with} $P_G(v_n|v_i) = \frac{1}{|\mathcal{V} \backslash \mathcal{N}_i|}$.}
% We will discuss the implementation of $P_{G}(v)$ later on.

\smallskip\noindent\textbf{Hypergraph-level Pretext Task.} Based on the proposed hypergraph structure, we design two self supervised learning tasks, namely \textit{hypergraph reconstruction} and \textit{hyperedge classification}, to inject prior knowledge and provide useful signals to guide road representation learning.

\begin{itemize}[leftmargin=*]
  \item \textit{Hypergraph Reconstruction.} Intuitively, the node and its hyperedge context should share similar representations. For example, the roads in a highway loop tend to serve similar functions. Thus, we propose to preserve this kind of correlations by reconstructing the incidence matrix $\mathbf{H}$ in $\mathcal{G_H}$. Similarly, the {\CHENG{loss}} function can be formulated as follows.
  
\begin{equation}
\begin{aligned}
    \mathcal{L}_{HR} = - \sum_{v_i \in \mathcal{V}} \sum_{e_j \in \mathcal{H}_i} \left(\mathbf{h}_{i}^{T} \mathbf{m}_{j} - \sum_{n=1}^{N_H} \mathbf{h}_{i}^{T} \mathbf{m}_{n} \right), 
\end{aligned}
\end{equation}
where $\mathcal{H}_i$ represents the neighbor hyperedge set of node $v_i$, $\mathbf{m}_j$ is the learned embedding for hyperedge $e_j$ in the last layer, and $N_H$ denotes the predefined negative size for the \textit{hypergraph reconstruction} task. 
For each positive pair $(v_i, e_j)$, {\LIANGTwo we fix road $v_i$ and sample other hyperedge $e_n \sim P_{H}(e_n|v_i)$ from outside $v_i$'s neighborhood in the hypergraph as negative pairs {\CHENGTwo with}
$P_H(e_n|v_i) = \frac{1}{|\mathcal{E_H} \backslash \mathcal{H}_i|}$.}
% {\CHENG{We will discuss the implementation of $P_{H}(e)$ later on.}}

% \begin{figure}[t!]
% 	\centering
% 	\includegraphics[width=0.50\textwidth]{inputs/Figs/Long_Range_Case.pdf}
% 	\caption{Road correlations for validating long-range relationships.}
% 	\label{fig:road_long}
% 	\vspace{-0.1cm}
% \end{figure}
  
\item \textit{Hyperedge Classification.} \label{sec:HC}
{\LIANG{As discussed in Section \ref{sec:intro}, roads far apart may have long-range relationships. 
{\LIANGTwo Specifically, such relationships can exist between roads involved in similar (or same) hyperedges {\bodyr {\LIANGMajor{(\emph{e.g.}, residential regions with similar urban functions)}}}.
% where hyperedges serve as ``bridges''.
For example, in Fig. \ref{fig:intro}, the hyperedges indicated by polygon regions of residential buildings shown as blue dotted frames involve few residential roads,
% (i.e., it is small), 
while the hyperedge indicated by the highway loop involves many highway roads.}
{\bodyr{\LIANGMajor{Therefore, we cluster the hyperedges into $K$ clusters 
% {\LIANGTwo as we did in data analysis,}
via $K$-means clustering, 
where hyperedges with similar sizes are in the same cluster.}}
% each cluster has similar hyperedges.
% The hyperedge clusters act as a medium to share information through a ``road-hyperedge-cluster-hyperedge-road'' scheme, which helps to capture the long-range relationships among roads to some extent. 
{\chengr The hyperedges' labels act as a medium to share information among roads that may be far apart through a ``road-hyperedge-cluster-hyperedge-road'' scheme for capturing long-range relationships.}
{\LIANGMajor{For example, ``Road B'' and ``Road C'' in Fig. \ref{fig:intro} are involved in hyperedges with similar sizes ({\chengr which} tend to have similar urban functions and belong to the same hyperedge cluster), and thus they would tend to have similar representations.}} }

Specifically, we assign each hyperedge $e_i \in \mathcal{E_H}$ a cluster label $c_i$. 
%
% {\color{red}Comments: Previously, I remember the area of the corresponding polygon for a hyperedge is also used for clustering. Now, this information is simply omitted?}
% 
% {\LIANG{In fact, I tried both area and number of sides. Using one of them can produce similar results. However, using both of them as features produce  results a little worse (maybe the feature scale problem when serving as a two-dimensional feature vectors in k-means). Currently, I simply use number of sided as features. So, I don't mention area here for avoiding possible misleading.}}
% 
% Given the label $c_i$ and its representation $\textbf{m}_i$, 
We then propose a new task named hyperedge classification, which predicts the cluster label $c_i$ for a given hyperedge $e_i$ based on its learned representation. 
Formally, we define a categorical cross-entropy loss function for the task as follows. 
\begin{equation}
\begin{aligned}
    \mathcal{L}_{HC} = - \sum_{e_i \in \mathcal{E_H}} \left(\sum_{k=1}^{K} \log \frac{\exp(x_{i,k})}{\sum_{k^\prime} \exp(x_{i,k^\prime})} \cdot y_{i,k} \right), 
\end{aligned}
\end{equation}
where $K$ is the number of clusters, $y_{i,k}$ is set to 1 when $k=c_i$ and 0 for other cases, $x_{i,k}$ is the $k$-th dimension of output vector $\mathbf{x}_i$ obtained by transforming hyperedge representation $\mathbf{m}_i$ as $\mathbf{x}_i = \text{MLP}(\mathbf{m}_i)$. 
}}
\end{itemize}
%

% {\color{red}Comments: The hyperedge prediction task should be revised further (its motivation, and the clustering process). }

% {\LIANG{Have changed.}}

Finally, we unify the different pretext tasks into a multi-task learning framework. The joint learning {\CHENG{loss}} function is as follows.

\begin{equation}
\label{eq:loss_base}
\mathcal{L} = \mathcal{L}_{GR} + \alpha \cdot (\mathcal{L}_{HR} + \mathcal{L}_{HC}).
\end{equation}
where $\alpha$ is hyper-parameter controlling the contributions from graph-level and hypergraph-level pre-training.

\section{HyperRoad: Extensions}
\label{sec:extension}
% In this section, we proceed to present how context information such as road attributes and trajectories can be integrated into HyperRoad to enable advanced models.
In this section, we extend the HyperRoad model to other problem settings with additional road attributes and/or additional trajectories.

\begin{figure}[t!]
	\centering
	\includegraphics[width=0.68\textwidth]{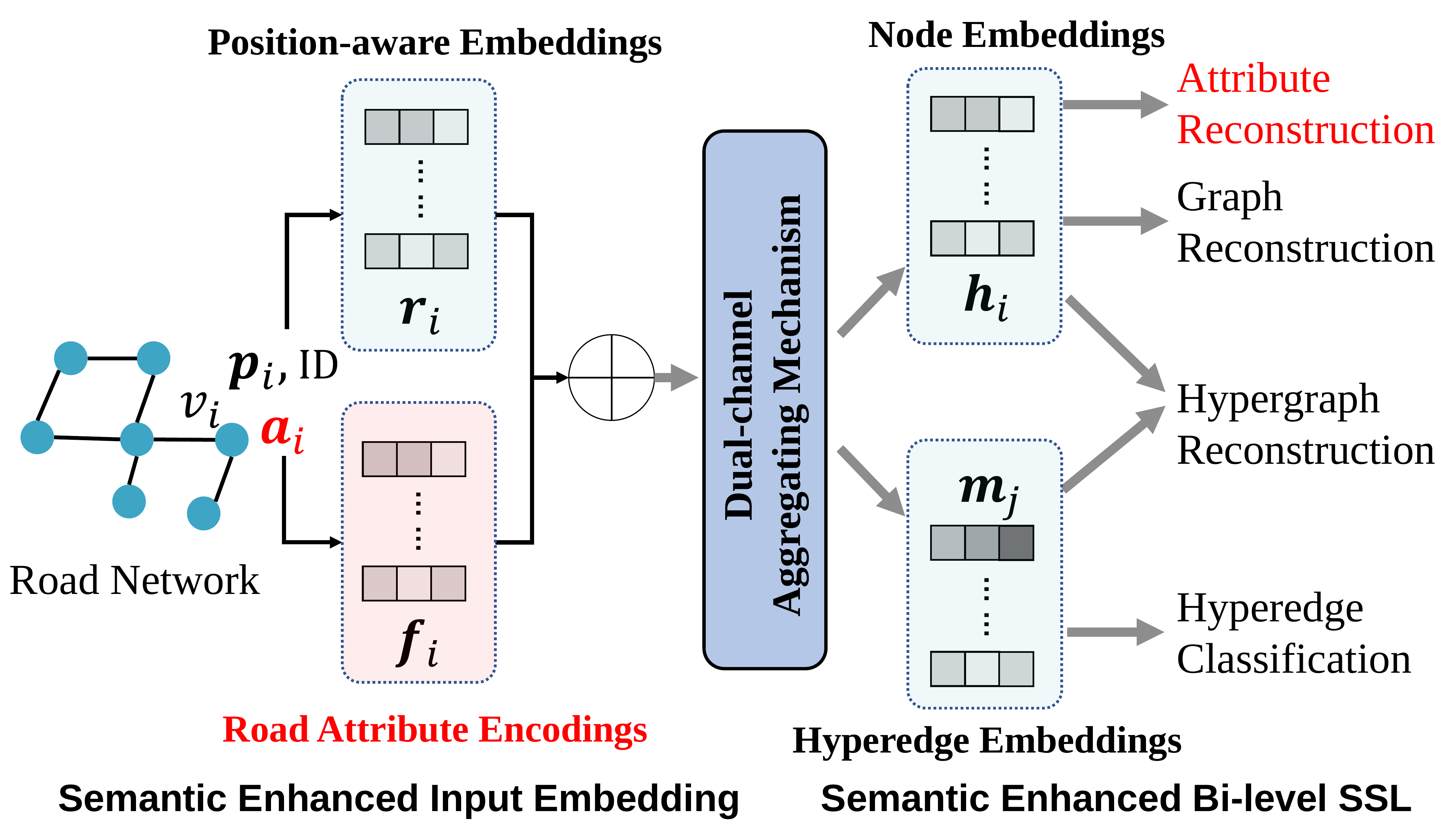}
	\caption{The architecture of HyperRoad-A.}
	\label{fig:extension-A}
\vspace{-0.1cm}
\end{figure}

% {\color{red}Comments on Figure 3 and 4: (1) Adjust the font size and figure size so that font sizes in the figures are comparable with those in the main texts and the figures can be read clearly; (2) For Fig. 3, specify the inputs including the road network (involving the spatial and structure information) and the attributes; (3) For Fig.4, change ``Trajectory Database'' to ``Trajectories'', specify the inputs including the road network and trajectories.}

% {\LIANG{Have changed.}}

\subsection{HyperRoad-A: with Additional Road Attributes}
When road attributes are available, 
% our model can not only capture the structural assumptions of road networks, but also incorporate semantic patterns in the representations. 
% To achieve this goal, 
we propose to extend HyperRoad by introducing road attributes into embedding layer and adding an auxiliary road attribute reconstruction task following the same encoder-decoder strategy as for modeling structural information. We name this advanced model as HyperRoad-A, and an overview of its framework is given in Fig. \ref{fig:extension-A} {\CHENG{(with the newly added components highlighted in red)}}. 

% Specifically, for a target road $v_i$, we retrieve its road attributes as $\mathbf{a}_i = \{a_{i,1}, a_{i,2}, ..., a_{i,m}\}$ where $m$ denotes the number of attributes. 
% In this paper, we use road tags as road attributes by following \cite{chen2021robust}. These tags are encoded by one-hot encoding model into a semantic feature vector $\mathbf{f}_i$. 

\smallskip\noindent\textbf{Semantic Enhanced Input Embedding.}
% \smallskip\noindent\textbf{Road Attribute Encoding.} 
In the encoder part, we model road attributes as another kind of semantic input information besides road ID and position. For a target road $v_i$, the new input embedding is defined as $\mathbf{r}_i \mathbin\Vert \mathbf{f}_i$, where $\mathbf{r}_i$ is the original input embedding of HyperRoad defined in Section \ref{sec:p_ie} and $\mathbf{f}_i$ is the embedding of $v_i$'s road attributes via one-hot encoding. 

\smallskip\noindent\textbf{Semantic Enhanced Self Supervised Learning.} 
To force the learned road representations preserve the semantic/attribute information, we add another auxiliary attribute reconstruction task in the self supervised learning module.
% for Equation \ref{eq:loss_base}. 
Specifically, given a road $v_i$ and its attributes $\{a_{i,1}, a_{i,2}, ..., a_{i,m}\}$, we aim at predicting these attributes based on the learned representation $\mathbf{h}_i$ with the additional loss defined as follows. 
\begin{equation}
\begin{aligned}
    \mathcal{L}_{AR} = - \sum_{v_i \in \mathcal{V}} \sum_{j=1}^{m} \left( \sum_{k=1}^{|A^j|} \log \frac{\exp(x^k_{i,j})}{\sum_{k^\prime} \exp(x^{k^\prime}_{i,j})} \cdot y^k_{i,j} \right), 
\end{aligned}
\end{equation}
where $m$ is the number of road attributes, $|A^j|$ is the number of possible values for the $j$-th attribute $a_{i,j}$, $x^{k}_{i,j}$ is the $k$-th dimension of an output vector $\mathbf{x}_{i,j}$ obtained by transforming the road representation $\mathbf{h}_i$ as $\mathbf{x}_{i,j} = \text{MLP}_{j}(\mathbf{h}_i)$, and $y^k_{i,j}$ equal to 1 when {\color{black}$a_{i,j}$ is equal to the $k^{th}$ attribute value} and 0 otherwise. 

{\LIANGTwo
Then, the joint learning loss function for HyperRoad-A can be defined as follows.
}
\begin{equation}
    \label{eq:loss_A}
    \mathcal{L} = \mathcal{L}_{GR} + \alpha \cdot (\mathcal{L}_{HR} + \mathcal{L}_{HC}) + \mathcal{L}_{AR}.
\end{equation}
%
% {\color{red}Comments: I suggest to change ``$\mathbf{x}^{i}_{j}$'' to ``$\mathbf{x}_{i,j}$'', ``$x^{i}_{j,k}$'' to ``$x^{k}_{i,j}$'', and ``$y^{i}_{j,k}$'' to ``$y^{k}_{i,j}$'' for better consistency with the notation $a_{i,j}$.}
 
% {\LIANG{Have changed.}}

\subsection{HyperRoad-T: with Additional Trajectories}
\label{sec:ext_tra}
\begin{figure}[t!]
	\centering
	\includegraphics[width=0.65\textwidth]{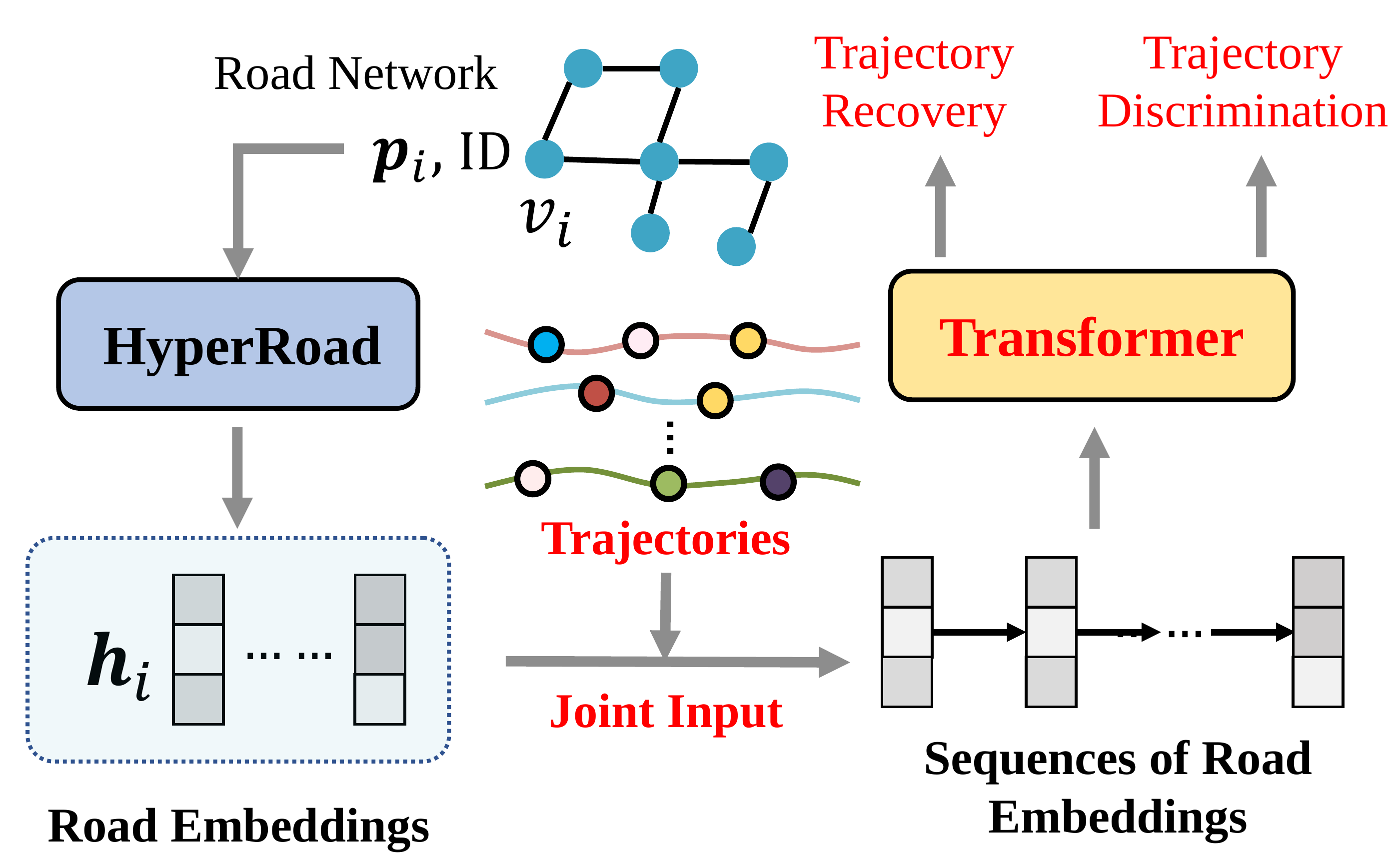}
	\caption{The architecture of HyperRoad-T.}
	\label{fig:extension-T}
\vspace{-0.1cm}
\end{figure}

Traveling patterns serve as a generic feature on road networks, which should be properly integrated so that they can precisely enrich the multi-faced knowledge about road networks such as underlying functional characteristics \cite{wu2020learning,chen2021robust}. 
% However, the trajectory information suffer from coverage problems and it is better to be encoded flexibly. 
In the section, we extend HyperRoad to the setting, where some trajectories on a road network are available.
We call the resulting model HyperRoad-T. The overview of the HyperRoad-T framework is shown in Fig. \ref{fig:extension-T}, where we follow a pre-training and refining strategy {\CHENG{(with the newly added components highlighted in red)}}. 

In the first stage, we train the road representations by optimizing the loss function defined in Equation \ref{eq:loss_base}. Then, these representations are treated as the input embeddings of a stacked bidirectional transformer. Following existing work \cite{chen2021robust}, we adopt two BERT-style self supervised learning tasks, i.e., trajectory recovery and trajectory discrimination to help trajectory modeling.

\smallskip\noindent\textbf{Trajectory Recovery.} Similar to the MLM (Masked Language Model) task used in BERT, we mask a sequence of consecutive roads to form a partially observed route given a trajectory. The transformer is then trained to predict these masked roads.

\smallskip\noindent\textbf{Trajectory Discrimination.} Similar to the NSP (Next Sentence Prediction) task used in BERT, we adopt a trajectory level task besides the token level recovery task. Specifically, we sample real trajectories from trajectory database and generate fake trajectories through random walks on road network. This task is to judge whether a given road sequence is a real human trajectory or not. 

\subsection{HyperRoad-AT: with Additional Attributes and Trajectories}
We further extend HyperRoad for the setting with additional road attributes and trajectories.
Specifically, it corresponds to the HyperRoad-T model, with the HyperRoad component replaced by HyperRoad-A.
% In this advanced model, we simply replace the component HyperRoad in Fig. \ref{fig:extension-T} with HyperRoad-A. 
That is, we first pre-train the attribute aware road representation by base model HyperRoad-A, and then finetune road embeddings through trajectory modeling following the same strategy in Section \ref{sec:ext_tra}. We call the resulting model HyperRoad-AT.

\section{Experimental Setup}
\label{sec:expsetup}

In order to evaluate the effectiveness of our proposed model HyperRoad and its extensions, we conduct extensive experiments to answer the following questions.

\textbf{RQ1}: Can HyperRoad outperform other models on the downstream tasks when only structural and spatial information of a road network is available (the default setting)?

\textbf{RQ2}: What roles do different components of HyperRoad play in improving the model performance?

\textbf{RQ3}: How would different hyper-parameters influence the performance of HyperRoad?

\textbf{RQ4}: What do the learned road embeddings represent? Can these embeddings capture the underlying characteristics of road networks?

\textbf{RQ5}: Can the extensions of HyperRoad outperform the state-of-the-art models when additional attributes and/or trajectories are available?

In this section, we first describe the experimental settings.

% {\color{red}Comments on Table I: (1) for numbers, it's better to use some format, e.g., for 33948, we write 33,948. Same for other numbers larger than 1,000 throughout the paper. (2) Change ``\# Road-Road'' to ``\# edges'' and present the corresponding row in the 2nd row; (3)}

% {\LIANG{Have changed.}}

\newcommand{\tabincell}[2]{\begin{tabular}{@{}#1@{}}#2\end{tabular}}  
\begin{table}[t!]
	\renewcommand\arraystretch{1.2}
	\centering
	\caption{Baseline setup under different settings.}
	\scriptsize
	
	\resizebox{0.90 \linewidth}{!}{
		\begin{threeparttable}
			\begin{tabular}{c|c|c}
				\toprule
				Setting & Baselines & Our Model  \\
				\midrule
				\tabincell{c}{Structural and spatial information} & \tabincell{c}{Node2Vec, LINE, SDNE, GAE, GraphSAGE \\ RFN, IRN2Vec, HRNR, Toast}  & HyperRoad \\  
				\midrule
				+ Road Attributes  & \tabincell{c}{RFN-A, IRN2Vec-A, HRNR-A, Toast-A} & HyperRoad-A  \\
				\midrule
				+ Trajectory & \tabincell{c}{HRNR-T, Toast-T} & HyperRoad-T \\ 
				\midrule
				+ Attributes \& Trajectory & \tabincell{c}{HRNR-AT, Toast-AT} & HyperRoad-AT \\ \bottomrule
			\end{tabular}
		\end{threeparttable}
	}
	\label{tab:baselines}
	\vspace{-0.1cm}
\end{table}

\subsection{Baseline models}
We compare our model with the following 9 baselines, as described as follows.

\begin{itemize}[leftmargin=*]
%   \item \textbf{Node2Vec \cite{grover2016node2vec}}: It adopts a biased random walk to generate node sequences and then learns node representations through skip-gram model. The walk length and the window size are set to be 20 and 5 respectively. The parameters $p$ and $q$ for biased random walk are tuned in \{0.25, 0.5, 1.0, 2.0\}.
  
%   \item \textbf{LINE \cite{tang2015line}}: It learns node representation by considering both first and second order node proximities. We use half dimensions to capture the first-order proximity, and the other half dimensions to capture the second-order information. 
  
%   \item \textbf{SDNE \cite{wang2016structural}}: It represents the network as an adjacency matrix, and obtains node embeddings through autoencoder. The encoder and decoder are implemented as a two-layer MLP with the number hidden dimensions varied from \{1024, 64\}.
  
%   \item \textbf{GAE \cite{kipf2016variational}}: It extends the idea of SDNE by using graph convolutional network (GCN) to learn node representations. The number of GCN layers is selected from \{2, 3, 4\}.
  
%   \item \textbf{GraphSAGE \cite{hamilton2017inductive}}: It is another GNN based unsupervised graph representation learning model. We use mean aggregator and tune the number of layers in \{2, 3, 4\}.
  \item \textbf{Node2Vec \cite{grover2016node2vec}, LINE \cite{tang2015line}, SDNE \cite{wang2016structural}, GAE \cite{kipf2016variational}, and GraphSAGE \cite{hamilton2017inductive}}: The former three are traditional graph representation learning models and the latter two are GNN-based graph representation learning models. They {\LIANGTwo are} compared under the default setting {\LIANGTwo since they are all general graph structure representation learning methods.}
  
  \item \textbf{RFN \cite{jepsen2020relational}}: RFN uses both primal and dual graphs to model road networks where roads are considered edges in the primal graph and nodes in the dual graph. The road representations are learned by fusing two views' information to a graph auto-encoder framework. 
%   %
%   {\color{red}Comments: better to include a bit more description of method (e.g., its major idea but not some terms like node-relational and edge-relational views - they cannot be followed.}
%   {\LIANG{Have changed.}}
%   The number of GNN layers is tuned the same with GraphSAGE. 
  %We name the version with pure spatial information as RFN and the version with road attributes as RFN-A. 
  
  We consider two versions, RFN and RFN-A. In RFN, we use the proposed position-aware embeddings as model input. 
  In RFN-A, we further incorporate the road attribute embedding as input. Note that it is non-trivial to extend the RFN model with additional inputs of trajectories.
  
  \item \textbf{IRN2Vec \cite{wang2019learning}}: IRN2Vec applies shortest path sampling and skip-gram model to predict the information of geo-locality, geo-shape, and road tags given road sequences. 
%   The parameters are set the same as that in Node2Vec. % We name the version with pure spatial information as IRN2Vec and the version with road attributes as IRN2Vec-A.
  We consider two versions, IRN2Vec and IRN2Vec-A.
  In IRN2Vec, we exclude the tag-based loss function, and in IRN2Vec-A, we follow the setting of the original paper~\cite{wang2019learning}. Note that it is non-trivial to extend the IRN2Vec model with additional inputs of trajectories.
  
  \item \textbf{HRNR \cite{wu2020learning}}: It learns road representation by preserving road network semantic and traffic information in a hierarchical manner. The attributes are used as a part of input embedding and trajectories are used as a extra loss function. %We name the version with pure spatial information as HRNR, the version with attribute as HRNR-A, the version with trajectory as HRNR-T, and the version with both information as HRNR-AT. 
  We consider four versions of HRNR, namely HRNR, HRNR-A, HRNR-T, and HRNR-AT.
  In HRNR, we use the position-aware embeddings as model input and exclude the trajectory-based loss function. In HRNR-A, we further incorporate the road attribute embedding as inputs. In HRNR-T, we add the trajectory-based loss function. Finally, in HRNR-AT, we combine HRNR-A and HRNR-T.
%   we  follow the setting of the original paper~\cite{wu2020learning}, which uses both attributes and trajectories.
  
  \item \textbf{Toast \cite{chen2021robust}}: It utilizes auxiliary traffic attributes to train a skip-gram model. Specifically, a trajectory-enhanced Transformer module is utilized to extract traveling semantics on trajectory data. %Again, we name the version with pure spatial information as Toast, the version with attribute as Toast-A, the version with trajectory as Toast-T, and the version with both information as Toast-AT in this paper.
  We consider four versions of Toast, namely Toast, Toast-A, Toast-T, and Toast-AT.
  In Toast, we exclude the attribute prediction task and trajectory-enhanced module (this would degrade to the DeepWalk model). In Toast-A, we add the auxiliary attributes in skip-gram model. In Toast-T, we add the trajectory-enhanced module. Finally, in Toast-AT, we combine Toast-A and Toast-T. 
  
\end{itemize}
A summary of baseline models under different settings is provided in Table \ref{tab:baselines}. 
{\color{black}For fair comparison, we incorporate the position-aware embeddings to those GNN based baseline models, including HRNR, GAE, GraphSAGE, and RFN ({\LIANGTwo these studies do not consider node position information by default}), so that they take the same inputs as our model HyperRoad. Among other baseline models, IRN2Vec uses spatial information as supervision signals and the rest models do not use the spatial information and it remains non-trivial to incorporate the spatial information to these models.}
{\bodyr {\LIANGMajor{Note that following the existing works \cite{jepsen2020relational, wang2019learning, wu2020learning}, we do not consider the specialized models such as \cite{pan2019urban, liang2019urbanfm, yao2019revisiting} which are designed for a specific task as baselines, because our goal is to evaluate the usefulness and generalization of the learned road representations. 
% These specialized methods are highly dependent on the task-specific model designs and usually take more factors as input and thus it is not a fair comparison.
These specialized methods are highly dependent on the task-specific model designs and usually take more {\chengr data} as input and 
{\chengr it would not be fair to compare our model with these methods.}
}}} 
Therefore we use the same task-specific components and make them as simple as possible for all the compared RNRL models.
\label{sec:R2_setting}

% {\color{red}Comments on the specifications of baseline models: (1) Specify the inputs for models; (2) Explain how the model can be modified to handle other settings of inputs; and (3) Comment on the settings the models cannot be adapted to.}

% {\LIANG{Have changed.}}

% {\color{red}Comments: (1) I have combined all graph representation learning methods with one bulletin points and drop their descriptions. This is because they are well-known models and it is fine we simply mention them. 
% (2) For the specifications of their parameters, we can move them to the parameter setting part. }

% {\LIANG{Have changed.}}

\subsection{Evaluation Tasks}
We consider two types of downstream tasks,
% for testing the effectiveness of learned road representations, including 
namely road based and trajectory based tasks, for evaluation. 
For each task, we employ a simple task-specific component (e.g., LSTM or Logistic Regression) with the learned road network representations as input. We exploit the same task-specific component for all compared models for fair comparison. The adopted tasks are described as follows.

\smallskip\noindent\textbf{Road based tasks.} Following the existing work \cite{yin2021multimodal}, we consider the following four important road attribute predictions as road based tasks: 1) one/two way classification, 2) lanes classification, 3) speed limit classification, and 4) road type classification. 
For each task, we choose the road attribute values as the labels, regard the learned road representations as input features, and use an one-vs-rest logistic regression classifier to make predictions. 

\smallskip\noindent\textbf{Trajectory based tasks.} Following the existing work \cite{wang2019learning}, we use travel time estimation task as the trajectory based task.
% to further verify the effectiveness and ubiquity of the road embeddings. 
Travel time estimation is a regression task, which predicts the travel time of a given moving path in road networks. We apply a simple Long Short-Term Memory (LSTM) layer followed by one fully-connected layer as the regression model, which takes the road representation as input and outputs the estimated travel time. 

\smallskip\noindent\textbf{Remarks.} When considering road attributes as extra information, we use one/two way, lanes and speed limit as road attributes, and only treat the road type classification as road based task to avoid data leakage. 

% \subsection{Evaluation Metrics}
% We adopt different evaluation metrics for the above two different types of tasks. 

% For road based tasks, Micro-F1, Macro-F1 and Weighted-F1 are used to evaluate the classification accuracy. Among them, the Micro-F1 computes a global average F1 score by counting the sums of the True Positive (TP), False Positive (FP), and False Negative (FN) of each class, while the Macro-F1 is computed by taking the arithmetic mean of all the per-class F1 scores. Finally, the Weighted-F1 score is obtained by taking the mean of all per-class F1 scores while considering each class’s support.

% For trajectory based tasks (travel time estimation), MAE (Mean Average Error) and RMSE (Root Mean Square Error) are used to measure the closeness between the predicted value and the real value. Among them, MAE measures the average absolute distance between predictions and ground-truths, and RMSE measures the average quadratic errors of all predictions.

\subsection{Datasets, Evaluation Metrics and Parameter Settings}

\noindent\textbf{Datasets.}
\label{sec:dataset}
% \subsection{Datasets}
We use the road networks of two cities, Singapore and Xi'an, in our experiments. 
The structural and spatial information and the road attributes are collected from OpenStreetMap.
% \footnote{https://www.openstreetmap.org/}. 
The trajectories are obtained from public datasets released by Grab 
% \footnote{https://www.didiglobal.com} 
and DiDi Chuxing
% \footnote{https://www.grab.com/sg/} \cite{huang2019grab}. 
for Singapore and Xi'an, respectively.
In trajectories, the raw GPS points are transformed by map matching \cite{yang2018fast} into road sequences. 
% We then apply map segmentation algorithm \cite{yuan2012discovering} on each road network to construct hypergraphs as detailed in Section \ref{sec:hypercon}. 
% In this paper, we use road tags as road attributes by following \cite{chen2021robust}.
The statistics of these datasets are shown in Table \ref{tab:data}.

\smallskip\noindent\textbf{Evaluation Metrics.}
For road based tasks, we adopt Micro-F1, Macro-F1 and Weighted-F1 to evaluate the classification accuracy, and for the trajectory based task (travel time estimation), we adopt MAE (Mean Average Error) and RMSE (Root Mean Square Error) to evaluate the regression accuracy.

{\bodyr{\LIANGMajor{\smallskip\noindent\textbf{Remarks.} For all the mentioned tasks, we apply 5-fold cross validation to evaluate the performance of all the methods. For example, for the road based tasks, we use 4-fold road representations and their tag labels as training data and {\liangr use} the rest 1-fold data as test set (we also randomly sampled 10\% of each training fold to act as a validation set during training process). {\liangr We report the average of} the results of 5-fold test sets.}} }

\smallskip\noindent\textbf{Parameter Settings.}
In our experiments, we set the embedding dimension to be 64 for all the models for fair comparison. For random walk based methods such as Node2Vec, IRN2Vec and Toast, the walk length and the window size are set to be 20 and 5 respectively. The number of walks per node are searched in \{30, 50, 80\}. The parameters $p$ and $q$ for Node2Vec are tuned in \{0.25, 0.5, 1.0, 2.0\}. We use half dimensions for first-order proximity and another half for second-order proximity in LINE. The encoder and decoder are implemented as a two-layer MLP with hidden dimensions of \{1024, 64\} for SDNE. The aggregation layers are selected from \{2, 3, 4\} for all the GNN-based methods. For HyperRoad, We use a mini-batch Adaptive Moment Estimation (Adam)~\cite{kingma2015adam} optimizer. The learning rate and batch size are set to be 0.001 and 1,024 respectively. The scale parameter $\phi$ and frequency parameter $\lambda$ in Equation \ref{eq:pe} are set to be 10 and 1,000 separately. The weighted parameter $\alpha$ in the loss function is set to be 0.1 for both datasets. The negative sizes for graph sampling and hypergraph sampling named $N_G$ and $N_H$ are set to be 5 and 2. The number of hyperedge clusters are set to be 8 and 7 for Singapore dataset and Xi'an dataset, respectively. We also tune the number of layers for the dual-channel aggregating mechanism in the set of \{2,3,4\}. Further, in HyperRoad-A, the weight for loss $\mathcal{L}_{AR}$ is set to be 1.0. In HyperRoad-T and HyperRoad-AT, 2 multi-head self-attention layers are stacked and the head number is set as 4.

{\CHENG{The codes and datasets can be found via this link \footnote{https://drive.google.com/drive/folders/1oz19-F0igSziViKU4qmhDH-Do1Oudrh\_?usp=sharing}. We implement HyperRoad using PyTorch \cite{paszke2019pytorch}. Our software environment contains Ubuntu 18.04.3, PyTorch 1.2.0 and python 3.7.3. All of the experiments are conducted on a machine with 2 GPUs (Tesla V100-SXM2 * 2), 20 CPUs (Inter i7 6850K) and 500G memories.}}

\begin{table}[t!]
	\renewcommand\arraystretch{1.05}
	\centering
	\caption{Statistics of the datasets.}
	\scriptsize
    
	\resizebox{0.45\linewidth}{!}{
		\begin{threeparttable}
			\begin{tabular}{c|c|c}
				\toprule[0.38pt]
				Dataset & Singapore &  Xi'an  \\
				\midrule[0.30pt]
				\# Roads & 33,948 & 53,067 \\  
				\# Edges & 147,550 & 249,848 \\ 
				\# Hyperedges & 3,772 & 6,953 \\
				% \# Road-Hyperedge & 54,031 & 86,301 \\ 
				\# Trajectories & 28,000 & 50,000 \\ 
				\# One/Two Way & 2 & 2 \\
				\# No. of Lanes & 6 & 6 \\
				\# Speed Limit & 6 & 9 \\
				\# Road Type & 5 & 5 \\
				\bottomrule[0.38pt]
			\end{tabular}
		\end{threeparttable}}
	\label{tab:data}
\vspace{-0.1cm}
\end{table}
\section{Experimental Results}
\label{sec:expresult}

% In this section, we first conduct comprehensive comparison with the state-of-the-art road network representation learning models in 5 downstream applications. To justify the module designs in HyperRoad and demonstrate the effectiveness, we then perform ablation studies in Section \ref{sec:q2}. Then, we give discussions on parameter analysis and present a case study on \textit{Singapore} dataset to analyze what the learned road embeddings represent in Section \ref{sec:q3} and \ref{sec:q4}. Note that these experiments are conducted when pure spatial information is considered, which is the most generic setting in real world and is applicable in any urban area. Finally, we demonstrate the effectiveness of HyperRoad extensions under different settings in Section \ref{sec:q5} to show our model's flexibility.  

\begin{table*}[t!]
	\renewcommand{\arraystretch}{1.05} 
	\centering 
	\caption{Results for road based applications.}
    \normalsize
	\resizebox{0.99\linewidth}{!}{
		\begin{tabular}{c|ccc|ccc|ccc|ccc}
			\toprule
			\multirow{4}{*}{Model} & \multicolumn{6}{c|}{{\textbf{ One/Two Way}}} & \multicolumn{6}{c}{{\textbf{No. of Lanes}}}  \\
			\cmidrule{2-13}
			& \multicolumn{3}{c|}{{\it Singapore}} & \multicolumn{3}{c|}{{\it Xi'an}} & \multicolumn{3}{c|}{{\it Singapore}} & \multicolumn{3}{c}{{\it Xi'an}} \\
			\cmidrule{2-13}
			& Micro-F1 & Macro-F1 & Weighted-F1 & Micro-F1 & Macro-F1 & Weighted-F1 & Micro-F1 & Macro-F1 & Weighted-F1 & Micro-F1 & Macro-F1 & Weighted-F1  \\ \midrule
			Node2Vec & 0.7213 & 0.4221 & 0.6059 & 0.5858 & 0.5638 & 0.5743 & 0.4773 & 0.1077 & 0.3087 & 0.5600 & 0.2675 & 0.4789\\
			LINE & \underline{0.7652} & 0.6018 & 0.7117 & 0.6032 & 0.5909 & 0.5985 & 0.4772 & 0.1094 & 0.3101 & 0.5504 & 0.1900 & 0.4070\\
			SDNE & 0.7199 & 0.4226 & 0.6058 & 0.5686 & 0.5114 & 0.5295 & 0.4778 & 0.1117 & 0.3120 & 0.5504 & 0.1833 & 0.3998\\
			\midrule
		    GAE & 0.7135 & 0.4935 & 0.6342 & \underline{0.6676} & \underline{0.6650} & \underline{0.6677} & 0.4790 & 0.1248 & 0.3186 & \underline{0.5827} & \underline{0.3383} & \underline{0.5235}\\
		    GraphSAGE & 0.7170 & 0.4829 & 0.6296 & 0.6452 & 0.6398 & 0.6438 & 0.4793 & 0.1228 & 0.3170 & 0.5794 & 0.2990 & 0.5181\\
		    \midrule
		    RFN & 0.7410 & 0.5995 & 0.6990 & 0.6305 & 0.6213 & 0.6267 & 0.4784 & 0.1177 & 0.3154 & 0.5596 & 0.2757 & 0.4737\\
		    IRN2Vec & 0.7085 & 0.4538 & 0.6110 & 0.6274 & 0.6100 & 0.6175 & \underline{0.4793} & \underline{0.1271} & \underline{0.3190} & 0.5677 & 0.2176 & 0.4508\\ 
		    HRNR & 0.7226 & \underline{0.6379} & \underline{0.7145} & 0.6659 & 0.6606 & 0.6652 & 0.4767 & 0.1076 & 0.3078 & 0.5601 & 0.2957 & 0.4547\\
		    Toast & 0.7313 & 0.5027 & 0.6517 & 0.5888 & 0.5647 & 0.5758 & 0.4766 & 0.1082 & 0.3089 & 0.5498 & 0.2009 & 0.4209 \\
		    \midrule
			HyperRoad & \textbf{0.7864} & \textbf{0.6995} & \textbf{0.7676} & \textbf{0.7051} & \textbf{0.7033} & \textbf{0.7054} & \textbf{0.4829} & \textbf{0.1534} & \textbf{0.3392} & \textbf{0.5955} & \textbf{0.3670} & \textbf{0.5459}\\
			Improvement & 2.77\% & 9.66\% & 7.43\% & 5.62\% & 5.76\% & 5.66\% & 0.75\% & 20.69\% & 6.33\% & 2.20\% & 8.48\% & 4.28\% \\
			\bottomrule
			\toprule
			\multirow{4}{*}{Model} & \multicolumn{6}{c|}{{\textbf{Speed Limit}}} & \multicolumn{6}{c}{{\textbf{Road Type}}}  \\
			\cmidrule{2-13}
			& \multicolumn{3}{c|}{{\it Singapore}} & \multicolumn{3}{c|}{{\it Xi'an}} & \multicolumn{3}{c|}{{\it Singapore}} & \multicolumn{3}{c}{{\it Xi'an}} \\
			\cmidrule{2-13}
			& Micro-F1 & Macro-F1 & Weighted-F1 & Micro-F1 & Macro-F1 & Weighted-F1 & Micro-F1 & Macro-F1 & Weighted-F1 & Micro-F1 & Macro-F1 & Weighted-F1 \\ \midrule
			Node2Vec & 0.5782 & 0.1273 & 0.4252 & 0.7661 & 0.7620 & 0.7679 & 0.4984 & 0.2733 & 0.4590 & 0.3791 & 0.2680 & 0.3298\\
			LINE & 0.5781 & 0.1221 & 0.4235 & 0.5920 & 0.6669 & 0.5917 & 0.5059 & 0.2715 & 0.4638 & 0.3892 & 0.2674 & 0.3358\\
			SDNE & 0.5766 & 0.1235 & 0.4243 & 0.4825 & 0.4288 & 0.4546 & 0.4806 & 0.2594 & 0.4422 & 0.3653 & 0.2270 & 0.2940\\
			\midrule
			GAE & \underline{0.5944} & \underline{0.1980} & \underline{0.4654} & 0.7661 & 0.8393 & 0.7636 & 0.5017 & \underline{0.3462} & 0.4762 & 0.4321 & \underline{0.3744} & \underline{0.4097}\\
			GraphSAGE & 0.5910 & 0.1864 & 0.4558 & \underline{0.7711} & \underline{0.8500} & \underline{0.7700} & 0.4955 & 0.3238 & 0.4648 & 0.4312 & 0.3715 & 0.4086\\
			\midrule
			RFN & 0.5904 & 0.1579 & 0.4471 & 0.6368	& 0.6711 & 0.6321 & \underline{0.5260} & 0.3207 & \underline{0.4877} & 0.4031 & 0.2937 & 0.3547\\
			IRN2Vec & 0.5869 & 0.1479 & 0.4341 & 0.6119 & 0.5444 & 0.5962 & 0.4773 & 0.2567 & 0.4353 & 0.3836 & 0.2924 & 0.3475 \\
			HRNR & 0.5868 & 0.1232 & 0.4341 & 0.6915 & 0.6052 & 0.6852 & 0.5050 & 0.2585 & 0.4399 & \underline{0.4485} & 0.3155 & 0.3914\\
			Toast & 0.5784 & 0.1231 & 0.4245 & 0.5820 & 0.5727 & 0.5814 & 0.4987 & 0.2697 & 0.4581 & 0.3697 & 0.2491 & 0.3139\\
			\midrule
			HyperRoad & \textbf{0.6066} & \textbf{0.2534} & \textbf{0.4970} & \textbf{0.7861} & \textbf{0.8566} & \textbf{0.7832} & \textbf{0.5785} & \textbf{0.3964} & \textbf{0.5482} & \textbf{0.4562} & \textbf{0.3957} & \textbf{0.4359}\\
			Improvement & 2.05\% & 27.98\% & 6.79\% & 1.95\% & 0.78\% & 1.71\% & 9.98\% & 14.50\% & 12.41\% & 1.72\% & 5.69\% & 6.39\% \\
			\bottomrule
	\end{tabular}}
	\label{tab:classify}
\vspace{-0.1cm}
\end{table*}

\subsection{Performance Comparison (RQ1)} \label{sec:q1}
We first evaluate the learned embeddings in the road based tasks. We show the performances of different models in Table~\ref{tab:classify}, where the best performance is in boldface and the second best is underlined. From the results, we have the following observations. First, the performances of GNN-based models are generally much better than traditional graph representation learning models, showing the advantages of deep neural networks in road network structure modeling over shallow representations. Second, the performances of road network representation learning models are the best in many cases among the baselines. For example, HRNR tends to perform better on one/two way prediction while RFN is better on road type classification task. However, the simple graph reconstruction objective of RFN and HRNR hinder their performance and superiority on other downstream applications. Third, our model performs consistently better than all the baselines in terms of all the metrics on the four tasks. {\LIANGTwo And the average margin in Macro-F1 reaches 18.20\% and 5.18\% for \textit{Singapore} and \textit{Xi'an} dataset respectively.} It shows that HyperRoad can better capture the underlying characteristics of road networks with help of the newly neighborhood aggregation mechanism and auxiliary hypergraph-level pre-training task. 

Next, we discuss the model performance on travel time estimation task in Table \ref{tab:tte}. It can be observed that HyperRoad consistently outperforms all competitors on both datasets. The result demonstrates that the road embeddings learned by HyperRoad can also be applicable for traffic level tasks. A possible reason is that the travel time is highly related to road information such as road type and road speed. HyperRoad captures the intrinsic properties of the road networks and thus improve the task performance. It verifies  the effectiveness of the learned road representations from another perspective.

\begin{table}[t!]
    \renewcommand{\arraystretch}{1.03} 
	\centering
	\tiny
	\caption{Results for travel time estimation.}
	\label{tab:tte}
	\resizebox{0.46 \linewidth}{!}{
		\begin{tabular}{@{}l|cc|cc@{}}
			\toprule
			\multirow{3}{*}{Model} & \multicolumn{2}{c|}{{\it Singapore}} & \multicolumn{2}{c}{{\it Xi'an}} \\
			\cmidrule{2-5}
			& MAE & RMSE & MAE  & RMSE  \\ \midrule
		    Node2Vec & 152.44 & 219.29 & 272.54 & 390.06 \\
		    LINE & 153.84 & 220.88 & 276.71 & 396.77 \\ 
		    SDNE & 162.43 & 231.80 & 277.62 & 397.26 \\
		    \midrule
		    GAE & 152.07 & 225.33 & \underline{271.43} & \underline{388.73} \\
		    GraphSAGE & 151.83 & 219.88 & 272.22 & 390.90 \\ 
		    \midrule
		    RFN & 154.76 & 219.70 & 273.35 & 395.48 \\ 
		    IRN2Vec & \underline{151.45} & \underline{218.06} & 281.89 & 398.86 \\
		    HRNR & 168.35 & 244.09 & 275.62 & 397.22 \\ 
		    Toast & 158.72 & 230.61 & 272.12 & 395.05  \\
		    \midrule
		    HyperRoad & \textbf{145.79} & \textbf{214.39} & \textbf{266.50} & \textbf{382.25} \\ 
		    Improvement & 3.73\% & 1.68\% & 1.82\% & 1.67\% \\
			\bottomrule
	\end{tabular}}
\vspace{-0.1cm}
\end{table}

\begin{table*}[t!]
	\renewcommand{\arraystretch}{1.15} 
	\centering 
	\caption{Results for ablation studies on Singapore dataset.}
    \normalsize
	\resizebox{0.99\linewidth}{!}{
		\begin{tabular}{c|ccc|ccc|ccc|ccc}
			\toprule
			\multirow{3}{*}{Model} & \multicolumn{3}{c|}{{\textbf{ One/Two Way}}} & \multicolumn{3}{c|}{{\textbf{No. of Lanes}}} & \multicolumn{3}{c|}{{\textbf{Speed Limit}}} & \multicolumn{3}{c}{{\textbf{Road Type}}} \\
			\cmidrule{2-13}
			& Micro-F1 & Macro-F1 & Weighted-F1 & Micro-F1 & Macro-F1 & Weighted-F1 & Micro-F1 & Macro-F1 & Weighted-F1 & Micro-F1 & Macro-F1 & Weighted-F1  \\ \midrule
			w/o PE & 0.7690 & 0.6619 & 0.7421 & 0.4762 & 0.1092 & 0.3091 & 0.5880 & 0.1324 & 0.4420 & \underline{0.5756} & 0.3193 & 0.5301\\
			\midrule
			w/o DAM & 0.7545 & 0.6221 & 0.7164 & 0.4784 & 0.1183 & 0.3145 & 0.5929 & 0.1928 & 0.4544 & 0.5497 & 0.3609 & 0.5114 \\ 
            {\bodyr{AM-KGNN}} & {\bodyr{0.7832}} & {\bodyr{0.6903}} & {\bodyr{0.7618}} & {\bodyr{0.4768}} & {\bodyr{0.1174}} & {\bodyr{0.3148}} & {\bodyr{0.5897}} & {\bodyr{0.1555}} & {\bodyr{0.4447}} & {\bodyr{0.5587}} & {\bodyr{0.3022}} & {\bodyr{0.5117}} \\
			DAM-ATT & 0.7434 & 0.6079 & 0.7051 & 0.4807 & 0.1339 & 0.3257 & 0.5888 & 0.2001 & 0.4600 & 0.5496 & \underline{0.3789} & 0.5203\\
		    DAM-MLP & 0.7773 & 0.6802 & 0.7545 & 0.4808 & 0.1395 & 0.3295 & 0.5945 & 0.1997 & 0.4653 & 0.5667 & 0.3350 & 0.5255\\
		    \midrule
			w/o GPT  & 0.7804 & 0.6875 & 0.7593 & 0.4793 & 0.1403 & 0.3293 & 0.5876 & 0.1952 & 0.4568 & 0.5742 & 0.3466 & 0.5316\\
			w/o HPT & 0.7402 & 0.5965 & 0.6980 & \underline{0.4827} & 0.1465 & 0.3327 & \underline{0.6016} & \underline{0.2341} & \underline{0.4885} & 0.5220 & 0.3598 & 0.4937\\
            {\bodyr{w/o HEC}} & {\bodyr{0.7239}} & {\bodyr{0.5442}} & {\bodyr{0.6648}} & {\bodyr{0.4794}} & {\bodyr{0.1287}} & {\bodyr{0.3243}} & {\bodyr{0.5945}} & {\bodyr{0.1602}} & {\bodyr{0.4572}} & {\bodyr{0.5553}} & {\bodyr{0.3679}} & {\bodyr{0.5217}} \\
		    HyperRoad-DBS  & \textbf{0.7879} & \textbf{0.7017} & \textbf{0.7693} & 0.4796 & \underline{0.1497} & \underline{0.3371} & 0.5962 & 0.2277 & 0.4838 & 0.5724 & 0.3674 & \underline{0.5349}\\
		    \midrule
			HyperRoad & \underline{0.7864} & \underline{0.6995} & \underline{0.7676} & \textbf{0.4829} & \textbf{0.1534} & \textbf{0.3392} & \textbf{0.6066} & \textbf{0.2534} & \textbf{0.4970} & \textbf{0.5785} & \textbf{0.3964} & \textbf{0.5482}\\ 
			\bottomrule
	\end{tabular}}
	\label{tab:ablation}
\vspace{-0.1cm}
\end{table*}

% {\color{red}Comments on Table VI: The font size is too small. Some margins can be removed for saving some space.}

% {\LIANG{Have changed.}}

\subsection{Ablation Study (RQ2)}  \label{sec:q2}
In this section, we perform ablation studies on {\it Singapore} dataset to inspect how each module of HyperRoad affects the model performance in four downtream tasks. 

There are three core modules in HyperRoad: positional encoding, dual-channel aggregation mechanism, self-supervised learning with both graph-level and hypergraph-level pretext tasks. By removing or replacing these modules, we can obtain different variants of HyperRoad. Specifically, we firstly remove the positional encoding module in the model variant w/o PE. Then, we remove the dual-channel aggregation mechanism and obtain the model variant w/o DAM, in which we only use the hyperedge channel aggregation to update the road representations. Note that we do not consider the case of using edge channel aggregation only, since it can not update and output the hyperedge representations necessary for subsequent hypergraph-level objectives. {\bodyr {\LIANGMajor{Considering the effectiveness of the newly proposed high-order graph neural networks \cite{morris2019weisfeiler} in capturing high-order node relations, we replace the dual-channel aggregation mechanism with k-GNN in \cite{morris2019weisfeiler} and name this model variant as AM-KGNN.}}}
In addition, to better validate the design of dual-channel aggregation, we use attention and MLP aggregators to implement $f(\cdot)$ in channel fusion and obtain two models named DAM-ATT and DAM-MLP respectively. Finally, models w/o GPT and w/o HPT can be obtained by removing the corresponding graph-level pretext task and hypergraph-level pretext task. {\bodyr {\LIANGMajor{To verify the importance of the newly proposed hyperedge classification task in road network, we obtain another model variant w/o HEC where only the hyperedge classification task is removed.}}} {\LIANGTwo To investigate the impacts of negative sampling methods, we replace the random sampling process with a \emph{distance based sampling} (DBS) process and obtain a model variant called HyperRoad-DBS. Here, in the distance based sampling process, the sampling probability of a candidate road $v_n$ in $P_G(v_n|v_i)$ 
% (or a candidate hyperedge $e_n$ in $P_H(e_n|v_i)$) 
is defined as the normalization of its shortest path distance from the anchor road $v_i$. The sampling probability for a candidate hyperedge $e_n$ in $P_H(e_n|v_i)$ is defined using the same strategy, where the distance between $e_n$ and $v_i$ is the average of shortest path distances between roads in $e_n$ and $v_i$.}
% We name the model variant using shortest path distance based sampling as HyperRoad-DBS. 
The ablation results are shown in Table \ref{tab:ablation}.

\smallskip\noindent\textbf{Contribution of the positional encoding.} By comparing % the performance of simplified model 
w/o PE and HyperRoad, we can find that the positional encoding module indeed brings substantive improvement in downstream task performances. {\LIANGTwo For example, HyperRoad outperforms the base model w/o PE by 40.42\% in Macro-F1 on average.}
The results justify the fact that the geo-location information is the core property of road network, and it can hardly be captured only by the graph structures. 
% Similar findings can be witnessed in Section \ref{sec:q1} by comparing IRN2Vec and its counterpart Node2Vec. 

\smallskip\noindent\textbf{Contribution of the dual-channel aggregation mechanism.} {\LIANGTwo The main difference between HyperRoad and w/o DAM is whether to aggregate information from two channels named edge channel and hyperedge channel at the same time.} 
As seen from Table \ref{tab:ablation}, HyperRoad brings at least 9.83\% improvement in terms of Macro-F1 compared with w/o DAM. This is because the two channels capture different graph structure information. Using only one of them will cause information loss and lead to inferior model performance. {\bodyr {\LIANGMajor{
% Similar {\chengr findings} can be observed by comparing HyperRoad with AM-KGNN. 
{\chengr In addition,} although AM-KGNN aggregates the information from high-order neighbors defined by the $k$-order cliques, these different structures are still defined on the same simple graph.
% , and thus only capturing one channel information. 
The results demonstrate the importance of our proposed dual channel aggregation mechanism.
% and it cannot be simply replaced by existing GNN components.
}}}
As for different channel fusion operations, 
% the MLP aggregation performs better on one/two way prediction task while attention aggregation performs better on road type classification task. However, it is interesting to find that 
the simple mean aggregation (adopted in HyperRoad) can achieve the best results.
% across all evaluation metrics. 
A possible reason might be that the extra MLP or attention layer introduces too much non-linearity for road representation learning, which is hard to train.
% and harmful to downstream applications.

\smallskip\noindent\textbf{Contribution of the self-supervised pretext tasks.} {\bodyr {\LIANGMajor{By comparing the performance of w/o GPT, w/o HPT, w/o HEC, HyperRoad-DBS and HyperRoad, we have the following observations. }}} 1) HyperRoad which considers both graph-level and hypergraph-level tasks generally has better performance than the models considering only single aspect, i.e., w/o GPT and w/o HPT. 
% For example, the full model HyperRoad brings 10.17\% and 14.37\% improvement (in terms of Macro-F1) compared with the base models w/o HPT and w/o GPT on road type classification task. 
The result demonstrates that both level tasks are important for effective road representations. 
% In fact, this conclusion is quit intuitive. For example, it is reasonable that the roads in a hyperedge (motorways around a lake) might have the same road type (motorways) but slightly different driving speed. Preserving only the hypergraph structure without graph homogeneity constraints may bring noise and lead to biased road representations. 
2) % By comparing w/o GPT with w/o HPT,  We can also find that 
The model w/o GPT performs better on one/two way classifications while w/o HPT is better on speed limit predictions. From another perspective, it verifies that each level task has its own merit to enhance the road representations. 
{\bodyr {\LIANGMajor{3) Model w/o HEC simply combines two commonly used reconstruction tasks, {\chengr namely graph reconstruction and hypergraph reconstruction}. Compared with w/o HPT which only applies {\chengr the} graph reconstruction task, w/o HEC performs better {\chengr for} Road Type prediction {\chengr but} performs worse {\chengr for} other applications. {\chengr The} results verify the advantages of introducing {\chengr the} hypergraph reconstruction task {\chengr to RNRL}. {\chengr It also} demonstrates that simply combining two commonly used reconstruction tasks {\chengr may} introduce noise for model learning. In contrast, HyperRoad achieves the best performance {\chengr for} all applications, showing the {\chengr benefit} of the newly proposed hyperedge classification task.}}}
4) {\LIANGTwo As for different sampling methods, 
% it is interesting to find that 
the random sampling model (adopted in HyperRoad) achieves 
% comparable and even 
better performances than the distance based sampling method. A possible reason is that roads far away on the map may still serve as similar urban functions and it is unreasonable to treat them as negative samples more frequently as HyperRoad-DBS does. This results also verify the importance of long-range relationship in road network from another perspective.}

% {\color{red}Comments: The description of the ablation study results can be shortened. Some summarization of the results would be good enough.}

% {\LIANG{Have deleted a few sentences.}}

\if 0 

\begin{figure}[t]
	\centering
	\begin{subfigure}[b]{0.50\textwidth}
		\includegraphics[width=\textwidth]{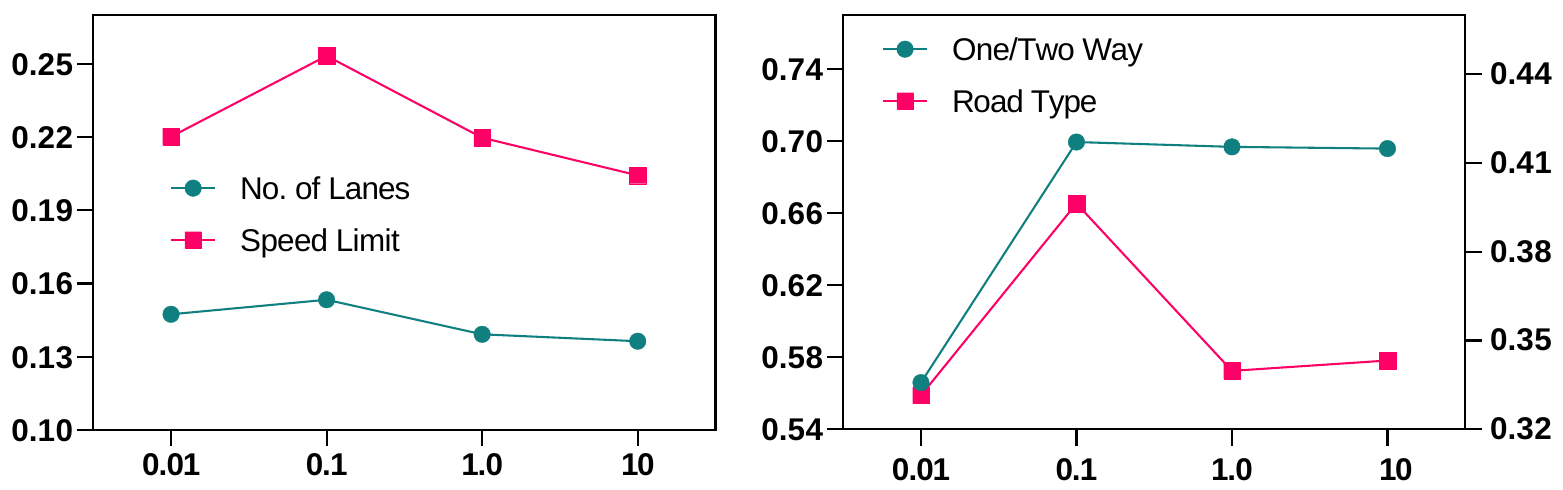}
		\caption{Parameter $\alpha$ (\textit{Singapore})}
		\label{fig:se_alpha}
	\end{subfigure}
	
	\begin{subfigure}[b]{0.50\textwidth}
		\includegraphics[width=\textwidth]{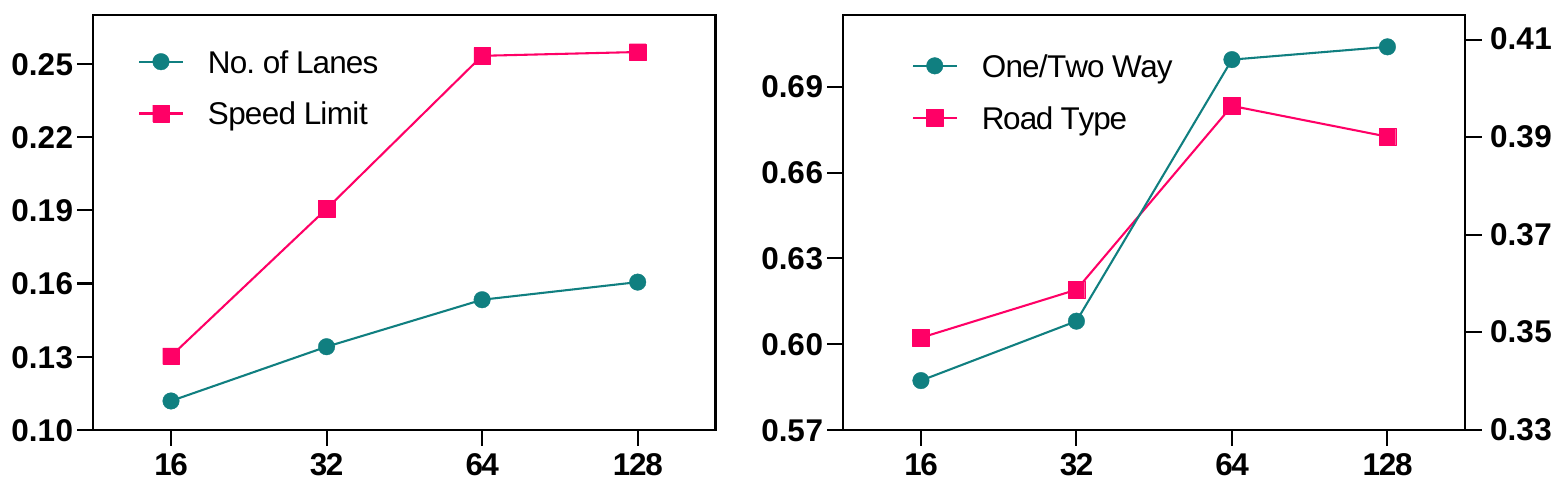}
		\caption{Parameter $d$ (\textit{Singapore})}
		\label{fig:se_d}
	\end{subfigure}
	
	\begin{subfigure}[b]{0.50\textwidth}
		\includegraphics[width=\textwidth]{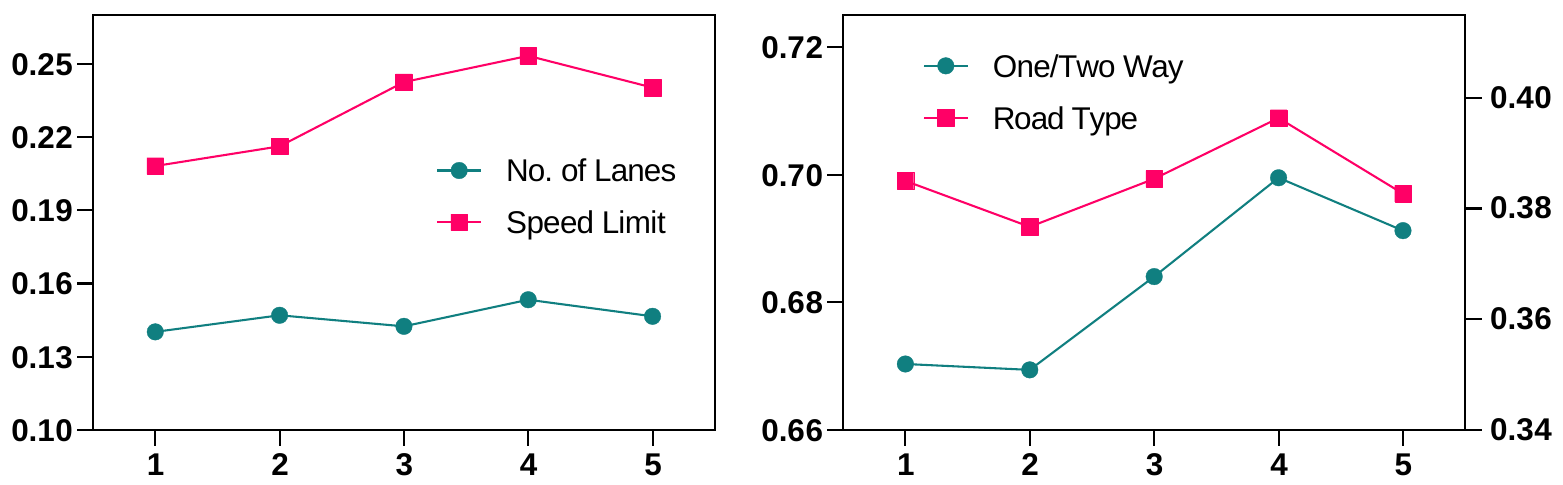}
		\caption{Parameter $L$ (\textit{Singapore})}
		\label{fig:se_layer}
	\end{subfigure}
	\caption{Impact of hyper-parameters $\alpha$, $d$ and $L$ on model performances.}
    \label{fig:se_analyse}
\vspace{-0.1cm}
\end{figure}

\fi 

\subsection{Parameter Sensitivity (RQ3)}  \label{sec:q3}

We conduct the parameter study on the weighted parameter $\alpha$, the embedding size $d$, and the number of layers $L$ on our example dataset \textit{Singapore} as follows. Fig. \ref{fig:se_analyse} shows the model performance in terms of Macro-F1 under different settings.  

\smallskip\noindent\textbf{The weighted parameter $\alpha$.} Generally, the model performances first increase along with $\alpha$, and then begin to drop when $\alpha$ is larger than 0.1. It is worth mentioning that HyperRoad would reduce to w/o HPT when $\alpha$ is zero, and to w/o GPT when $\alpha$ is infinity. The discovered trend conforms to the their performances in ablation study and again shows the necessity of introducing both self-supervised learning tasks.

\smallskip\noindent\textbf{The embedding size $d$.} The model performances increase with embedding size in most cases, since large embedding size tends to have stronger representational power. However, using high-dimensional representation does not always produce the best results. For example, the Macro-F1 achieves the best when $d=64$ rather than $d=128$ on road type classification.

\smallskip\noindent\textbf{The number of layers $L$.} The model performances remain relatively stable when $L$ increases, and the best results for all application tasks can be achieved when $L = 4$. The model performance at $L = 5$ is slightly worse than that of $L = 4$, showing potential for overfitting and oversmoothing.

\begin{figure}[t]
	\centering
	\begin{subfigure}[b]{0.50\textwidth}
		\includegraphics[width=\textwidth]{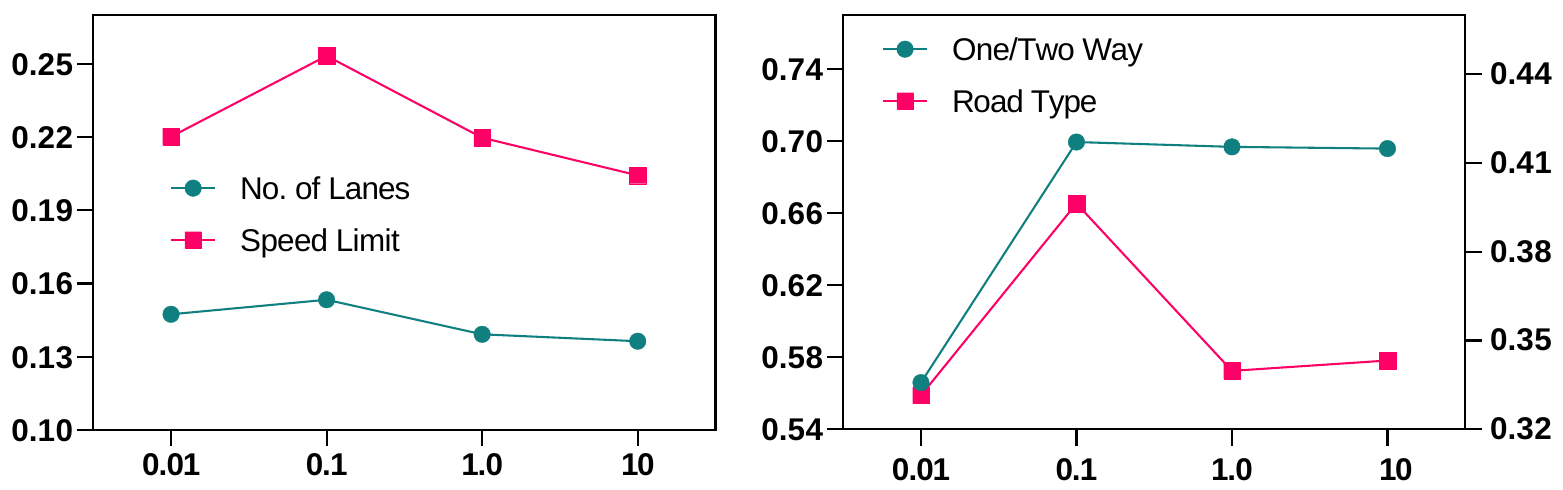}
		\caption{Parameter $\alpha$ (\textit{Singapore})}
		\label{fig:se_alpha}
	\end{subfigure}
	
	\begin{subfigure}[b]{0.50\textwidth}
		\includegraphics[width=\textwidth]{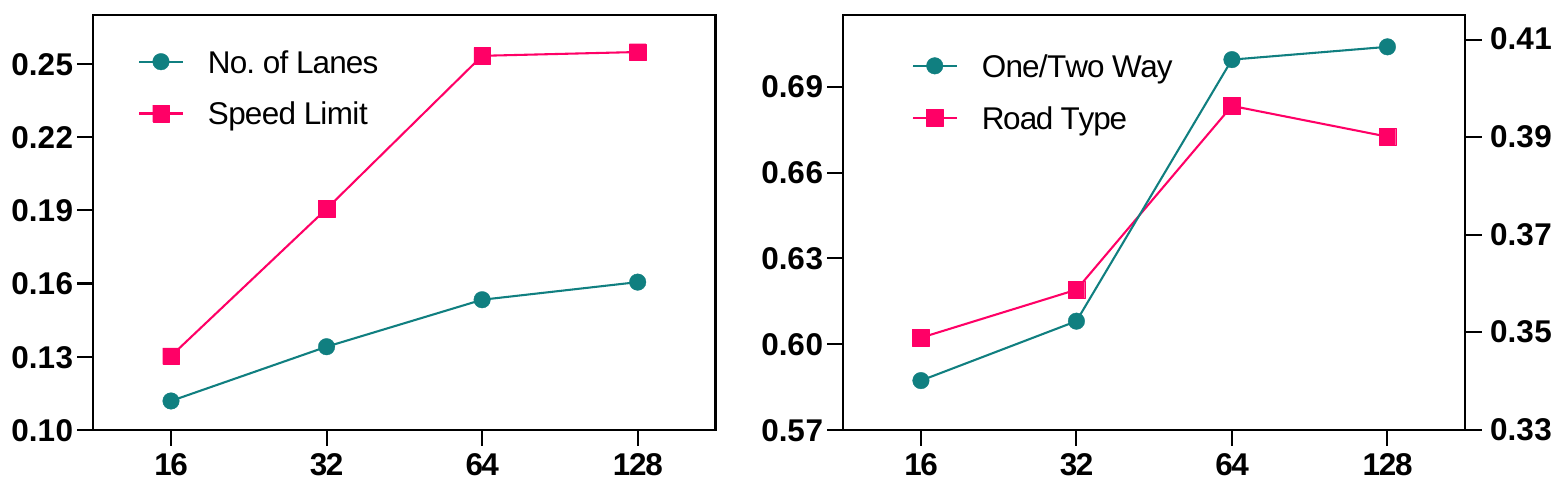}
		\caption{Parameter $d$ (\textit{Singapore})}
		\label{fig:se_d}
	\end{subfigure}
	
	\begin{subfigure}[b]{0.50\textwidth}
		\includegraphics[width=\textwidth]{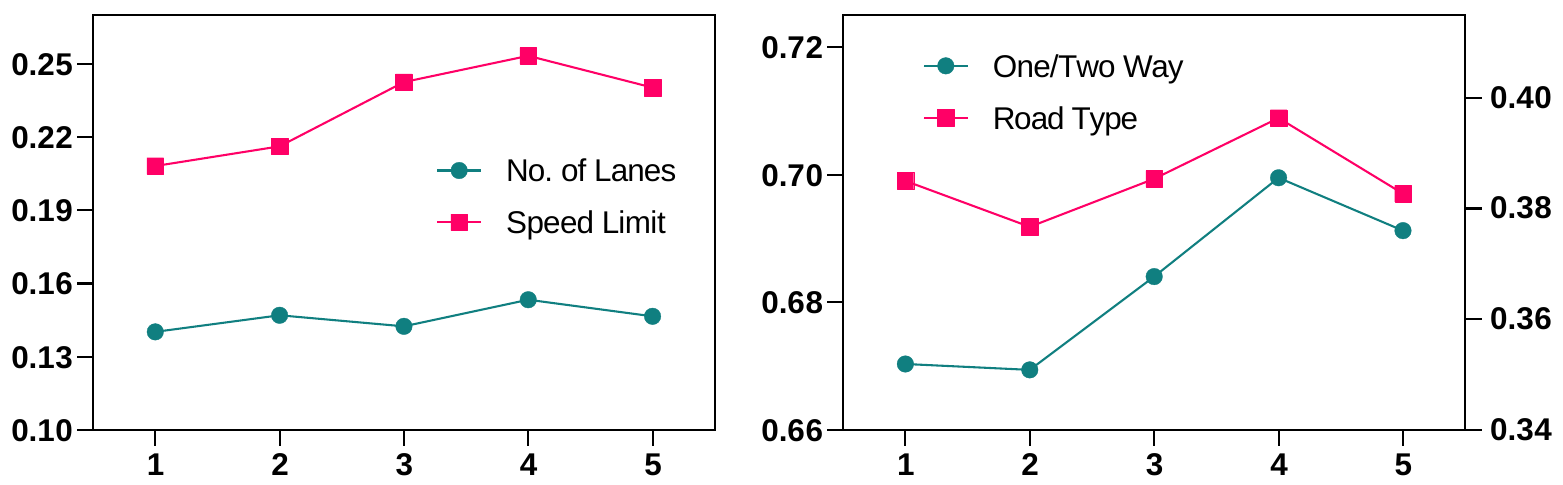}
		\caption{Parameter $L$ (\textit{Singapore})}
		\label{fig:se_layer}
	\end{subfigure}
	\caption{Impact of hyper-parameters $\alpha$, $d$ and $L$ on model performances.}
    \label{fig:se_analyse}
\vspace{-0.1cm}
\end{figure}

\begin{figure*}[t!]
	\centering
	\begin{subfigure}[b]{0.175\textwidth}
		\includegraphics[width=\textwidth]{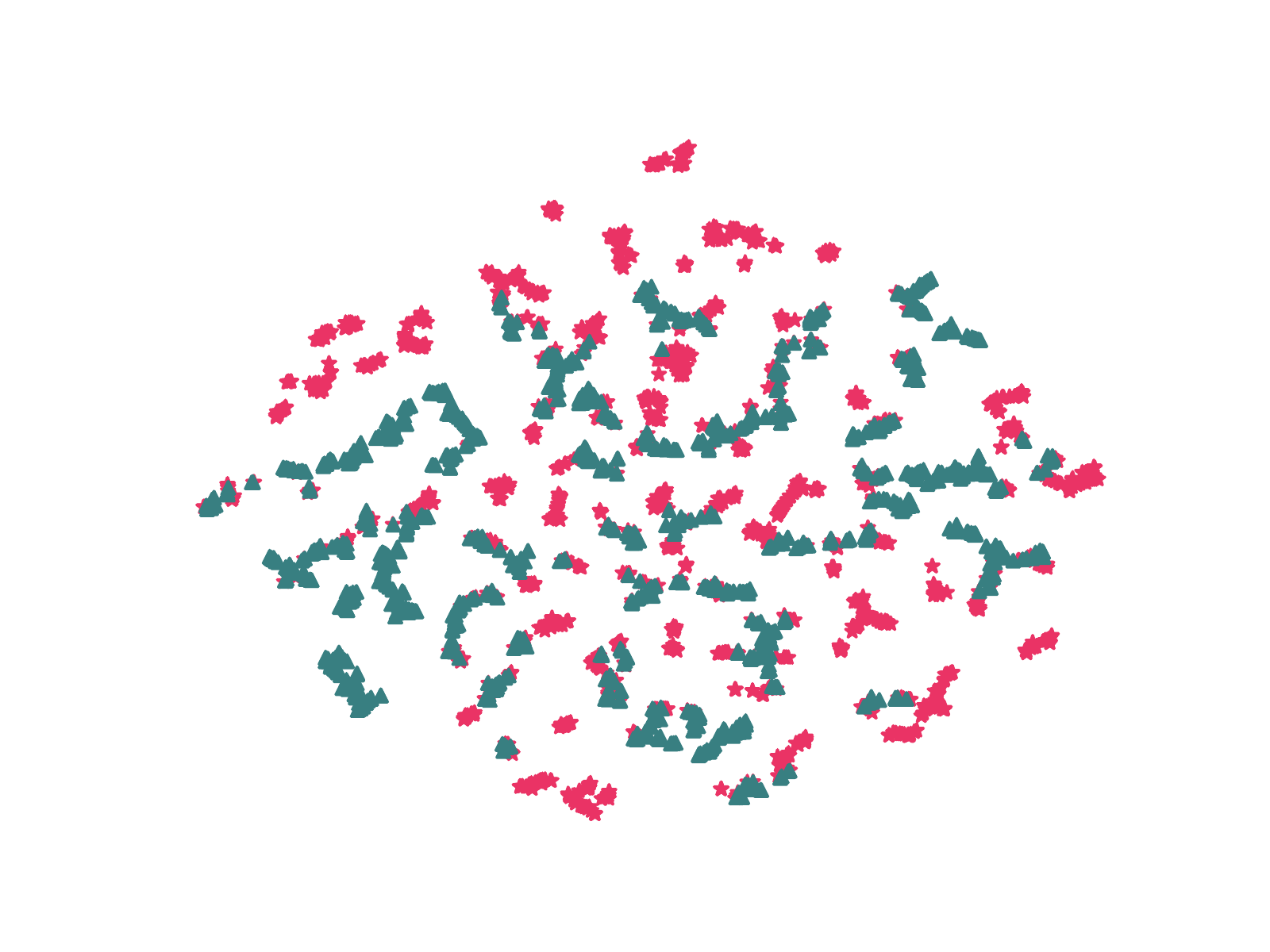}
		\caption{Node2Vec}
		\label{fig:nodev}
	\end{subfigure}
	\hspace{0.01\textwidth}
	\begin{subfigure}[b]{0.175\textwidth}
		\includegraphics[width=\textwidth]{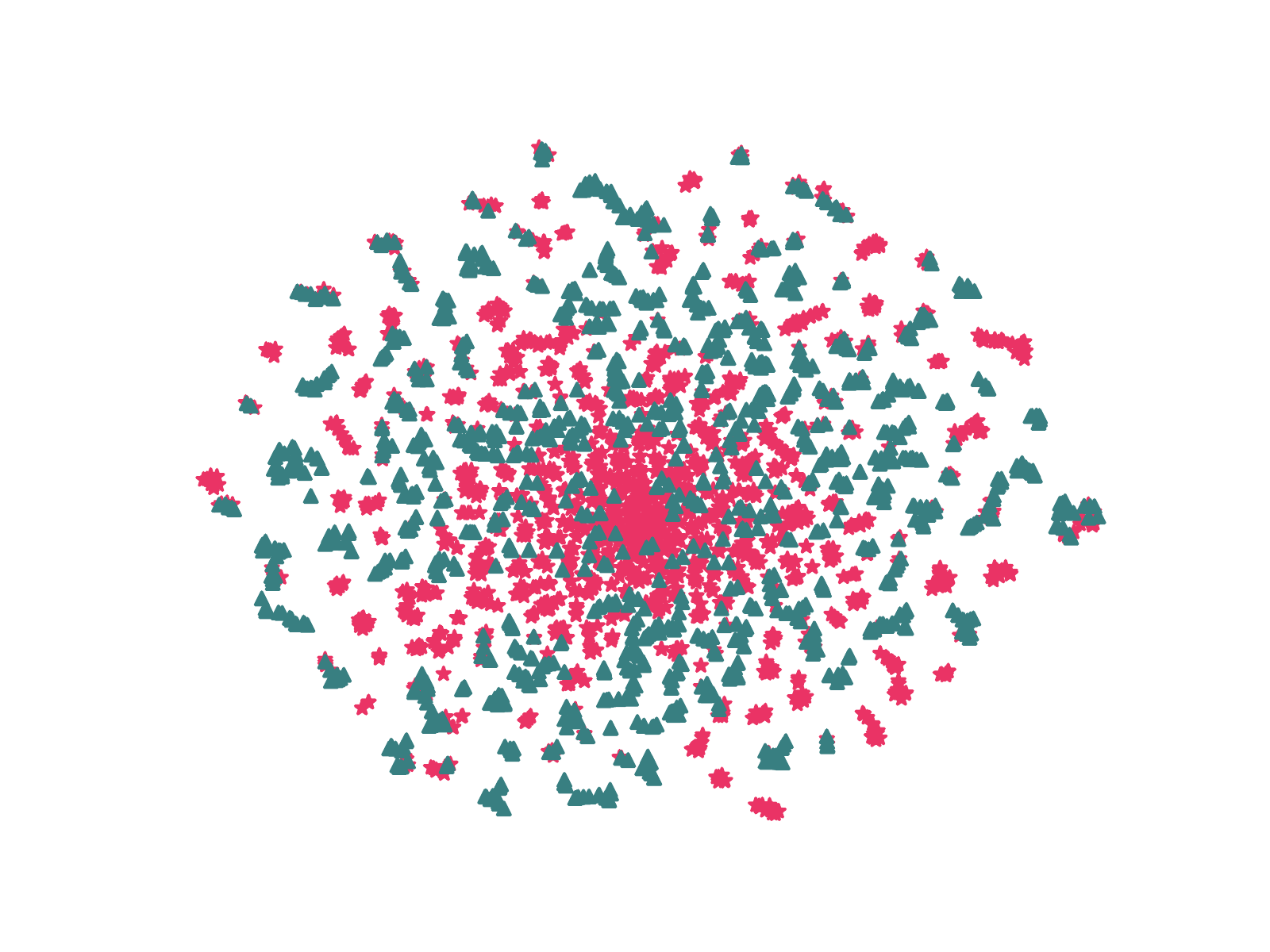}
		\caption{LINE}
		\label{fig:line}
	\end{subfigure}
	\hspace{0.01\textwidth}
	\begin{subfigure}[b]{0.175\textwidth}
		\includegraphics[width=\textwidth]{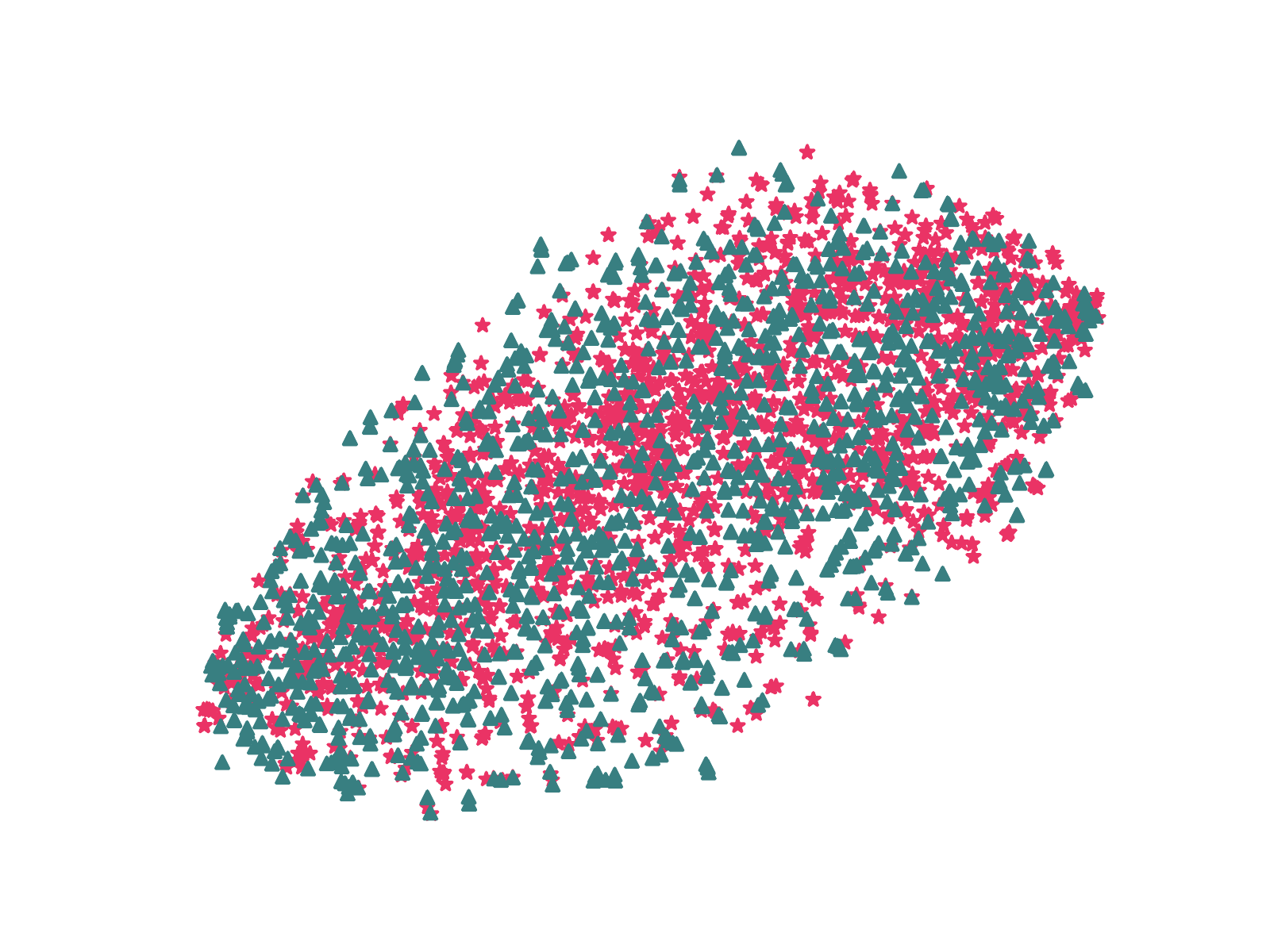}
		\caption{SDNE}
		\label{fig:sdne}
	\end{subfigure}
	\hspace{0.01\textwidth}
	\begin{subfigure}[b]{0.175\textwidth}
		\includegraphics[width=\textwidth]{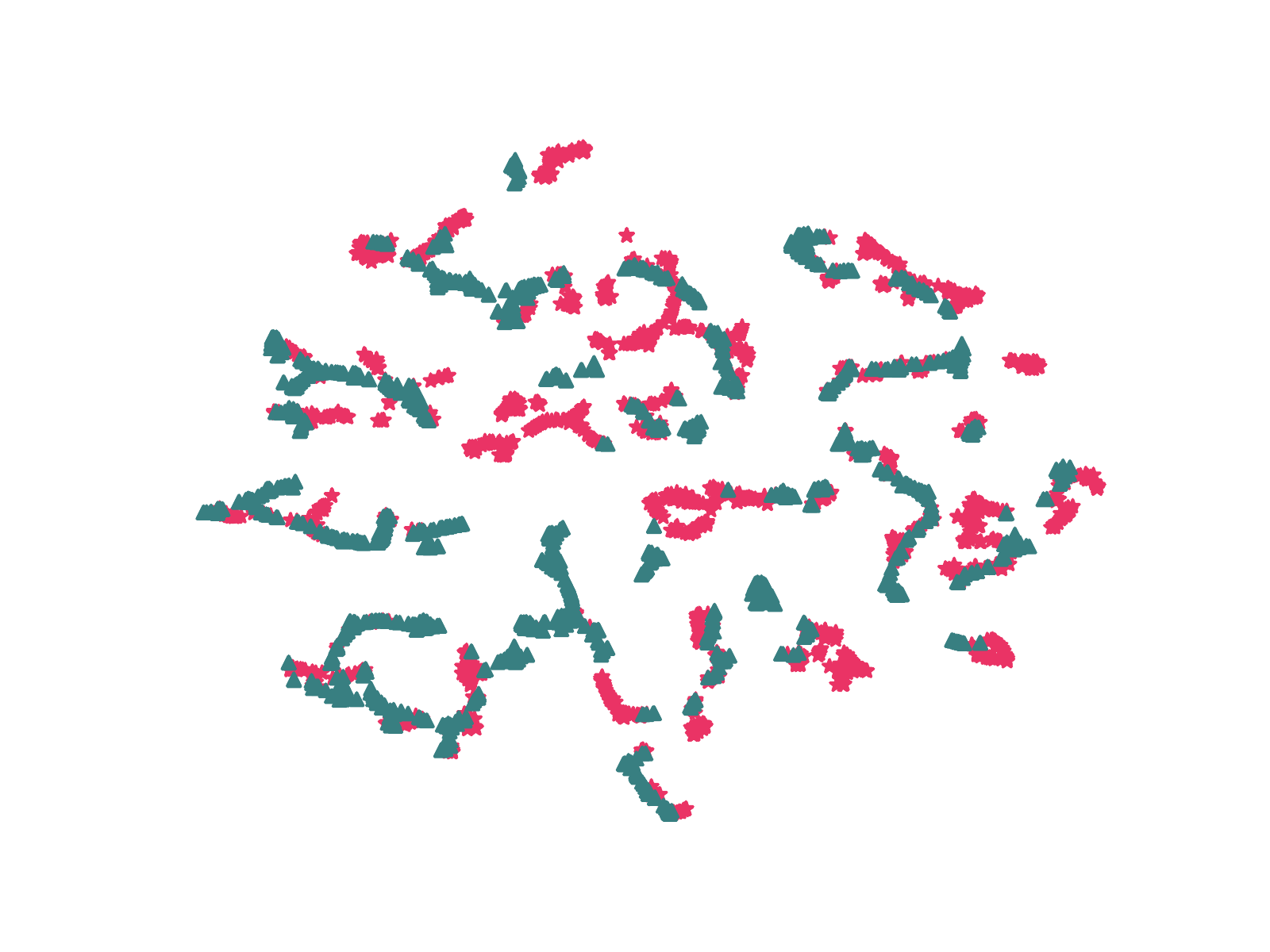}
		\caption{GAE}
		\label{fig:gae}
	\end{subfigure}
	\hspace{0.01\textwidth}
	\begin{subfigure}[b]{0.175\textwidth}
		\includegraphics[width=\textwidth]{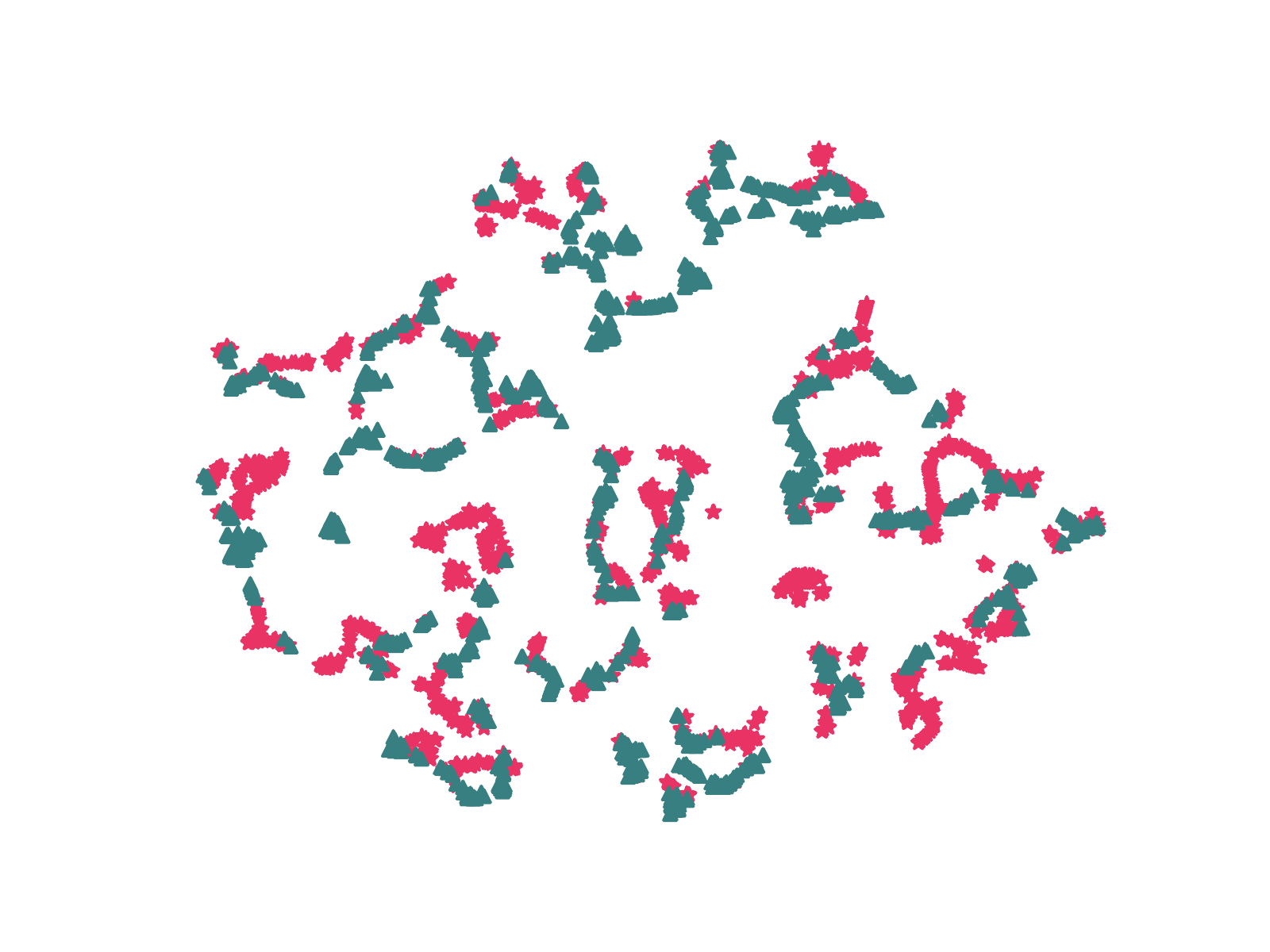}
		\caption{GraphSAGE}
		\label{fig:sage}
	\end{subfigure}
	\hspace{0.01\textwidth}
    
    \begin{subfigure}[b]{0.175\textwidth}
		\includegraphics[width=\textwidth]{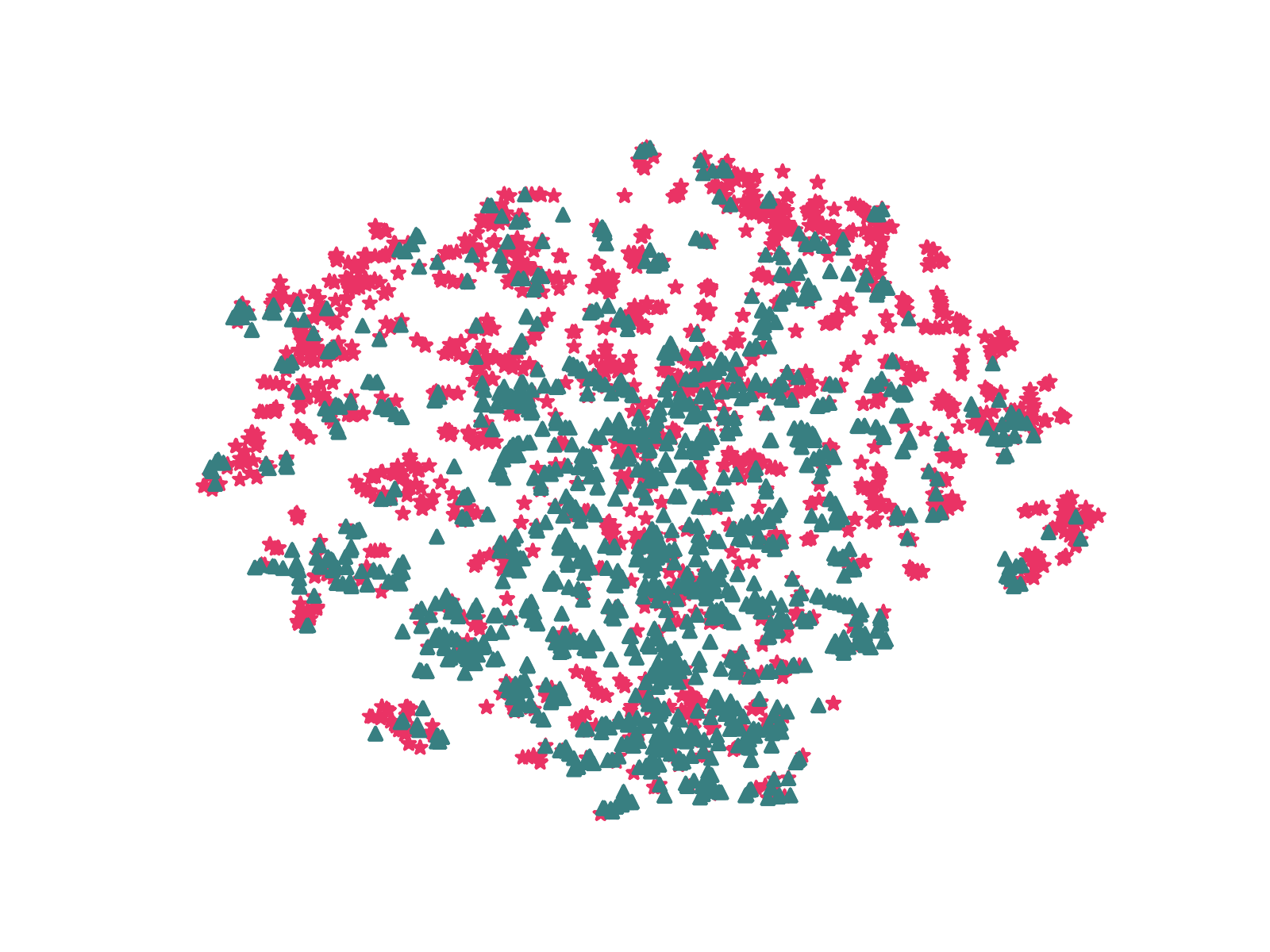}
		\caption{RFN}
		\label{fig:rfn}
	\end{subfigure}
	\hspace{0.01\textwidth}
	\begin{subfigure}[b]{0.175\textwidth}
		\includegraphics[width=\textwidth]{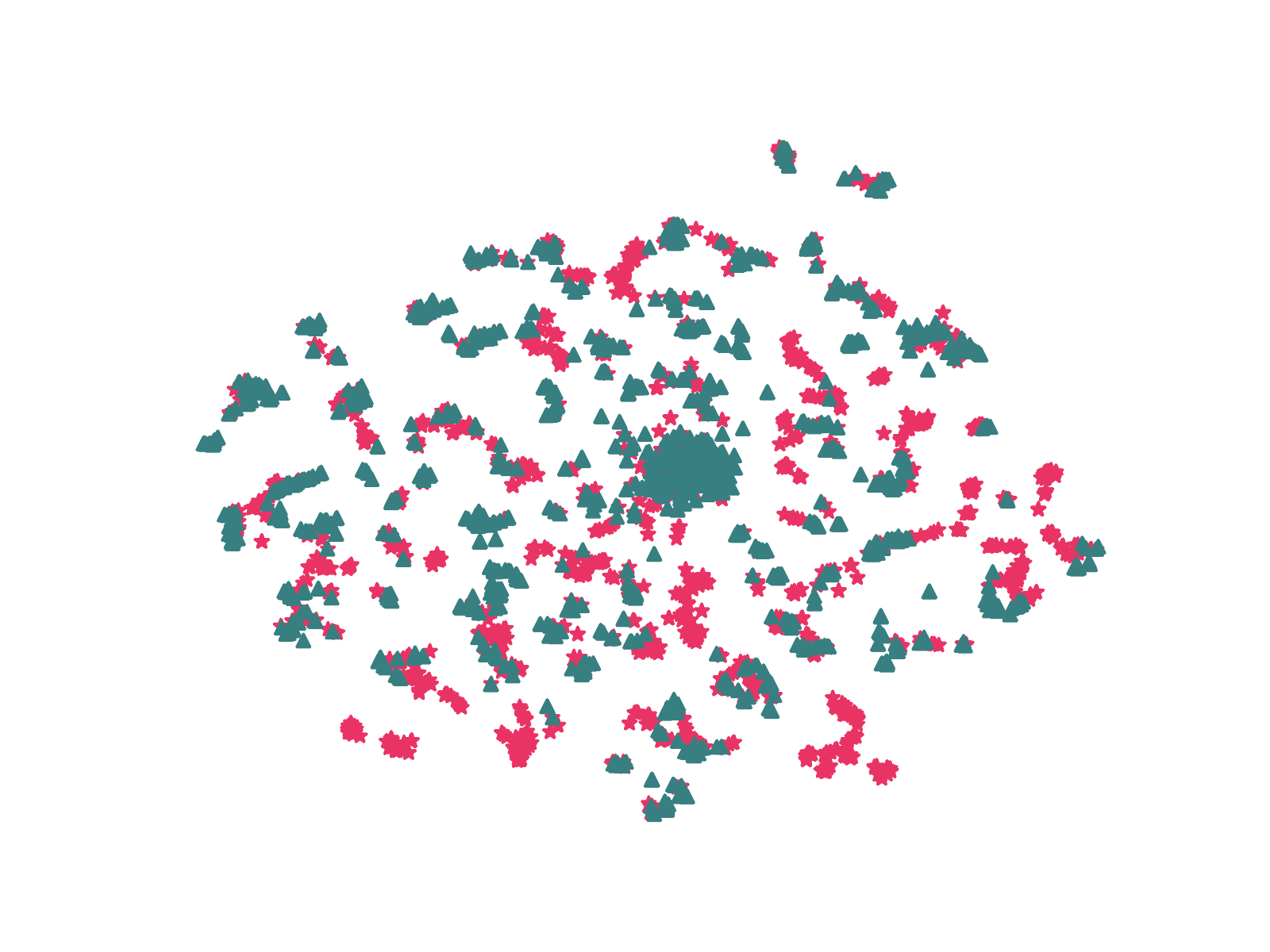}
		\caption{IRN2Vec}
		\label{fig:irn}
	\end{subfigure}
	\begin{subfigure}[b]{0.175\textwidth}
		\includegraphics[width=\textwidth]{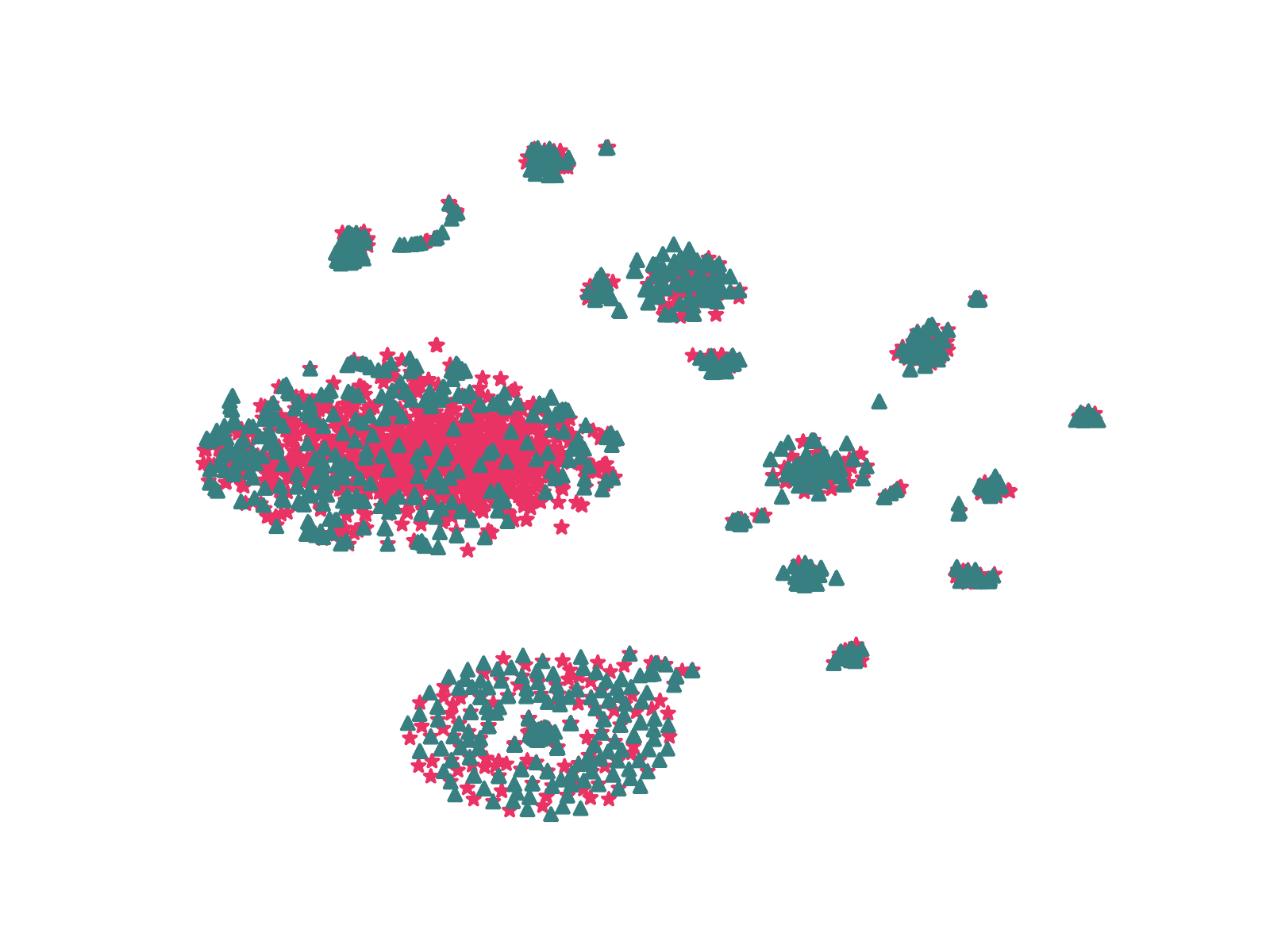}
		\caption{HRNR}
		\label{fig:hrnr}
	\end{subfigure}
	\hspace{0.01\textwidth}
	\begin{subfigure}[b]{0.175\textwidth}
		\includegraphics[width=\textwidth]{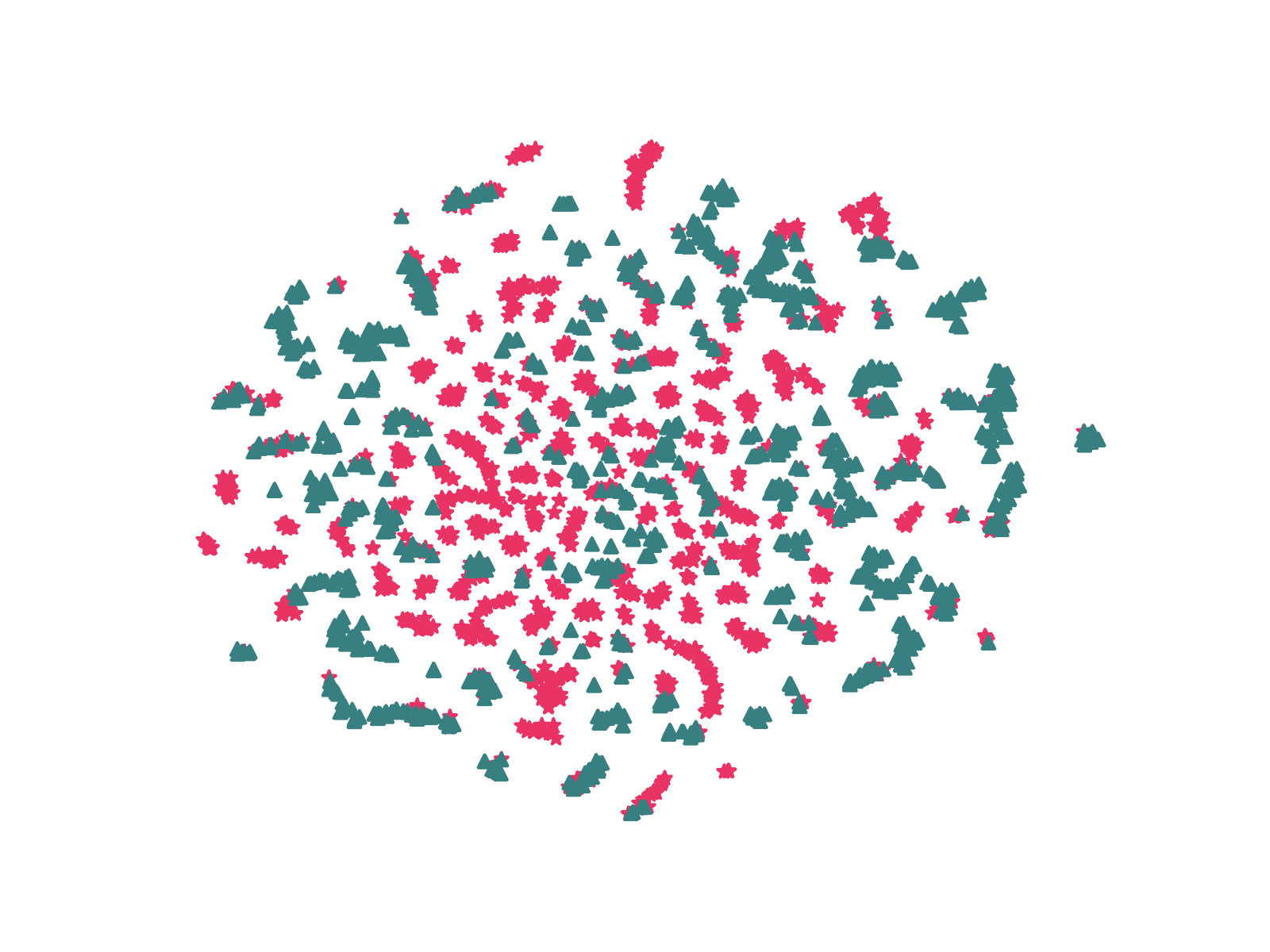}
		\caption{Toast}
		\label{fig:dw}
	\end{subfigure}
    \hspace{0.01\textwidth}
	\begin{subfigure}[b]{0.175\textwidth}
		\includegraphics[width=\textwidth]{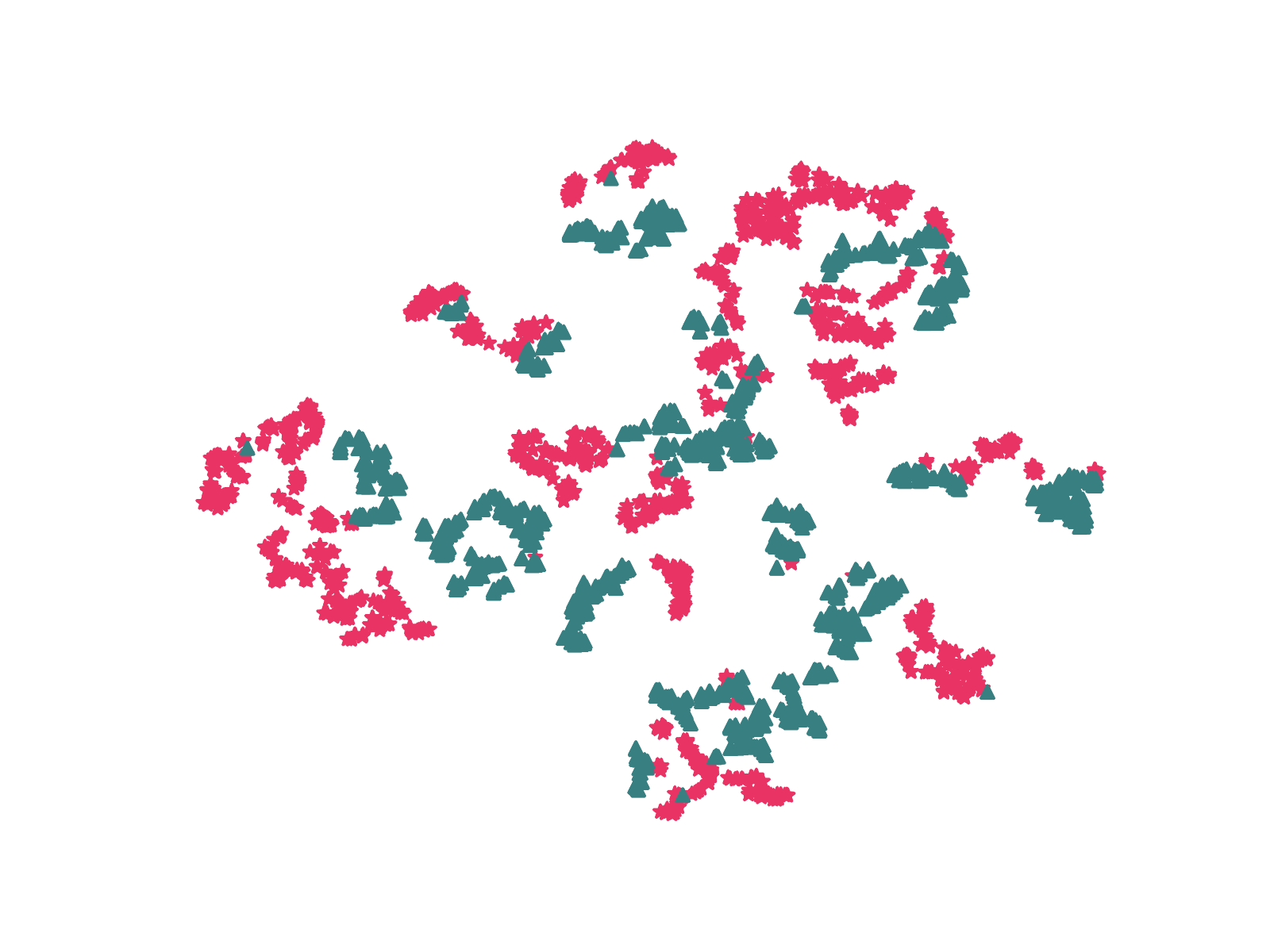}
		\caption{HyperRoad}
		\label{fig:hyper}
	\end{subfigure}
% 	\vspace{-2mm}
	\caption{Road embedding visualizations. Color of a node indicates its road type. Green: "motorway", Red: "residential".}
	\label{fig:embedding}
\vspace{-0.0cm}
\end{figure*}

% {\color{red}Comments on Figure 6: Present the results following the previous ordering of models for better consistency?}

% {\LIANG{Have changed.}}

% {\color{red}If necessary, we can omit the figures for parameter studies for saving space.}

\subsection{Case Study for Representation Understanding (RQ4)}  \label{sec:q4}
In this section, we use auxiliary information such as road attributes to help uncover what HyperRoad has learned. Note that these information is not available for model training in our default problem setting.

We first exploit the network visualization approach to evaluate node embeddings learned by different models. Here, we take a rectangular region of \textit{Singapore} as an example. The longitude is between $103^\circ 77^\prime \text{E}$ and $103^\circ 84^\prime \text{E}$, and the latitude is between $1^\circ 33^\prime \text{N}$ and $1^\circ 43^\prime \text{N}$. Then, we select roads located in this region and employ the t-SNE model \cite{van2008visualizing} to project road embeddings to a 2-dimensional space. For the sake of visualization, we present the results for two road types with different semantics namely motorway and residential. The projected results of different models are shown in Fig. \ref{fig:embedding}.
% , using the color of a node to indicate its road type. 
From the results, we make the following observations. 1) The shallow network representation models failed to separate the two road types clearly. For example, in Node2Vec and IRN2Vec, two road types are mixed together in the middle of Fig. \ref{fig:nodev} and Fig. \ref{fig:irn}. 
% Although IRN2Vec considers the geo-location and shape information in road network compared with Node2Vec, it still can not achieve satisfactory visualizing results as shown in Fig. \ref{fig:irn}. 
The reason is that random walks may not preserve the node homogeneity, and thus lead to similar representations for roads with different semantics. 2) There are different clusters in the results of HRNR as shown in Fig. \ref{fig:hrnr}. This is because HRNR is a hierarchical GNN structure, and each node can aggregate information from too many neighbors through the hidden nodes. This aggregation mechanism leads to the oversmoothing problem. 3) RFN, GraphSAGE and GAE provide 
% better performance than other baselines and obtain 
more reasonable visualizing results. However, some clusters are still mixed with each other and there is no clear margin between different classes. 4) In contrast, HyperRoad gives the most satisfying identification with relatively clear cluster boundaries.
% , which demonstrates that the road functionalities are well preserved by our model. 
The reason is that the proposed dual-channel aggregation mechanism can help the model to find high-order neighbors with similar semantics, and the hypergraph-level tasks provides extra prior knowledge for road representation learning. 

\begin{figure*}[t]
	\centering

	\begin{subfigure}[b]{0.40\textwidth}
		\includegraphics[width=\textwidth]{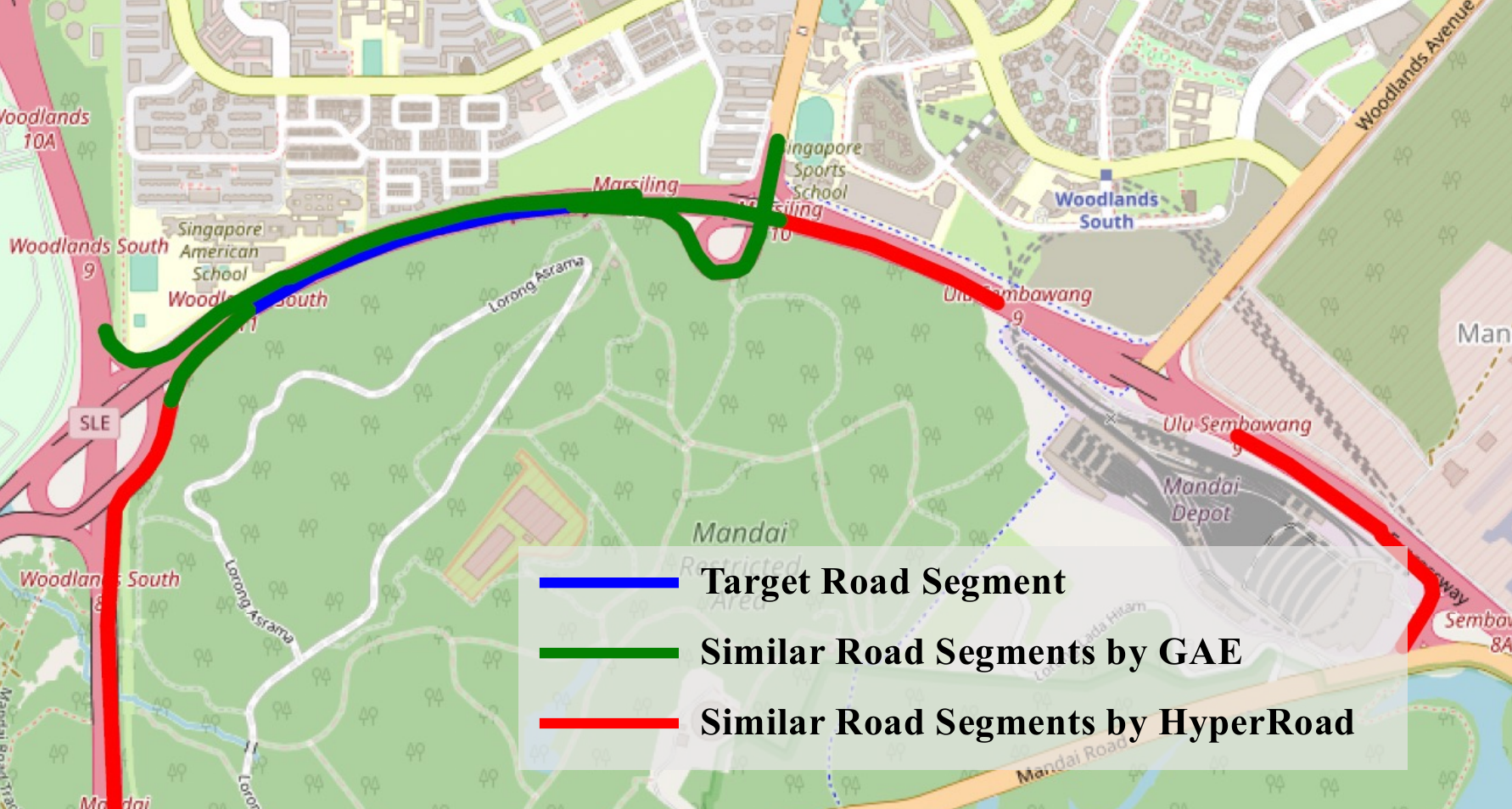}
		\caption{Case one: $r5028$}
		\label{fig:se_alpha}
	\end{subfigure}
	\begin{subfigure}[b]{0.40\textwidth}
		\includegraphics[width=\textwidth]{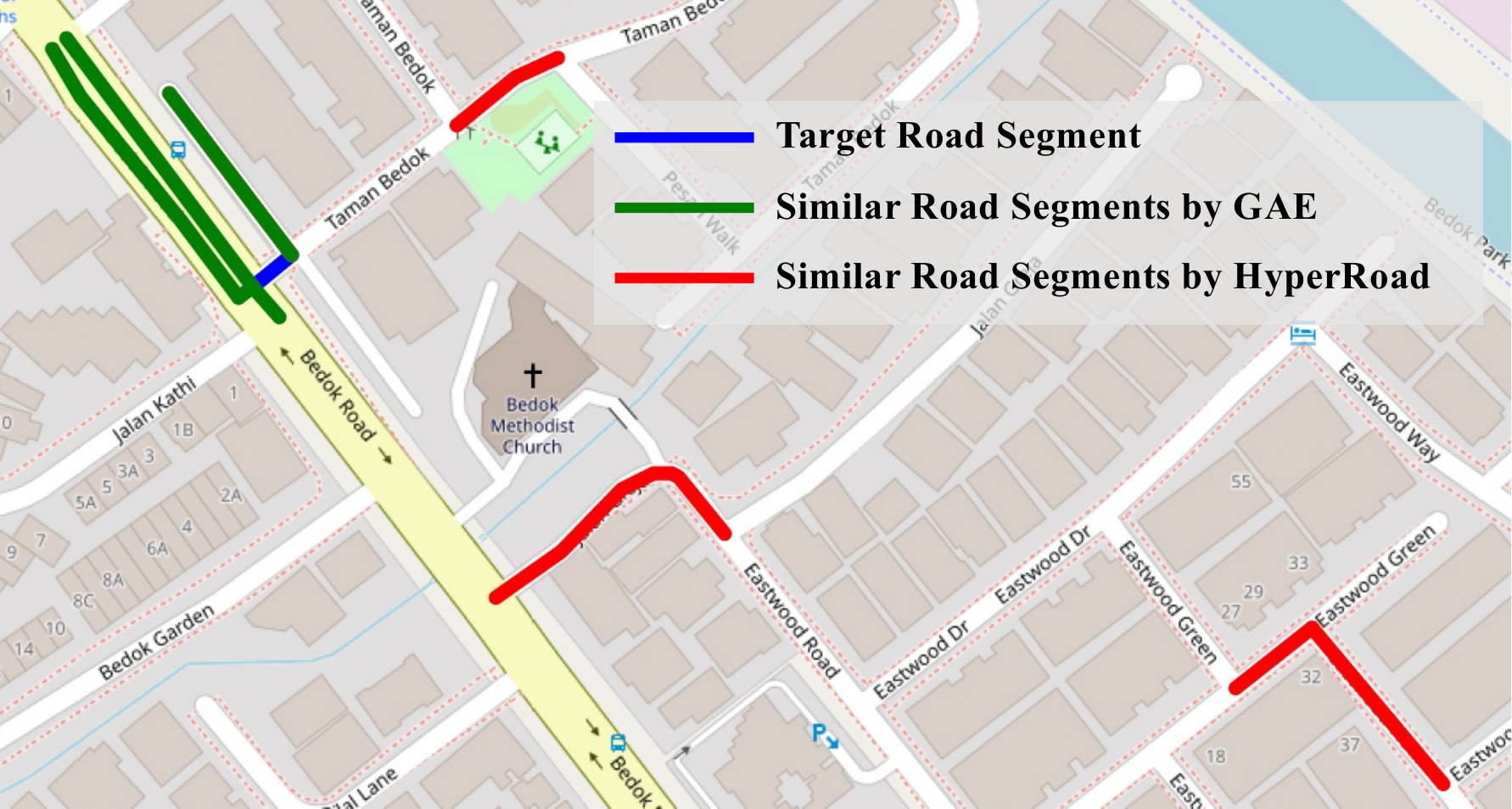}
		\caption{Case two: $r30396$}
		\label{fig:se_d}
	\end{subfigure}
	\caption{Road similarity visualizations.}
	\label{fig:simi}
\vspace{-0.1cm}
\end{figure*}

We further pick two specific cases to explain the road embeddings learned by different models in details. GAE is selected as a reference model given its satisfactory task performances and visualizing results. 
Specifically, we first randomly sample two roads $r5028$ and $r30396$ with distinct semantic information as target roads. For each road, we extract its top 5 most similar roads in GAE and HyperRoad with cosine similarity. The road attributes are shown in Table \ref{tab:simi} and the geo-locations are plotted in Fig. \ref{fig:simi}. It is clear to see that GAE mainly recalls very close neighbors in geo-space although their attributes are significantly different. For example, 
% in case one, the ramp is taken as the similar road for the target road with higher speed. And 
the motorways are considered similar to residential roads in case two. The results show that the adopted graph reconstruction objective {\LIANGTwo in GAE} is not effective in modeling complex road networks. In contrast, our model captures both location and semantic information together for querying similar roads. For example, in both case one and case two, our model can find the roads with quite similar functions in a reasonable long range, which justifies the necessity of the proposed hypergraph modeling.

\begin{table*}[t!]
	\renewcommand{\arraystretch}{1.10} 
	\centering 
	\caption{A show case of road similarity.}
    \LARGE
	\resizebox{0.77 \linewidth}{!}{
		\begin{tabular}{c|ccccc|ccccc}
			\toprule
			\multirowcell{3}{Category} & \multicolumn{5}{c|}{{\textbf{ Case One}}} & \multicolumn{5}{c}{{\textbf{Case Two}}} \\
			\cmidrule{2-11}
			& ID & \tabincell{c}{One/Two \\ Way} & \tabincell{c}{No. of \\ Lanes} & \tabincell{c}{Speed \\ Limit} & \tabincell{c}{Road \\ Type} & ID & \tabincell{c}{One/Two \\ Way} & \tabincell{c}{No. of \\ Lanes} & \tabincell{c}{Speed \\ Limit} & \tabincell{c}{Road \\ Type}  \\ \midrule
			Target road & $r5028$ & one way & 3 & 90 & motorway & $r30396$ & two way & 2 & 50 & residential \\
			\midrule
			\multirowcell{5}{Similar roads \\ by \\ GAE} & $r6832$ & one way & 4 & 90 & motorway & $r2133$ & two way & 2 & 50 & residential \\
			& $r16803$ & one way & 3 & 90 & motorway & $r21127$ & one way & 1 & 50 & motorway  \\

			& $r5027$ & one way & 2 & 50 & motorway & $r5514$ & one way & 1 & 50 & motorway  \\ 
			& $r16804$ & one way & 2 & 50 & motorway & $r5995$ & one way & 1 & 50 & motorway \\
			& $r6831$ & one way	& 1 & 50 & motorway & $r27877$ &	one way & 1 & 50 & motorway  \\ 
			\midrule
			\multirowcell{5}{Similar roads \\ by \\ HyperRoad} & $r16803$ & one way & 3 & 90 & motorway & $r6328$ & two way & 2 & 50 & residential \\
			& $r28968$ & one way & 3 & 90 & motorway & $r2133$ & two way & 2 & 50 & residential \\
			& $r25372$ & one way & 4 & 90 & motorway & $r32071$ & two way & 2 & 50 & residential \\
			& $r6832$ & one way & 4 & 90 & motorway & $r9356$ & two way & 2 & 50 & residential \\
			& $r3221$ & one way & 4 & 70 & motorway & $r29281$ & two way & 2 & 50 & residential \\
			\bottomrule
	\end{tabular}}
	\label{tab:simi}
\vspace{-0.1cm}
\end{table*}

\subsection{Extensions to Other Settings (RQ5)}  \label{sec:q5}

% In this section, we examine whether HyperRoad extensions can outperform the state-of-the-art models on \textit{Singapore} dataset in settings where additional attributes and/trajectories are available. 
In this section, we study road type classification task (by excluding the road type attributes in the inputs) and the travel time estimation task on the Singapore dataset.
The model performances under various settings are shown in Table \ref{tab:A_T_AT}. 
% In this table, besides the innate structure and spatial information of road network, we incorporate extra semantic attributes and human trajectory information flexibly into the model training. 
Based on the results, we have the following observations.

\smallskip\noindent\textbf{Comparisons among problem settings.} In general, including additional inputs can help to learn better road network representations. For example, HyperRoad-AT significantly exceeds its counterpart HyperRoad with relative improvements of 33.5\% and 11.2\% in terms of Macro-F1 and RMSE, respectively. Similar findings can also be witnessed by comparing any extended model with its base model. 
In addition, we find that using road attributes is more effective for road type classification, while using trajectory is more effective for travel time estimation. 

\smallskip\noindent\textbf{Comparisons among models.} By comparing the performances of HyperRoad extensions with other baseline models, we can find that the proposed model achieves the best results on two tasks across all the settings studied. For example, HyperRoad-AT outperforms the best baseline Toast-AT by 16.8\% in terms of Macro-F1 on road type classification task. 
% This results demonstrate that our model can not only support a newly constructed road network (or those in underdeveloped areas) without any extra information, but also it can be used when various extra information is available.

\begin{table}[t!]
    % \vspace{-2mm}
    \renewcommand{\arraystretch}{1.05} 
	\centering
	\tiny
	\caption{Model performances under various settings.}
	\label{tab:A_T_AT}
	\resizebox{0.70 \linewidth}{!}{
		\begin{tabular}{@{}l|ccc|cc@{}}
			\toprule
			\multirow{3}{*}{+ Attributes} & \multicolumn{3}{c|}{{\it Road Type Classification}} & \multicolumn{2}{c}{{\it Travel Time Estimation}} \\
			\cmidrule{2-6}
			& Micro-F1 & Macro-F1 & Weighted-F1 & MAE & RMSE  \\ \midrule
		    RFN-A & 0.7264 & \underline{0.4606} & \underline{0.6919} & 152.20 & 211.16 \\ 
		    IRN2Vec-A & 0.5054 & 0.2732 & 0.4628 & \underline{150.44} & 210.99 \\
		    HRNR-A & 0.6776 & 0.3670 & 0.6212 & 165.21 & 232.26 \\ 
		    Toast-A & \underline{0.7353} & 0.4254 & 0.6792 & 155.06 & \underline{208.78}  \\
		    \midrule
		    HyperRoad-A & \textbf{0.7497} & \textbf{0.5166} & \textbf{0.7110} & \textbf{145.67} & \textbf{205.69} \\ Improvement & 1.96\% & 12.16\% & 2.76\% & 3.17\% & 1.48\%  \\
			\bottomrule
			\toprule
			
			\multirow{3}{*}{+ Trajectory} & \multicolumn{3}{c|}{{\it Road Type Classification}} & \multicolumn{2}{c}{{\it Travel Time Estimation}} \\
			\cmidrule{2-6}
			& Micro-F1 & Macro-F1 & Weighted-F1 & MAE & RMSE  \\ \midrule
		    HRNR-T & 0.5989 & 0.3244 & 0.5492 & 155.12 & 213.22 \\ 
		    Toast-T & \underline{0.6210} & \underline{0.3364} & \underline{0.5695} & \underline{144.44} & \underline{205.18}  \\
		    \midrule
		    HyperRoad-T & \textbf{0.6327} & \textbf{0.4294} & \textbf{0.5994} & \textbf{141.28} & \textbf{195.42} \\ Improvement & 1.88\% & 27.65\% & 5.25\% & 2.19\% & 4.76\%  \\
			\bottomrule
			\toprule
			
			\multirowcell{3}{+ Attributes \& \\ Trajectory} & \multicolumn{3}{c|}{{\it Road Type Classification}} & \multicolumn{2}{c}{{\it Travel Time Estimation}} \\
			\cmidrule{2-6}
			& Micro-F1 & Macro-F1 & Weighted-F1 & MAE & RMSE  \\ \midrule
		    HRNR-AT & 0.7060 & 0.3823 & 0.6473 & 154.98 & 208.87 \\ 
		    Toast-AT & \underline{0.7512} & \underline{0.4533} & \underline{0.6996} & \underline{142.23} & \underline{199.52}  \\
		    \midrule
		    HyperRoad-AT & \textbf{0.7621} & \textbf{0.5294} & \textbf{0.7233} & \textbf{137.93} & \textbf{189.39} \\
			Improvement & 1.45\% & 16.79\% & 3.39\% & 3.03\% & 5.08\%  \\
			\bottomrule
	\end{tabular}}
\vspace{-0.1cm}
\end{table}

\section{Related Work}\label{sec:relate}

% Our work is related to the following topics including graph representation learning, road networks representation modelling and hypergraph learning.  

% \subsection{Road Networks Modelling and Representation Learning} 
\noindent\textbf{Road Networks Modeling and Representation Learning.}
Road network has been used in various  applications, such as traffic inference and forecasting \cite{hu2019stochastic, guo2019attention}, road tags prediction \cite{yin2021multimodal}, and region functionality modeling \cite{wozniak2021hex2vec}. 
In this studies, road embeddings are indirectly learned as a byproduct to incorporate structural and spatial information. However, the methods in these studies are tailored for specific tasks, and the representations learned for one task can hardly be transferred to other tasks. 
Recently, a few studies propose to learn generic road representations by extending graph representation learning to road networks \cite{jepsen2020relational, wang2019learning, wu2020learning, chen2021robust}. In specific, IRN2Vec \cite{wang2019learning} extends the skip-gram model by considering geo-locality and geo-shape information. Toast \cite{chen2021robust} extends skip-gram model by adding auxiliary attributes prediction task and trajectory-enhanced transformer module. 
RFN \cite{jepsen2020relational} and HRNR \cite{wu2020learning} extend GCN by considering dual node-relational and edge-relational views and hierarchical graph structures respectively.
However, none of these methods are sufficient in capturing the high-order relationships and long-range relationships, as has been explained in Section~\ref{sec:intro}.
% propose to learn high-order road interactions, which is vital for high-quality representation learning. They suffer from either shallow representation or expressive power issues. 

% {\color{red}I believe this subsection should be covering existing RNRL methods?}

% {\LIANG{Have changed.}}

% \subsection{Graph Representation Learning}
\smallskip\noindent\textbf{Graph Representation Learning.}
Recent years have witnessed the great success of
graph representation learning.
% in modelling graph data. %\cite{goyal2018graph}. 
The earliest efforts in this field mainly focus on shallow representations. For example, DeepWalk \cite{perozzi2014deepwalk} and Node2Vec \cite{grover2016node2vec} use random walks on the graph to generate node sequences and then apply skip-gram model \cite{mikolov2013distributed} to learn node representations. LINE \cite{tang2015line} aims to preserve the first and second-order proximities by explicitly modeling the corresponding objectives. All these models are actually equivalent to the simple matrix factorization framework with closed forms \cite{qiu2018network}. 
Inspired by the recent huge success of deep learning, graph neural networks have been introduced for graph representation learning. Along this line, graph convolution networks (GCN) \cite{kipf2016semi} and graph attention network (GAT) \cite{velivckovic2017graph} are two classic models. These GNN models mainly follow the paradigm of message passing and local neighborhood aggregation and differ in various aggregation operations \cite{xu2018powerful, hamilton2017inductive, zhu2020bilinear}. 
However, all these methods are designed for general graphs, but not road networks, and fail to capture some unique information of road networks such as the spatial information and high-order % and long-range 
relationships among the roads.
% the long range road segment correlations.

% \subsection{Hypergraph Learning}
\smallskip\noindent\textbf{Hypergraph Learning.}
Hypergraph learning is firstly introduced in \cite{zhou2006learning} to generalize the methodology of spectral clustering to hypergraphs, which naturally models high-order relations of three or more entities. The subsequent studies adopt ranking based methods \cite{li2013link} and non-negative matrix factorization methods \cite{zhang2018beyond} for hypergraph mining and hyperedge prediction. Recently, deep learning based methods have been developed in this field. In specific, deep neural networks and auto-encoder are applied in DHNE \cite{tu2018structural} to capture first-order proximity and second-order proximity respectively for hyperedge representation learning. Zhang et al. \cite{zhang2019hyper} applied attention mechanism and MLP network for both dynamic and static hyperedge representations learning to improve hyperedge prediction performance. Recently, hypergraph neural networks (HGNNs) \cite{feng2019hypergraph, jiang2019dynamic} 
have been proposed to generalize the neighborhood aggregation operation to the hypergraph. 
%
% However, to our best knowledge, our proposed method is the first attempt to shift the power of hypergraph to the road network modeling. 
%
{\color{black}Our HyperRoad model differs from these models in (1) the dual-channel aggregating mechanism, which involves propagations via both edges and hyperedges (vs. propagations via hyperedges only in these models), (2) the bi-level self supervised learning module, which involves not only the hypergraph reconstruction task, but also some other tasks such as hyperedge classification in the road network context (this is new), and (3) its extended versions with additional attributes and trajectories as inputs (which are not available in these studies).}

% We design meaningful hypergraph construction strategy in road network and for the first time, and then propose hypergraph based dual-channel aggregation mechanism and self supervised learning tasks for effective representation learning, which compose the key contributions.

% {\color{red}Make sure we don't over claim some ideas here: I feel our aggregating mechanism based on the hypergraph channel is not and it is likely that our hypergraph reconstruction task is not new.}

% {\LIANG{Have changed. I delete a few sentences because it has been emphasized in the introduction.}}
% \vspace{-0.2cm}
\section{Conclusion}
In this paper, we propose a novel model called HyperRoad for road network representation learning (RNRL). 
% HyperRoad is based on three key ideas: (1) constructs a \emph{hypergraph}, where each hyperedge corresponds to a set of multiple roads forming a region, (2) propagates information via both edges (of a simple graph capturing the pairwise relationships) and hyperedges (of the constructed hypergraph) in a context of graph neural network, and (3) optimizes the road representations with a few pretext tasks based on both the simple graph (i.e., graph reconstruction) and the constructed hypergraph (including hypergraph reconstruction and hyperedge classification). 
HyperRoad is superior over existing models since it captures pairwise relationships, as well as high-order relationships and long-range relationships among roads, with the help of a hypergraph constructed on top of the roads. We further extend HyperRoad to other problem settings with additional road attributes and/or trajectories. Extensive experiment results demonstrate that our model achieves impressive improvements compared with existing baselines in all the experimental settings considered. In the future, we plan to study the RNRL problem under settings with incomplete data (e.g., the road attributes are available for some roads only)
% and/or the trajectories traverse a fraction of roads only) 
and/or with types of data that has not been explored before (e.g., satellite images of roads, etc).

% In this paper, we study the road network representation learning problem under different settings, and introduce a new structure named hypergraph into this problem. Based on that we propose a novel end-to-end pre-training framework HyperRoad, which features in the schemes of dual-channel aggregation mechanism to enhance model expressive power, and bi-level self-supervised pretext tasks (graph-level and hypergraph-level respectively) to capture both local and global road correlations. To cope with the potential extra road network information, we propose three advanced models to extend HyperRoad flexibly. Extensive experiments on five downstream tasks demonstrate the great improvements of our model compared with existing baselines. The ablation and case study results also validate the necessity of introducing hypergraphs into road network modelling.   

\section*{ACKNOWLEDGMENTS}
This study is supported under the RIE2020 Industry Alignment Fund Industry Collaboration Projects (IAF-ICP) Funding Initiative, as well as cash and in kind contribution from Singapore Telecommunications Limited Singtel, through Singtel Cognitive and Artificial Intelligence Lab for Enterprises (SCALE@NTU). This research is also supported by the Ministry of Education, Singapore, under its Academic Research Fund (Tier 2 Award MOE-T2EP20221-0013). Any opinions, findings and conclusions or recommendations expressed in this material are those of the author(s) and do not reflect the views of the Ministry of Education, Singapore.

% \section{Acknowledgments}

%%
%% Print the bibliography
%%
\printbibliography

@inproceedings{xia2021spatial,
  title={Spatial-Temporal Sequential Hypergraph Network for Crime Prediction with Dynamic Multiplex Relation Learning.},
  author={Xia, Lianghao and Huang, Chao and Xu, Yong and Dai, Peng and Bo, Liefeng and Zhang, Xiyue and Chen, Tianyi},
  booktitle={IJCAI},
  pages={1631--1637},
  year={2021}
}

@article{luo2022directed,
  title={Directed hypergraph attention network for traffic forecasting},
  author={Luo, Xiaoyi and Peng, Jiaheng and Liang, Jun},
  journal={IET Intelligent Transport Systems},
  volume={16},
  number={1},
  pages={85--98},
  year={2022},
  publisher={Wiley Online Library}
}

@inproceedings{wang2021tree,
  title={Tree decomposed graph neural network},
  author={Wang, Yu and Derr, Tyler},
  booktitle={Proceedings of the 30th ACM International Conference on Information \& Knowledge Management},
  pages={2040--2049},
  year={2021}
}

@inproceedings{yuan2020effective,
  title={Effective travel time estimation: When historical trajectories over road networks matter},
  author={Yuan, Haitao and Li, Guoliang and Bao, Zhifeng and Feng, Ling},
  booktitle={Proceedings of the 2020 acm sigmod international conference on management of data},
  pages={2135--2149},
  year={2020}
}

@inproceedings{yuan2012discovering,
  title={Discovering regions of different functions in a city using human mobility and POIs},
  author={Yuan, Jing and Zheng, Yu and Xie, Xing},
  booktitle={Proceedings of the 18th ACM SIGKDD international conference on Knowledge discovery and data mining},
  pages={186--194},
  year={2012}
}

@inproceedings{chen2020simple,
  title={Simple and deep graph convolutional networks},
  author={Chen, Ming and Wei, Zhewei and Huang, Zengfeng and Ding, Bolin and Li, Yaliang},
  booktitle={International conference on machine learning},
  pages={1725--1735},
  year={2020},
  organization={PMLR}
}

@article{hu2021adaptive,
  title={Adaptive hypergraph auto-encoder for relational data clustering},
  author={Hu, Youpeng and Li, Xunkai and Wang, Yujie and Wu, Yixuan and Zhao, Yining and Yan, Chenggang and Yin, Jian and Gao, Yue},
  journal={IEEE Transactions on Knowledge and Data Engineering},
  year={2021},
  publisher={IEEE}
}

@article{yang2018fast,
  title={Fast map matching, an algorithm integrating hidden Markov model with precomputation},
  author={Yang, Can and Gidofalvi, Gyozo},
  journal={International Journal of Geographical Information Science},
  volume={32},
  number={3},
  pages={547--570},
  year={2018},
  publisher={Taylor \& Francis}
}

@inproceedings{perozzi2014deepwalk,
  title={Deepwalk: Online learning of social representations},
  author={Perozzi, Bryan and Al-Rfou, Rami and Skiena, Steven},
  booktitle={Proceedings of the 20th ACM SIGKDD international conference on Knowledge discovery and data mining},
  pages={701--710},
  year={2014}
}

@inproceedings{grover2016node2vec,
  title={node2vec: Scalable feature learning for networks},
  author={Grover, Aditya and Leskovec, Jure},
  booktitle={Proceedings of the 22nd ACM SIGKDD international conference on Knowledge discovery and data mining},
  pages={855--864},
  year={2016}
}

@inproceedings{tang2015line,
  title={Line: Large-scale information network embedding},
  author={Tang, Jian and Qu, Meng and Wang, Mingzhe and Zhang, Ming and Yan, Jun and Mei, Qiaozhu},
  booktitle={Proceedings of the 24th international conference on world wide web},
  pages={1067--1077},
  year={2015}
}

@inproceedings{wang2016structural,
  title={Structural deep network embedding},
  author={Wang, Daixin and Cui, Peng and Zhu, Wenwu},
  booktitle={Proceedings of the 22nd ACM SIGKDD international conference on Knowledge discovery and data mining},
  pages={1225--1234},
  year={2016}
}

@inproceedings{morris2019weisfeiler,
  title={Weisfeiler and leman go neural: Higher-order graph neural networks},
  author={Morris, Christopher and Ritzert, Martin and Fey, Matthias and Hamilton, William L and Lenssen, Jan Eric and Rattan, Gaurav and Grohe, Martin},
  booktitle={Proceedings of the AAAI conference on artificial intelligence},
  volume={33},
  number={01},
  pages={4602--4609},
  year={2019}
}

@inproceedings{yao2019revisiting,
  title={Revisiting spatial-temporal similarity: A deep learning framework for traffic prediction},
  author={Yao, Huaxiu and Tang, Xianfeng and Wei, Hua and Zheng, Guanjie and Li, Zhenhui},
  booktitle={Proceedings of the AAAI conference on artificial intelligence},
  volume={33},
  number={01},
  pages={5668--5675},
  year={2019}
}

@inproceedings{liang2019urbanfm,
  title={Urbanfm: Inferring fine-grained urban flows},
  author={Liang, Yuxuan and Ouyang, Kun and Jing, Lin and Ruan, Sijie and Liu, Ye and Zhang, Junbo and Rosenblum, David S and Zheng, Yu},
  booktitle={Proceedings of the 25th ACM SIGKDD international conference on knowledge discovery \& data mining},
  pages={3132--3142},
  year={2019}
}

@inproceedings{pan2019urban,
  title={Urban traffic prediction from spatio-temporal data using deep meta learning},
  author={Pan, Zheyi and Liang, Yuxuan and Wang, Weifeng and Yu, Yong and Zheng, Yu and Zhang, Junbo},
  booktitle={Proceedings of the 25th ACM SIGKDD international conference on knowledge discovery \& data mining},
  pages={1720--1730},
  year={2019}
}

@article{kipf2016variational,
  title={Variational graph auto-encoders},
  author={Kipf, Thomas N and Welling, Max},
  journal={arXiv preprint arXiv:1611.07308},
  year={2016}
}

@article{hamilton2017inductive,
  title={Inductive representation learning on large graphs},
  author={Hamilton, Will and Ying, Zhitao and Leskovec, Jure},
  journal={Advances in neural information processing systems},
  volume={30},
  year={2017}
}

@inproceedings{wang2019learning,
  title={Learning embeddings of intersections on road networks},
  author={Wang, Meng-xiang and Lee, Wang-Chien and Fu, Tao-yang and Yu, Ge},
  booktitle={Proceedings of the 27th ACM SIGSPATIAL International Conference on Advances in Geographic Information Systems},
  pages={309--318},
  year={2019}
}

@inproceedings{wu2020learning,
  title={Learning effective road network representation with hierarchical graph neural networks},
  author={Wu, Ning and Zhao, Xin Wayne and Wang, Jingyuan and Pan, Dayan},
  booktitle={Proceedings of the 26th ACM SIGKDD International Conference on Knowledge Discovery \& Data Mining},
  pages={6--14},
  year={2020}
}

@inproceedings{chen2021robust,
  title={Robust Road Network Representation Learning: When Traffic Patterns Meet Traveling Semantics},
  author={Chen, Yile and Li, Xiucheng and Cong, Gao and Bao, Zhifeng and Long, Cheng and Liu, Yiding and Chandran, Arun Kumar and Ellison, Richard},
  booktitle={Proceedings of the 30th ACM International Conference on Information \& Knowledge Management},
  pages={211--220},
  year={2021}
}

@inproceedings{yin2021multimodal,
  title={Multimodal Fusion of Satellite Images and Crowdsourced GPS Traces for Robust Road Attribute Detection},
  author={Yin, Yifang and Tran, An and Zhang, Ying and Hu, Wenmiao and Wang, Guanfeng and Varadarajan, Jagannadan and Zimmermann, Roger and Ng, See-Kiong},
  booktitle={Proceedings of the 29th International Conference on Advances in Geographic Information Systems},
  pages={107--116},
  year={2021}
}

@article{kingma2015adam,
	title={Adam: A method for stochastic optimization},
	author={Kingma, Diederik P and Ba, Jimmy},
	journal={ICLR},
	year={2015}
}

@article{van2008visualizing,
  title={Visualizing data using t-SNE.},
  author={Van der Maaten, Laurens and Hinton, Geoffrey},
  journal={Journal of machine learning research},
  volume={9},
  number={11},
  year={2008}
}

@inproceedings{qiu2018network,
  title={Network embedding as matrix factorization: Unifying deepwalk, line, pte, and node2vec},
  author={Qiu, Jiezhong and Dong, Yuxiao and Ma, Hao and Li, Jian and Wang, Kuansan and Tang, Jie},
  booktitle={Proceedings of the eleventh ACM international conference on web search and data mining},
  pages={459--467},
  year={2018}
}

@article{mikolov2013distributed,
  title={Distributed representations of words and phrases and their compositionality},
  author={Mikolov, Tomas and Sutskever, Ilya and Chen, Kai and Corrado, Greg S and Dean, Jeff},
  journal={Advances in neural information processing systems},
  volume={26},
  year={2013}
}

@article{kipf2016semi,
  title={Semi-supervised classification with graph convolutional networks},
  author={Kipf, Thomas N and Welling, Max},
  journal={arXiv preprint arXiv:1609.02907},
  year={2016}
}

@article{velivckovic2017graph,
  title={Graph attention networks},
  author={Veli{\v{c}}kovi{\'c}, Petar and Cucurull, Guillem and Casanova, Arantxa and Romero, Adriana and Lio, Pietro and Bengio, Yoshua},
  journal={arXiv preprint arXiv:1710.10903},
  year={2017}
}

@article{xu2018powerful,
  title={How powerful are graph neural networks?},
  author={Xu, Keyulu and Hu, Weihua and Leskovec, Jure and Jegelka, Stefanie},
  journal={arXiv preprint arXiv:1810.00826},
  year={2018}
}

@inproceedings{hu2019stochastic,
  title={Stochastic weight completion for road networks using graph convolutional networks},
  author={Hu, Jilin and Guo, Chenjuan and Yang, Bin and Jensen, Christian S},
  booktitle={2019 IEEE 35th International Conference on Data Engineering (ICDE)},
  pages={1274--1285},
  year={2019},
  organization={IEEE}
}

@inproceedings{guo2019attention,
  title={Attention based spatial-temporal graph convolutional networks for traffic flow forecasting},
  author={Guo, Shengnan and Lin, Youfang and Feng, Ning and Song, Chao and Wan, Huaiyu},
  booktitle={Proceedings of the AAAI conference on artificial intelligence},
  volume={33},
  number={01},
  pages={922--929},
  year={2019}
}

@inproceedings{wozniak2021hex2vec,
  title={hex2vec: Context-Aware Embedding H3 Hexagons with OpenStreetMap Tags},
  author={Wo{\'z}niak, Szymon and Szyma{\'n}ski, Piotr},
  booktitle={Proceedings of the 4th ACM SIGSPATIAL International Workshop on AI for Geographic Knowledge Discovery},
  pages={61--71},
  year={2021}
}

@article{zhou2006learning,
  title={Learning with hypergraphs: Clustering, classification, and embedding},
  author={Zhou, Dengyong and Huang, Jiayuan and Sch{\"o}lkopf, Bernhard},
  journal={Advances in neural information processing systems},
  volume={19},
  year={2006}
}

@inproceedings{li2013link,
  title={Link prediction in social networks based on hypergraph},
  author={Li, Dong and Xu, Zhiming and Li, Sheng and Sun, Xin},
  booktitle={Proceedings of the 22nd international conference on world wide web},
  pages={41--42},
  year={2013}
}

@inproceedings{zhang2018beyond,
  title={Beyond link prediction: Predicting hyperlinks in adjacency space},
  author={Zhang, Muhan and Cui, Zhicheng and Jiang, Shali and Chen, Yixin},
  booktitle={Proceedings of the AAAI Conference on Artificial Intelligence},
  volume={32},
  number={1},
  year={2018}
}

@inproceedings{tu2018structural,
  title={Structural deep embedding for hyper-networks},
  author={Tu, Ke and Cui, Peng and Wang, Xiao and Wang, Fei and Zhu, Wenwu},
  booktitle={Proceedings of the AAAI Conference on Artificial Intelligence},
  volume={32},
  number={1},
  year={2018}
}

@article{zhang2019hyper,
  title={Hyper-SAGNN: a self-attention based graph neural network for hypergraphs},
  author={Zhang, Ruochi and Zou, Yuesong and Ma, Jian},
  journal={arXiv preprint arXiv:1911.02613},
  year={2019}
}

@inproceedings{feng2019hypergraph,
  title={Hypergraph neural networks},
  author={Feng, Yifan and You, Haoxuan and Zhang, Zizhao and Ji, Rongrong and Gao, Yue},
  booktitle={Proceedings of the AAAI Conference on Artificial Intelligence},
  volume={33},
  number={01},
  pages={3558--3565},
  year={2019}
}

@inproceedings{jiang2019dynamic,
  title={Dynamic Hypergraph Neural Networks.},
  author={Jiang, Jianwen and Wei, Yuxuan and Feng, Yifan and Cao, Jingxuan and Gao, Yue},
  booktitle={IJCAI},
  pages={2635--2641},
  year={2019}
}

@article{devlin2018bert,
  title={Bert: Pre-training of deep bidirectional transformers for language understanding},
  author={Devlin, Jacob and Chang, Ming-Wei and Lee, Kenton and Toutanova, Kristina},
  journal={arXiv preprint arXiv:1810.04805},
  year={2018}
}

@article{mai2020multi,
  title={Multi-scale representation learning for spatial feature distributions using grid cells},
  author={Mai, Gengchen and Janowicz, Krzysztof and Yan, Bo and Zhu, Rui and Cai, Ling and Lao, Ni},
  journal={arXiv preprint arXiv:2003.00824},
  year={2020}
}

@article{zhu2020bilinear,
  title={Bilinear graph neural network with neighbor interactions},
  author={Zhu, Hongmin and Feng, Fuli and He, Xiangnan and Wang, Xiang and Li, Yan and Zheng, Kai and Zhang, Yongdong},
  journal={arXiv preprint arXiv:2002.03575},
  year={2020}
}

@inproceedings{he2016deep,
  title={Deep residual learning for image recognition},
  author={He, Kaiming and Zhang, Xiangyu and Ren, Shaoqing and Sun, Jian},
  booktitle={Proceedings of the IEEE conference on computer vision and pattern recognition},
  pages={770--778},
  year={2016}
}

@article{vaswani2017attention,
  title={Attention is all you need},
  author={Vaswani, Ashish and Shazeer, Noam and Parmar, Niki and Uszkoreit, Jakob and Jones, Llion and Gomez, Aidan N and Kaiser, {\L}ukasz and Polosukhin, Illia},
  journal={Advances in neural information processing systems},
  volume={30},
  year={2017}
}

@article{lu2019vilbert,
  title={Vilbert: Pretraining task-agnostic visiolinguistic representations for vision-and-language tasks},
  author={Lu, Jiasen and Batra, Dhruv and Parikh, Devi and Lee, Stefan},
  journal={Advances in neural information processing systems},
  volume={32},
  year={2019}
}

@inproceedings{paszke2019pytorch,
	title={PyTorch: An imperative style, high-performance deep learning library},
	author={Paszke, Adam and Gross, Sam and Massa, Francisco and Lerer, Adam and Bradbury, James and Chanan, Gregory and Killeen, Trevor and Lin, Zeming and Gimelshein, Natalia and Antiga, Luca and others},
	booktitle={Advances in Neural Information Processing Systems},
	pages={8024--8035},
	year={2019}
}

@article{jepsen2020relational,
  title={Relational fusion networks: Graph convolutional networks for road networks},
  author={Jepsen, Tobias Skovgaard and Jensen, Christian S and Nielsen, Thomas Dyhre},
  journal={IEEE Transactions on Intelligent Transportation Systems},
  volume={23},
  number={1},
  pages={418--429},
  year={2020},
  publisher={IEEE}
}

@String{Computing = "Computing" }

@String{Computer = "{IEEE} Computer" }

@ArtifactSoftware{R,
    title = {R: A Language and Environment for Statistical Computing},
    author = {{R Core Team}},
    organization = {R Foundation for Statistical Computing},
    address = {Vienna, Austria},
    year = {2019},
    url = {https://www.R-project.org/},
}

%%
%% If your work has an appendix, this is the place to put it.
% \appendix

% \input{inputs/response.tex}

\end{document}